%% file: main.tex
\documentclass{article}

\input{preamble/packages}

\input{preamble/variables}

\input{preamble/commands}

\input{preamble/icml-fixes}

\begin{document}

\twocolumn[
  \input{front/titlepage}

  \vskip 0.3in
]

\printAffiliationsAndNotice{} 

\setcounter{tocdepth}{3}
\etocsettocstyle{}{}
\etocsetnexttocdepth{-2}
\tableofcontents

\input{front/abstract}

\begin{figure}[ht]
    \centering
    \includegraphics[width=0.85\columnwidth]{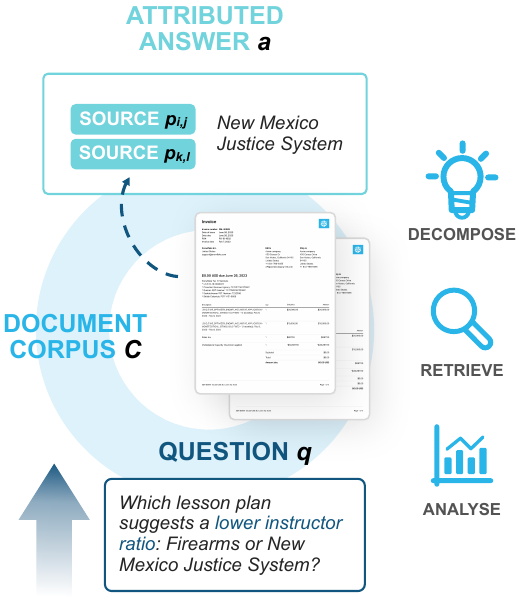}
    \caption{Given a query $q$ over corpus $\mathcal{C}$, the system iteratively retrieves pages, reasons over visual and textual content, and aggregates evidence from multiple pages $\mathcal{E} = \{p_{i,j}, \ldots\}$ to produce a grounded answer $a$ with attribution. The process typically requires decomposing $q$, iterative retrieval, and synthesizing across $\mathcal{E}$.}
    \label{fig:hero}
\end{figure}


\input{sections/introduction}
\input{sections/dataset}
\input{sections/evaluation}
\input{sections/baselines}

\input{sections/experiments}
\input{sections/conclusion}

\clearpage

\input{sections/limitations}



\bibliography{bib/references}
\bibliographystyle{icml2026}


\input{appendix/appendix-main}

\end{document}

%% file: preamble/packages.tex
\usepackage{microtype}
\usepackage{graphicx}
\usepackage{subcaption}
\usepackage{booktabs}

\usepackage[normalem]{ulem}

\usepackage[linkcolor={1 0 0}]{hyperref}

\usepackage{amsmath}
\usepackage{amssymb}
\usepackage{mathtools}
\usepackage{amsthm}

\usepackage[capitalize,noabbrev]{cleveref}

\usepackage[textsize=tiny]{todonotes}

\usepackage{varwidth}
\usepackage[dvipsnames*,svgnames]{xcolor}
\usepackage{multirow}
\usepackage{bigdelim}
\usepackage{pifont}
\usepackage{algorithm}
\usepackage{algpseudocode}
\usepackage{longtable}
\usepackage{tabularx}
\usepackage{adjustbox}
\usepackage[toc,page,header]{appendix}
\usepackage{etoc}
\usepackage{makecell}
\usepackage[most]{tcolorbox}
\usepackage{array}
\usepackage{fontawesome}
\usepackage{wasysym}
\usepackage[table]{xcolor}

\usepackage[inline]{enumitem}

\usepackage{caption}

\DeclareCaptionType{prompt}[Prompt][List of Prompts]

\usepackage[english]{babel}
\usepackage[autostyle, english = american]{csquotes}
\MakeOuterQuote{"}

\colorlet{saved}{.}  

\hypersetup{allcolors=saved}

%% file: preamble/variables.tex
\newcommand{\benchmarkname}{Multimodal Agentic Document QA}
\newcommand{\benchmarknameshort}{\textsc{MADQA}}

\newcommand{\taskname}{Agentic Document Collection Visual Question Answering}
\newcommand{\tasknameshort}{Agentic Document Collection VQA}

\newcommand{\runningtitle}{\benchmarkname}

\newcommand{\papertitle}{Strategic Navigation or Stochastic Search?\\How Agents and Humans Reason Over Document Collections}

\newcommand{\anlsstar}{ANLS*}
\newcommand{\pagef}{Page F1}
\newcommand{\docf}{Doc F1}
\newcommand{\kuiper}{Kuiper}

\newcommand{\numquestions}{2,250}
\newcommand{\numdocs}{800}
\newcommand{\numpages}{18,619}
\newcommand{\numdomains}{13}
\newcommand{\numcategories}{63}
\newcommand{\multihoppercent}{17.3\%}

\newcommand{\testsize}{500}
\newcommand{\devsize}{200}
\newcommand{\trainsize}{1,550}

%% file: preamble/commands.tex
\theoremstyle{plain}
\newtheorem{theorem}{Theorem}[section]

\theoremstyle{definition}
\newtheorem{definition}[theorem]{Definition}

\theoremstyle{remark}

\definecolor{snow}{HTML}{53B0E2}
\definecolor{codepurple}{HTML}{D45B90}
\definecolor{valencia}{HTML}{FF9F36}
\definecolor{star}{HTML}{75CDD7}
\definecolor{snowpurple}{HTML}{7254A3}
\definecolor{snowlight}{HTML}{7AC5EA}
\definecolor{snowdark}{HTML}{1F4463}
\definecolor{snoworange}{HTML}{FF9F36}
\definecolor{snowpink}{HTML}{D45B90}
\definecolor{snowgray}{HTML}{5B5B5B}
\definecolor{hyperlinkgray}{HTML}{333333}
\definecolor{snowstar}{HTML}{75CDD7}
\definecolor{gray}{HTML}{BDBDBD}

\newcommand{\cmark}{\tikz[baseline=-0.6ex]{}{\color{star}\ding{51}}}
\newcommand{\xmark}{{\color{codepurple}\ding{55}}}

\newcommand{\medium}{%
  \tikz[baseline=-0.6ex]{\fill[valencia] (0,0) circle (3pt);} medium%
}
\newcommand{\high}{%
  \tikz[baseline=-0.6ex]{\fill[star] (0,0) circle (3pt);} high%
}
\newcommand{\low}{%
  \tikz[baseline=-0.6ex]{\fill[codepurple] (0,0) circle (3pt);} low%
}

\newcolumntype{P}[1]{>{\centering\arraybackslash}p{#1}}

\makeatletter
\newtcbox{\bluebox}[1][]{enhanced jigsaw, 
      sharp corners,
      colback=LightSteelBlue,
      center,
      varwidth upper=0.8\linewidth,
      frame hidden,
      #1}
\makeatother

\renewcommand{\thepart}{}
\renewcommand{\partname}{}

\newcommand{\jvl}[1]{#1}

\newcommand{\hflink}{https://huggingface.co/spaces/Snowflake/MADQA-Leaderboard}
\newcommand{\ghlink}{https://github.com/OxRML/MADQA}
\newcommand{\hflinkshow}{Snowflake/MADQA-Leaderboard}
\newcommand{\ghlinkshow}{OxRML/MADQA}
\newcommand{\dslink}{https://huggingface.co/datasets/Snowflake/MADQA}

\newcommand{\datalink}{https://huggingface.co/datasets/OxRML/MADQA}
\newcommand{\datalinkshow}{OxRML/MADQA}

\def\snowone{\raisebox{-2.9pt}{\includegraphics[height=1.4em]{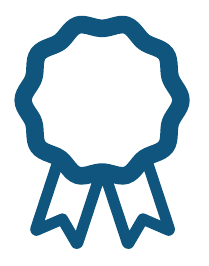}}\hspace{0.3em}}

\def\snowthree{\raisebox{-2.9pt}{\includegraphics[height=1.4em]{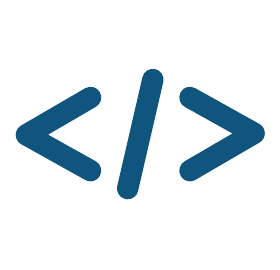}}\hspace{0.3em}}

\def\snowfour{\raisebox{-2.9pt}{\includegraphics[height=1.4em]{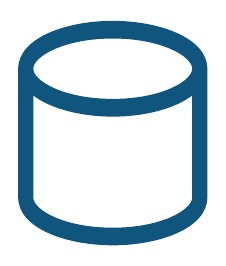}}\hspace{0.3em}}

\tcbset{
    promptbox/.style={
        enhanced,
        colback=white,
        colframe=snowdark,
        boxrule=1pt,
        arc=2pt,
        left=0pt, right=0pt, top=0pt, bottom=0pt,
        fonttitle=\sffamily,
        coltitle=white,
        colbacktitle=snowdark,
        attach boxed title to top left={yshift=-2mm, xshift=4mm},
        boxed title style={boxrule=0pt, arc=1pt}
    },
    promptbox-notitle/.style={
        enhanced,
        colback=white,
        colframe=snowdark,
        boxrule=1pt,
        arc=2pt,
        left=0pt, right=0pt, top=0pt, bottom=0pt,
        fonttitle=\sffamily,
        coltitle=white,
        colbacktitle=snowdark,
    }
}

\tcbset{
    promptsection/.style={
        enhanced,
        boxrule=0pt,
        leftrule=3pt,
        arc=0pt,
        left=6pt, right=6pt, top=4pt, bottom=4pt
    }
}

\tcbset{
    psec-role/.style={promptsection, colback=gray!5, colframe=gray!50},
    psec-input/.style={promptsection, colback=snowpurple!5, colframe=snowpurple!50},
    psec-warning/.style={promptsection, colback=valencia!10, colframe=valencia},
    psec-output/.style={promptsection, colback=snowstar!10, colframe=snowstar},
    psec-steps/.style={promptsection, colback=snowlight!5, colframe=snowlight!80!black},
    psec-error/.style={promptsection, colback=codepurple!5, colframe=codepurple},
    psec-correct/.style={promptsection, colback=star!10, colframe=star}
}

%% file: preamble/icml-fixes.tex
\let\oldaddcontentsline\addcontentsline

\expandafter\def\csname ver@algorithmic.sty\endcsname{}

\usepackage[preprint]{icml2026} 

\let\addcontentsline\oldaddcontentsline

\icmltitlerunning{\runningtitle}

\ificmlshowauthors
  \newcommand{\affilicon}[1]{%
    \hspace{1ex}\raisebox{-1ex}{\includegraphics[height=3.7ex]{icons/#1.png}}%
  }
  \newcommand{\affilicontop}[1]{%
    \hspace{1pt}\raisebox{0.4ex}{\includegraphics[height=2.3ex]{icons/#1.png}}}
    
  \makeatletter
  \renewcommand{\@pa}[1]{%
    \ifcsname @icmlsymbol#1\endcsname
      \textsuperscript{\csname @icmlsymbol#1\endcsname\,}%
    \else
      \ifcsname the@affil#1\endcsname\else
        \stepcounter{@affiliationcounter}\newcounter{@affil#1}\setcounter{@affil#1}{\value{@affiliationcounter}}%
      \fi
      \ifnum\pdf@strcmp{#1}{snowflake}=0 \affilicontop{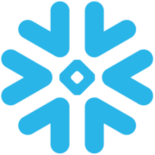}%
      \else\ifnum\pdf@strcmp{#1}{instabase}=0 \affilicontop{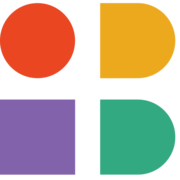}%
      \else\ifnum\pdf@strcmp{#1}{oxford}=0 \affilicontop{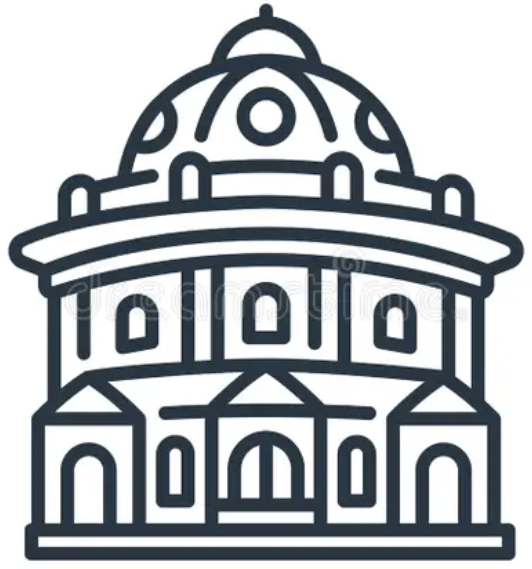}%
      \else\ifnum\pdf@strcmp{#1}{huggingface}=0 \affilicontop{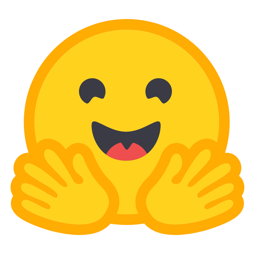}%
      \else\ifnum\pdf@strcmp{#1}{cvc}=0 \affilicontop{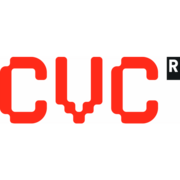}%
      \else\ifnum\pdf@strcmp{#1}{unc}=0 \affilicontop{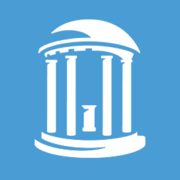}%
      \else\textsuperscript{\arabic{@affil#1}}\fi\fi\fi\fi\fi\fi 
    \fi
  }

  \renewcommand{\printAffiliationsAndNotice}[1]{\global\icml@noticeprintedtrue%
    \stepcounter{@affiliationcounter}%
    {\let\thefootnote\relax\footnotetext{%
      \vspace{0.2em}\hspace*{-2.2em}%
      \begin{minipage}{1.15\linewidth}%
      \setlength{\parindent}{0pt}%
      \sloppy
      \raggedright
      \ificmlshowauthors #1\fi%
        \forloop{@affilnum}{1}{\value{@affilnum} < \value{@affiliationcounter}}{%
          \expandafter\let\expandafter\tempaffilname\csname @affilname\the@affilnum\endcsname
          \textsuperscript{%
             \ifnum\value{@affilnum}=1 \affilicon{snowflake}%
             \else\ifnum\value{@affilnum}=2 \affilicon{instabase}%
             \else\ifnum\value{@affilnum}=3 \affilicon{oxford}%
             \else\ifnum\value{@affilnum}=4 \affilicon{huggingface}%
             \else\ifnum\value{@affilnum}=5 \affilicon{unc}%
             \else\ifnum\value{@affilnum}=6 \affilicon{cvc}%
             \else\arabic{@affilnum}\fi\fi\fi\fi\fi\fi
          }\,\tempaffilname
          \space
        }%
        \ifdefined\icmlcorrespondingauthor@text
           \vspace{0.7em}\\\hspace{0.3em}Correspondence to: \icmlcorrespondingauthor@text
        \fi
        \end{minipage}%
      }%
    }%
  }
  \makeatother
\fi

%% file: front/titlepage.tex

\icmltitle{\papertitle}

\icmlsetsymbol{equal}{*}

\begin{icmlauthorlist} 
  \icmlauthor{Łukasz Borchmann}{snowflake}
  \icmlauthor{Jordy Van Landeghem}{instabase}
  \icmlauthor{Michał Turski}{snowflake}
  \icmlauthor{Shreyansh Padarha}{oxford}\\
  \icmlauthor{Ryan Othniel Kearns}{oxford}
  \icmlauthor{Adam Mahdi}{oxford}
  \icmlauthor{Niels Rogge}{huggingface}
  \icmlauthor{Clémentine Fourrier}{huggingface}\\
  \icmlauthor{Siwei Han}{unc}
  \icmlauthor{Huaxiu Yao}{unc}
  \icmlauthor{Artemis Llabrés}{cvc}
  \icmlauthor{Yiming Xu}{cvc}
  \icmlauthor{Dimosthenis Karatzas}{cvc}\\
  \icmlauthor{Hao Zhang}{snowflake}
  \icmlauthor{Anupam Datta}{snowflake}
\end{icmlauthorlist}

\icmlaffiliation{snowflake}{Snowflake\,AI\,Research}
\icmlaffiliation{instabase}{Instabase}
\icmlaffiliation{oxford}{University\,of\,Oxford}
\icmlaffiliation{huggingface}{Huggingface}
\icmlaffiliation{cvc}{Computer\,Vision\,Center}
\icmlaffiliation{unc}{UNC-Chapel\,Hill}

\icmlcorrespondingauthor{}{lukasz.borchmann@snowflake.com}

\icmlkeywords{Machine Learning, ICML}


\begin{center}
\vspace{0.4cm}
{\hypersetup{allcolors=snowpurple}\begin{tabular}{rllrllll}
\snowone \href{\hflink}{\hflinkshow} &
\snowthree \href{\ghlink}{\ghlinkshow} &
\snowfour \href{\datalink}{\datalinkshow}
\end{tabular}}
\vspace{-0.1cm}
\end{center}

%% file: front/abstract.tex

\begin{abstract}
Multimodal agents offer a promising path to automating complex document-intensive workflows. Yet, a critical question remains: do these agents demonstrate genuine strategic reasoning, or merely stochastic trial-and-error search?  To address this, we introduce \benchmarknameshort{}, a benchmark of \numquestions{} human-authored questions grounded in \numdocs{} heterogeneous PDF documents. Guided by \textit{Classical Test Theory}, we design it to maximize discriminative power across varying levels of agentic abilities. To evaluate agentic behaviour, we introduce a novel evaluation protocol measuring the accuracy-effort trade-off. Using this framework, we show that while the best agents can match human searchers in raw accuracy, they succeed on largely different questions and rely on brute-force search to compensate for weak strategic planning. They fail to close the nearly 20\% gap to oracle performance, persisting in unproductive loops. We release the dataset and evaluation harness to help facilitate the transition from brute-force retrieval to calibrated, efficient reasoning.
\end{abstract}

%% file: sections/introduction.tex

\section{Introduction}

We introduce \textit{\benchmarkname{}} benchmark (\benchmarknameshort{}).
It focuses on the capabilities of multimodal large language model (MLLM) based \textit{agentic} systems to handle complex, multi-stage information retrieval and reasoning tasks that one would encounter in enterprise settings.   

\subsection{Motivation and Related Works}

\input{tables/benchmark_comparison}

Current benchmarks in this area operate in a fragmented landscape, as detailed in Table~\ref{tab:benchmark_comparison} (Appendix~\ref{app:related} provides per-benchmark justifications). 

\paragraph{\textit{(i)} Format.}
Agent-centric benchmarks such as Researchy Questions \cite{10.1145/3726302.3730275} and BRIGHT \cite{su2025brightrealisticchallengingbenchmark} capture the complexity of ``agentic research.'' 
However, they rely on HTML or plain text, disregarding visual comprehension required for real-world documents.

\paragraph{\textit{(ii)} Scope.}
Domain-specific PDF benchmarks such as FinRAGBench-V \cite{zhao2025finragbenchvbenchmarkmultimodalrag} and ViDoSeek \cite{wang2025vidoragvisualdocumentretrievalaugmented} confront visually-rich PDFs but restrict evaluation to narrow verticals like finance or rely on single-step metrics that fail to capture iterative planning and refinement.

\paragraph{\textit{(iii)} Data Provenance.}\looseness=-1
General-purpose document benchmarks attempt to bridge these gaps but suffer from methodological concerns. ViDoRE \cite{faysse2025colpaliefficientdocumentretrieval} relies on MLLM-generated questions, risking bias toward similar models. ViDoRe v3 \cite{vidorev3} mitigates this with contextually-blind generation and extensive human verification, but the generating model still shapes the distribution of question types---a concern absent from fully human-authored benchmarks. 
Similarly, VIMDoc \cite{kim2025hybrid} and DOUBLE-BENCH \cite{shen2025we} recycle documents from older datasets (e.g., DocVQA, Wikipedia), increasing the risk of data contamination. 

\noindent We addresses all three limitations: combine the high layout diversity of a large-scale, heterogeneous PDF collection \textit{(format)} with broad domain coverage and multi-step reasoning requirements \textit{(scope)}, grounded in rigorous, fully human-authored questions over fresh documents \textit{(data prevenance)}.

\subsection{Guiding Principles}
\label{sec:problem_statement}

\benchmarknameshort{} exemplifies \textit{\taskname}.
Given a corpus $\mathcal{C}$ of multi-page documents and a natural language query $q$, the task is to produce an answer $a$ and a minimal evidence set $\mathcal{E} \subseteq \mathcal{C}$ (see Appendix~\ref{app:problem_statement} for full formulation). Six properties distinguish this task from standard document QA:
\begin{table}[H]
    \centering
    \small
    \label{tab:properties}
    \vspace{0.3em}
    \begin{tabular}{@{}llp{5cm}@{}}
        \toprule
        \textbf{\#} & \textbf{Property} & \textbf{Definition} \\
        \midrule
        1 & Extractive & Answer tokens must appear physically in the evidence set $\mathcal{E}$. \\
        2 & Multi-Hop & $\mathcal{E}$ may span disjoint pages (cross-page) or documents (cross-doc). \\
        3 & Closed-World & Answer derived solely from $\mathcal{C}$; no external parametric knowledge. \\
        4 & Grounded & $\mathcal{E}$ must entail $a$ and be minimal (no superfluous pages). \\
        5 & Agentic & No single retrieval query $q'$ may exist such that $\textsc{Retrieve}(q') \supseteq \mathcal{E}$. \\
        6 & Visual & Answering may require non-textual information (layout, tables, figures) in $\mathcal{E}$. \\
        \bottomrule
    \end{tabular}
    \vspace{-1em}
\end{table}

    
    

Properties 1, 3, and 4 are \textit{enforced by construction}: answers must be \textit{extractive} (tokens appear in evidence), derived under a \textit{closed-world} assumption, and \textit{attributed} to minimal evidence such that evaluation rewards genuine corpus use.

Properties 2, 5, 6 are \textit{targeted by design}: our annotation protocol encourages questions requiring \textit{multi-hop} reasoning across disjoint pages or documents, under conditions where single-query retrieval is unlikely to surface all evidence (\textit{agentic}), or where \textit{visual} comprehension of layout structure is beneficial.

The \textit{agentic} property, when satisfied, necessitates planning (decomposing $q$ into sub-queries), navigation (iterating on intermediate findings), and aggregation (synthesizing partial answers).

\subsection{Contributions}

\paragraph{Task Formalization.} We formally define \emph{\tasknameshort{}} 
with six core properties 
that distinguish it from prior document QA formulations (\S\ref{sec:problem_statement}).

\paragraph{Validated Benchmark.}  We release a fully human-authored dataset of \numquestions{} questions over \numdocs{} heterogeneous, fresh PDF documents  (\S\ref{sec:dataset}). Unlike prior work, we operationalize a Construct Validity framework \cite{bean2025measuringmattersconstructvalidity} to certify the benchmark's integrity. 
(\S\ref{sec:validity}-\ref{sec:subset}).

\paragraph{Evaluation Protocol.} We provide principled methods of measuring answer correctness, 
evidence attribution, and a novel metric for effort calibration, all directly motivated by our formal properties (\S\ref{sec:evaluation}).

\paragraph{Principled Split Creation.} We eploy \textit{Classical Test Theory} to derive a test set with strong rank correlation to the full benchmark while reserving hard items for long-term relevance, enabling low-cost experimentation (\S\ref{sec:subset}).

\paragraph{Human vs.\ Agent Comparison.} We provide the first comparative study of human vs.\ agentic research behaviors in this domain. Results reveal a critical \textit{efficiency gap} and fundamentally different competencies (\S\ref{sec:human-agent}). 

\paragraph{Efficiency Analysis.} We compare static RAG, unconstrained Recursive Language Models (RLM), and tool-augmented Agents, and demonstrate that constrained agency significantly outperforms static RAG while avoiding the catastrophic effort overhead of RLMs (\S\ref{sec:experiments}). 

%% file: tables/benchmark_comparison.tex
\begin{table*}[h]
\caption{\textbf{Comparing \benchmarknameshort{} with existing benchmarks.} Our work is distinguished by its focus on a collection of complex PDFs with \emph{fresh documents} (not recycled from existing benchmarks) and \emph{fully human-authored questions}, designed to test agentic reasoning. We denote diversity levels using \textcolor{snowlight}{$\bullet$} high, \textcolor{valencia}{$\bullet$} medium, and \textcolor{snowpink}{$\bullet$} low as coarse, relative categories (see Appendix~\ref{app:related} for detailed assessment).
}\label{tab:benchmark_comparison}
\footnotesize
\renewcommand{\arraystretch}{1.1}
\begin{tabular}{@{}p{5.2cm} p{3.45cm} P{2.9cm} P{1.5cm} p{2.1cm}}
\toprule
\multirow{2}{*}{\textbf{Name and Reference}} & \multirow{2}{*}{\textbf{Input File(s)}} & \textbf{Diversity} & \textbf{Human} & \makecell[c]{\textbf{Problem}} \\
& & \textbf{(Domains~/~Layouts)} & \textbf{-annotated} & \makecell[c]{\textbf{Framing}} \\
\midrule
DocVQA \cite{mathew2021docvqadatasetvqadocument} & Single document image & \medium{} / \medium{}
& \cmark & \rdelim\}{7}{0pt}[{\parbox[t]{2.5cm}{\vspace{-35px}\raggedright ~\textbf{Document} \\ ~\textbf{VQA} \vspace{0.5em}\\ ~Question\\~grounded\\~on a single\\~rich document}}]
\\
InfographicVQA \cite{mathew2021infographicvqa} &
Single image & \medium{} / \high{}
&
\cmark
\\
TAT-DQA \cite{Zhu_2022}
 &
Mostly single-page PDF
& \low{} / \low{}
&
\cmark
\\
DUDE \cite{vanlandeghem2023documentunderstandingdatasetevaluation}
& Multi-page PDF file
& \high{} / \high{}
&
\cmark
\\
MP-DocVQA \cite{tito2023hierarchicalmultimodaltransformersmultipage} &
Multi-page PDF
& \medium{} / \medium{}
&
\cmark{} / \xmark{}
\\
SlideVQA \cite{tanaka2023slidevqadatasetdocumentvisual} &
Series of slides & \medium{} / \low{}
& 
\cmark
\\
M-LongDoc \cite{chia2024mlongdocbenchmarkmultimodalsuperlong} &
Multi-page PDF
& \medium{} /  \medium{}
& 
\cmark{} / \xmark{}
\\
MMLongBench-Doc \cite{ma2024mmlongbenchdocbenchmarkinglongcontextdocument} &
Multi-page PDF
& \high{} / \high{}
& 
\cmark
\vspace{0.6em}
\\
MuRAR \cite{zhu2025murarsimpleeffectivemultimodal} &
Collection of web pages
&  \low{} / \medium{} & 
\cmark{} / \xmark{}
& \rdelim\}{3}{0pt}[\parbox{2.5cm}{\raggedright ~\textbf{Multimodal} \\ ~\textbf{RAG}}]
\\
M$^2$RAG \cite{liu2025benchmarkingretrievalaugmentedgenerationmultimodal} &
Collection of web pages & \high{} / \high{} &
\cmark{} / \xmark{} \\
MR$^2$-Bench \cite{zhou2025mr} &
Interleaved image-text & \high{} / \high{}& \cmark{} / \xmark{}
\vspace{0.6em}
\\
ViDoRE v3 \cite{vidorev3} & 
Collection of PDFs & \high{} / \high{}
 & \cmark{} / \xmark{}
 & \rdelim\}{8}{0pt}[\parbox{2.5cm}{\raggedright ~\textbf{Document} \\ ~\textbf{RAG}  \vspace{0.5em}\\~Single-step\\~retrieval\\~and~answer}]
\\
DocBench \cite{zou2024docbenchbenchmarkevaluatingllmbased}
& Collection of PDFs & \high{} / \high{}
 & \cmark{} / \xmark{}
\\
M3DocRAG \cite{cho2024m3docragmultimodalretrievalneed} & PDFs made from web pages
& \medium{} / \low{} & \cmark{} / \xmark{}
\\
MMDocIR \cite{dong2025mmdocirbenchmarkingmultimodalretrieval} &
Collection of PDFs
& \high{} / \high{} & \cmark{} / \xmark{}
\\
FinRAGBench-V \cite{zhao2025finragbenchvbenchmarkmultimodalrag} & Collection of PDFs & \low{} / \high{} & \cmark{}
\\
VIMDoc \cite{kim2025hybrid} &
Collection of PDFs & \high{} / \high{} 
& \cmark{} / \xmark{}
\\
JINA-VDR \cite{gunther2025jina} &
Collection of PDFs & \high{} / \high{}
& \cmark{} / \xmark{}
\\
UniDoc-Bench \cite{peng2026unidocbenchunifiedbenchmarkdocumentcentric} & Collection of PDFs & \high{} / \high{}
& \cmark{} / \xmark{}
\vspace{0.6em}
\\
BRIGHT \cite{su2025brightrealisticchallengingbenchmark} &
Collection of web pages & \high{} / \medium{} & \cmark{}
& \rdelim\}{6}{0pt}[\parbox{2.5cm}{\raggedright ~\textbf{Agentic} \\ ~\textbf{Research} \vspace{0.5em}\\~Multi-step\\~answering\\~with tools}]
\\ 
Researchy Questions \cite{10.1145/3726302.3730275}
 & Collection of texts & \high{} / \high{}
 & \cmark{} / \xmark{}
 \\
ViDoSeek \cite{wang2025vidoragvisualdocumentretrievalaugmented} & Collection of slide decks &  \low{} / \high{}
&  \cmark{} / \xmark{}
\\
DOUBLE-BENCH \cite{shen2025we} &
Collection of PDFs & \medium{} / \medium{}
& \cmark{} / \xmark{} \\
MRMR \cite{zhang2025mrmr} & Interleaved image-text & \high{} / \low{} & \cmark{} / \xmark{} \\
\textbf{Our benchmark} & Collection of PDFs
& \high{} / \high{} & \cmark{}
\\ 
\bottomrule
\end{tabular}
\end{table*}

%% file: sections/dataset.tex

\section{Dataset Construction and Validity}\label{sec:dataset}

To ensure high-quality, solvable, and unambiguous benchmarks, we orchestrated a rigorous human annotation pipeline involving over 1,200 hours of professional work.

\subsection{Sourcing and Filtering Documents}

We manually curated PDFs from DocumentCloud,\footnote{\url{https://www.documentcloud.org/documents/}} intentionally seeking clusters of up to 30 related documents (e.g., sequential reports or menus from different restaurants). It was crucial for enabling realistic cross-document multi-hop questions (e.g., comparing values between documents). 

The created corpus covers \numdocs{} PDFs divided into \numcategories{} fine-grained categories in \numdomains{} high-level domains (as detailed in Table~\ref{tab:domain_distribution}), ranging from single-page summaries to extensive 800+ page filings (see Appendix~\ref{app:domains} for detailed statistics; Appendix~\ref{app:dataset_card} provides a full Dataset Card).

\paragraph{Layout and Domains Diversity.}
We extract layout elements and compute per-element z-score normalization to highlight domain-specific patterns (see Appendix~\ref{app:layout_elemt_density}). Figure~\ref{fig:domain_layout_heatmap} visualizes the structural heterogeneity of our corpus by showing layout element density across domains: financial and government documents exhibit high table density, while technical documents are figure-heavy. Legal documents show elevated text density with minimal visuals, and Reference materials (catalogs, guides) contain diverse lists.

\begin{figure}[!h]
    \centering
    \includegraphics[width=0.8\linewidth,trim={1cm 1cm 0cm 1cm},clip]{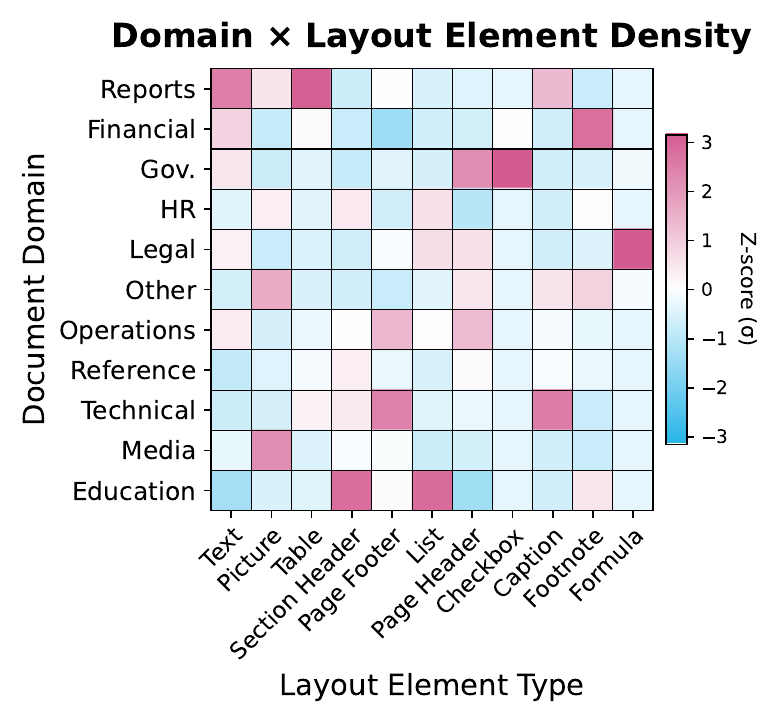}
    \caption{\textbf{Layout element density across document domains in \benchmarknameshort{}.}
    The heatmap shows the standardized (z-scored) concentration of individual layout elements within each domain.
    \textbf{\textcolor{codepurple}{Pink}} indicates above-average density, while
    \textbf{\textcolor{snowlight}{cyan}} indicates below-average density.
    A detailed discussion is provided in Appendix~\ref{app:layout_elemt_density}.}
    \label{fig:domain_layout_heatmap}
\end{figure}

\subsection{Question Annotation and Quality Assurance}
We contracted a professional data vendor with full-time employees who were experienced in labeling QA datasets over PDF documents.

\paragraph{Annotation Protocol.} We formulated strict guidelines to ensure solvability and lack of ambiguity (detailed in Appendix~\ref{app:guidelines}):
(1) Questions must be answerable entirely from the provided documents, strictly excluding external world knowledge. (2) Questions must be specific enough to pinpoint a unique answer but must not reveal the source location too easily. (3) For every question, collect the minimal set of evidence pages required to answer (tagging all pages contributing to the final answer for multi-hop queries).

\paragraph{Annotator Selection.} We provided 20 annotators with the same guidelines to annotate the same group of documents.
Their annotations were subjected to a two-step verification: (1) We provided GPT-5 with the human-annotated oracle evidence pages. If the model failed to answer correctly using the perfect context, the instance was flagged for manual check. (2) Only annotators with zero errors in manual review participated in \benchmarknameshort{} annotation.

\paragraph{Supervision, Checks, and Corrections.} Domain expert from the authors' institution maintained constant synchronization with contractors to resolve ambiguities in real-time and enforce strict adherence to the guidelines.
After the initial version of the dataset was composed, we reviewed the questions that were not answered correctly by any of the tested baseline models or any of real humans employed to establish human baseline (\S\ref{sec:baselines}). This led to replacing ($<1\%$, mostly typos) or extending gold standard ($<5\%$, mostly adding optional, desirable context).

\paragraph{Annotation Statistics.} Table~\ref{tab:domain_distribution} presents the distribution of \numquestions{} QA pairs. We targeted approximately 20\% multi-hop questions in our annotation guidelines, though without strict enforcement. The resulting distribution spread of \multihoppercent{} multi-hop, 82.7\% single-hop instances closely matches this target. 8.3\% require synthesizing information from multiple pages within the same document (\textit{X-Page}), while 9.0\% demand cross-document reasoning (\textit{X-Doc}).

\begin{table}[h]
\centering
\caption{\textbf{Constituents of ~\benchmarknameshort{}.} The corpus spans documents (median 5 pages, mean 23.3, max 859) totaling 12.2M tokens. 
}\label{tab:domain_distribution}
\footnotesize\setlength{\tabcolsep}{4pt}
\begin{tabular}{lrrrrr}
\toprule
\textbf{Domain} & \textbf{Docs} & \textbf{Pages} & \textbf{Pg/Doc} & \textbf{Qs} & \textbf{Q/Doc} \\
\midrule
Financial & 131 & 6,149 & 46.9 & 460 & 3.5 \\
Reports & 127 & 2,665 & 21.0 & 360 & 2.8 \\
Gov/Regulatory & 105 & 702 & 6.7 & 304 & 2.9 \\
Legal & 69 & 1,154 & 16.7 & 182 & 2.6 \\
HR/Employment & 68 & 813 & 12.0 & 159 & 2.3 \\
Reference & 62 & 1,292 & 20.8 & 218 & 3.5 \\
Miscellaneous & 56 & 155 & 2.8 & 92 & 1.6 \\
Events & 43 & 117 & 2.7 & 88 & 2.0 \\
Financial/Tax & 39 & 2,925 & 75.0 & 82 & 2.1 \\
Media/Publishing & 31 & 1,492 & 48.1 & 113 & 3.6 \\
Technical & 29 & 842 & 29.0 & 68 & 2.3 \\
Education & 26 & 255 & 9.8 & 68 & 2.6 \\
Cases/Logs & 14 & 58 & 4.1 & 55 & 3.9 \\
\midrule
\textbf{Total} & \textbf{800} & \textbf{\numpages} & \textbf{23.3} & \textbf{2,250} & \textbf{2.8} \\
\bottomrule
\end{tabular}
\end{table}

\begin{figure}[!t]
    \centering\vspace{1em}
    \begin{tcolorbox}[
        enhanced,
        boxrule=0pt,
        frame hidden,
        colback=snowdark!8,
        arc=2pt,
        left=8pt, right=8pt, top=13pt, bottom=17pt,
        width=\linewidth,
    ]
    \centering
    \textit{What was the total excess permit revenue in Minnesota for the 2014-2019 period?}\vspace{1em}\\
    \fcolorbox{snowdark!25}{white}{\includegraphics[width=0.87\linewidth]{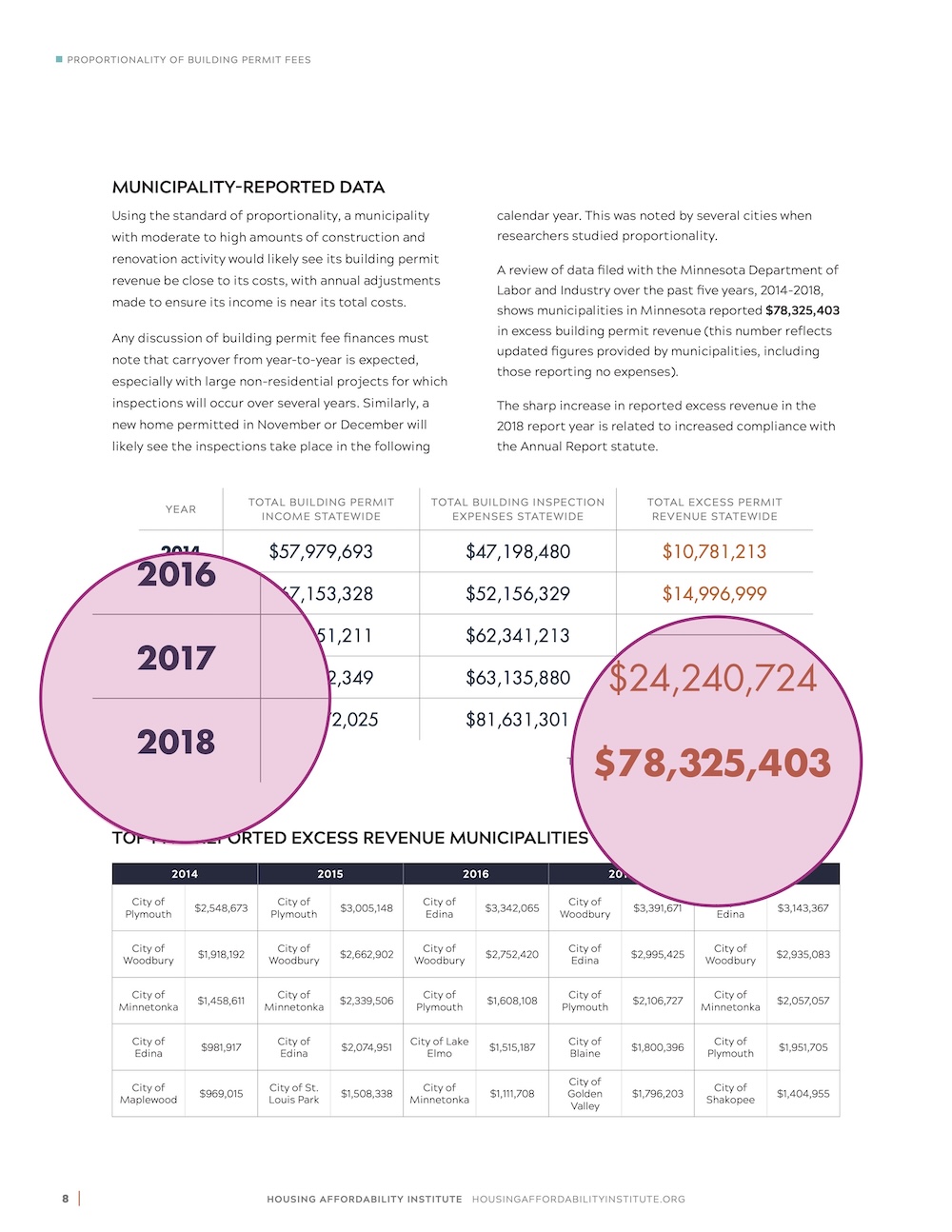}}
    \vspace{0.5em}\\
    \fcolorbox{snowdark!25}{white}{\includegraphics[width=0.87\linewidth]{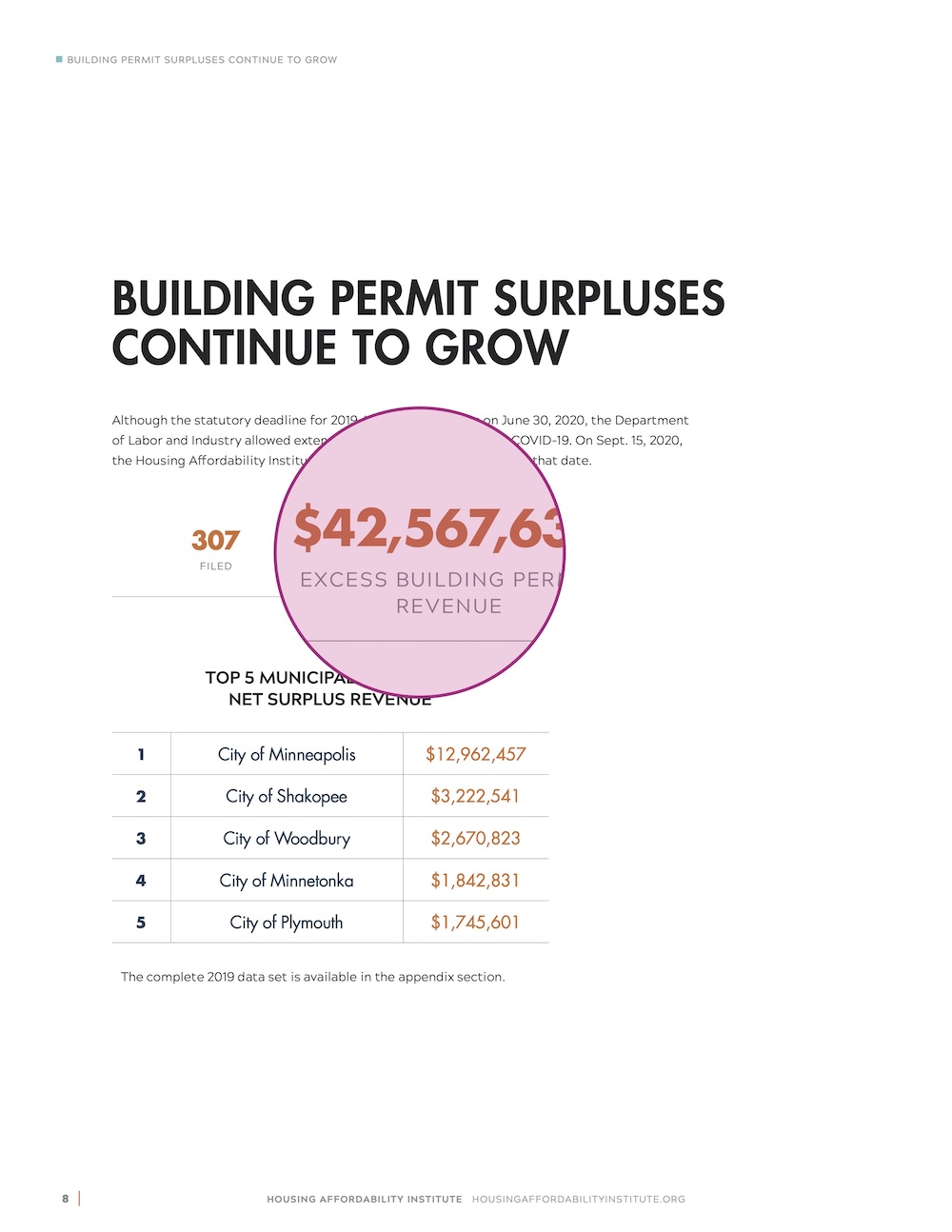}}
    \end{tcolorbox}
    \caption{\textbf{Sample question (X-Doc).} No single document covers the full period. The agent must retrieve both the 2018 report (covering 2014--2018) and the 2019 report, then extract and sum the relevant values. More examples provided in Appendix~\ref{app:examples}.}
    \label{fig:example_permit}
\end{figure}

\subsection{Construct Validity Analysis}
\label{sec:validity}

We measure to what extend guidelines from Section~\ref{sec:problem_statement} are satisfied, explicitly dissociating the construct from confounders \cite{bean2025measuringmattersconstructvalidity}.


\paragraph{Lexical Overlap vs.\ Reasoning.}
To verify that \benchmarknameshort{} requires \emph{planning retrieval trajectories}, 
we check if gold evidence can be retrieved based on question n-grams.
Unigram matching yields a median of $\sim$4k pages per query with only 0.03\% precision---questions are drowned in false positives.
Bigram matching reduces hits to $\sim$24 pages but precision remains at 2.6\%.
Trigram matching achieves $\sim$1 page hit, yet recall drops to 51\%. 
This confirms that solving our benchmark requires semantic understanding, not just lexical overlap (Appendix~\ref{app:lexical-solvability}).

\paragraph{Parametric Knowledge vs.\ Grounding.}
To confirm that the benchmark measures the ability to \emph{synthesize a faithful answer} solely from $\mathcal{C}$, we prompt six frontier models to guess answers based solely on question text. 
Then, we categorize correct guesses into three types: yes/no questions, other binary-choice questions, and memorization (facts recalled from the training data). Across models, measured guessability ranges from 9.1\% (Claude Haiku) to 15.2\% (GPT-5), with an average of 11.2\%. Based on our question classification, 3\% stems from random chance on yes/no and binary questions, while the remaining 8\% reflects training data contamination---models correctly recalling facts from public documents. 

When models achieve 80\%+ accuracy with document evidence, the additional 70+ percentage points represent genuine comprehension.
See Appendix~\ref{app:guessability} for detailed results. 

\begin{figure}[h]
    \centering
    \includegraphics[width=\linewidth]{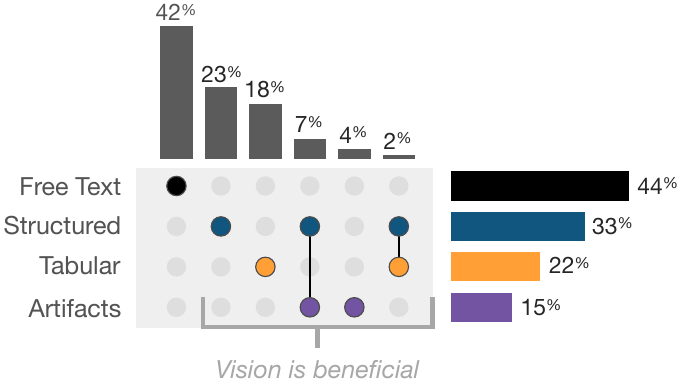}
    \caption{\textbf{Visual necessity in \benchmarknameshort{}.} 58\% of the questions benefit from understanding \textit{Structured} layouts, \textit{Tabular} data, or Visual \textit{Artifacts} (e.g., charts, stamps). The matrix highlights that multi-category dependencies (e.g., \textit{Structured} + \textit{Artifacts}) are a significant driver of benchmark difficulty.}
    \label{fig:visual_necessity}
\end{figure}

\paragraph{Visual Perception.}
To quantify visual understanding (Property~6), we categorize questions by whether answers are extractable from running text, benefit from layout comprehension (forms, tables), or require visual artifacts (checkboxes, figures).

Figure~\ref{fig:visual_necessity} shows that only 42\% of questions can be answered from free text alone. Structure is not the only visual challenge: 15\% involve visual artifacts, with 7\% requiring both---e.g., interpreting a checkbox within a form field. While our documents are text-heavy, this text is rarely unstructured, and relationships between elements must be taken into account. While tabular relationships can sometimes be inferred from linearized text, our fine-grained annotations enable measuring where visual encoders, layout-aware parsing, or pure text extraction each succeed.
See Appendix~\ref{sec:visual_modality_taxonomy} for the full taxonomy.



\subsection{Principled Splits Creation }\label{sec:subset}

\begin{figure}[h]
    \centering
    \includegraphics[width=\linewidth]{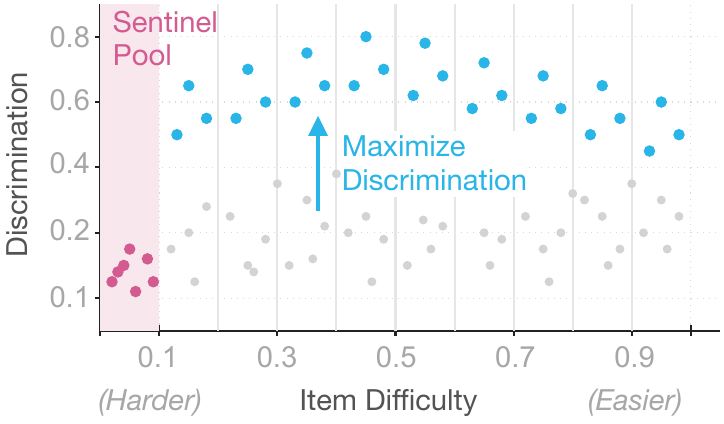}
    \caption{\textbf{Principled dev/test set selection.} We evaluate every question based on Difficulty (mean accuracy) and Discrimination (point-biserial correlation). The Sentinel Pool (\textcolor{snowpink}{$\bullet$}) captures the hardest items to preserve headroom, regardless of discrimination scores. For the remaining budget, we stratify questions into difficulty bins and greedily select those with the highest discrimination signal (\textcolor{snowlight}{$\bullet$}), discarding questions with lower predictive power (\textcolor{gray}{$\bullet$}). Data on the plot are illustrative.}
    \label{fig:ctt}
\end{figure}

To reduce evaluation costs while maintaining statistical power, we apply \textit{Classical Test Theory} \cite{crocker1986introduction}.
We evaluate each questions's Difficulty and Discrimination, prioritizing questions that best distinguish between strong and weak models (Figure~\ref{fig:ctt}, Appendix~\ref{app:subset-selection}).

Crucially, we explicitly reserve 20\% of the test set (the "Sentinel Pool") for items that no current model can solve. This design guarantees that the benchmark retains significant headroom (0\% baseline accuracy on this subset) even as models improve on the discriminatory set.

This approach yields non-overlapping Test ($n=\testsize$) and Development ($n=\devsize$) sets. The Test set achieves a strong rank correlation with the complete benchmark (Spearman's $\rho > 0.85$) while retaining 100 items that are too complex for current models to ensure long-term relevance. The remaining $\trainsize$ items are provided as a Train set.

%% file: sections/evaluation.tex

\section{Evaluation Protocol}\label{sec:evaluation}

\subsection{Answer Correctness (Property: \textit{Extractive})}\label{sec:accuracy}
We use LLM-based \textbf{Accuracy} to balance two needs: answers must be concrete values suitable for downstream automation, yet the metric should accept semantically correct responses even when they differ in surface form from ground truth.\footnote{We initially considered ANLS$^*$~\cite{peer2025anlsuniversaldocument}, but found it too strict---even after adding alternative answers, 35\% of predictions where ANLS$^*$ assigned zero score were actually correct per human review. This motivated our LLM-based judge.} By focusing on the \textit{Extractive} property, we ensure the benchmark rewards answers grounded in the corpus, not stylistic choices like formatting or verbosity.

Answers are represented as lists of strings to accommodate multi-part responses (e.g., ``list all board members''). The evaluation prompt was iteratively refined through human calibration rounds, addressing edge cases in list formatting, verbosity tolerance, and unit qualifier handling. Excluding $\sim$50\% exact match cases where  agreement is perfect by construction, the final setup achieves a quadratic-weighted Cohen's $\kappa = 0.88$ with human judgments, indicating almost perfect agreement~\cite{Landis77}.

Finally, to ensure statistical validity, we measured the calibrated judge's sensitivity and specificity on human-annotated samples, and apply bias correction to aggregate scores following \citet{lee2025llmjudge} (see Appendix~\ref{app:anls_star}).

\subsection{Retrieval and Attribution (Property: \textit{Grounded})}
As our questions are by design unanswerable using the general knowledge, we chose the \textbf{\pagef{}} metric (Appendix~\ref{app:retrieval_metrics}) to serve as a proxy for the \textit{Context Relevance} component of the RAG Triad \cite{truera2024ragtriad}.

It measures the overlap between the unique set of pages cited by the agent and the human-annotated minimal evidence set $\mathcal{E}$. A high score certifies that the agent successfully navigated to the precise location of the answer, penalizing both ``lazy citations'' (failure to cite necessary pages) and ``spurious citations'' (citing irrelevant pages).

We also report \textbf{\docf{}}, which relaxes the constraint to the document level. Comparing \docf{} against \pagef{} allows us to diagnose ``last-mile'' navigation failures, where an agent correctly identifies the relevant document but fails to pinpoint the specific page containing the evidence. We annotate at page level rather than bounding-box level; Appendix~\ref{sec:bbox_rationale} discusses this design choice.

\subsection{Efficiency and Calibration (Property: \textit{Agentic})}\label{sec:kuiper}

The defining feature of the Agentic property is the ability to invest variable compute into a problem. To measure whether this investment is rational, we design a metric based on the Cumulative Difference method \cite{kloumann2024cumulativedifferencespairedsamples},
treating the number of discrete steps (e.g., search iterations) as a proxy for implicit uncertainty. 

Specifically, given evaluation tuples $\{(s_i,y_i)\}_{i=1}^N$, corresponding to different test set items, with effort $s_i\in\mathbb{N}$ and correctness $y_i\in\{0,1\}$, we sort by nondecreasing effort via permutation $\pi$ and define mean accuracy $\bar y=\frac1N\sum_{i=1}^N y_i$.
The cumulative deviation curve:
\[
D_0=0, \quad D_k=\sum_{j=1}^k \bigl(y_{\pi(j)}-\bar y\bigr)
\]
\looseness=-1 tracks how accuracy conditioned on increasing effort departs from the global mean: upward (downward) segments indicate effort ranges with above-mean (below-mean) accuracy.

\begin{figure}[h]
    \centering
    \includegraphics[width=0.92\linewidth,trim={0 0 0 0.2cm},clip]{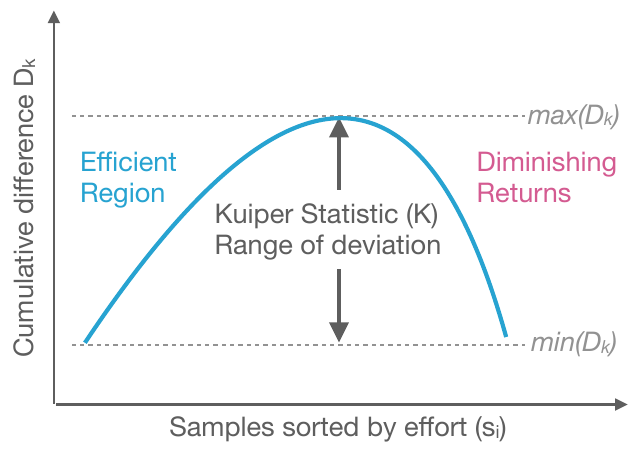}
    \caption{\textbf{Illustration of the cumulative difference curve.} Upward segments indicate above-average accuracy at low effort, while downward indicate below-average accuracy at higher effort. The Kuiper statistic measures the total range of this deviation.}
    \label{fig:kuipe}
\end{figure}

We quantify the dependency between effort and accuracy using the \textbf{\kuiper{}} range statistic (Figure~\ref{fig:kuipe}, Appendix~\ref{app:calibration_metrics}):
\[
K=\max_{0\le k\le N} D_k \;-\; \min_{0\le k\le N} D_k
\]

A low \kuiper{} score indicates stable, ``effort-invariant'' performance. A high score reveals poor calibration, specifically identifying regimes where the agent is expending significant budget on complex queries it ultimately fails to solve. In practice, we interpret Kuiper jointly with accuracy and grounding: low values indicate weak dependence on effort, which can reflect either well-calibrated termination or uniformly effort-invariant behavior.

The choice of effort measure has limited impact on calibration scores: step counts and token-based measures correlate strongly (Spearman $\rho > 0.85$), with Kuiper values varying by less than 20\% across definitions (Appendix~\ref{app:effort-sensitivity}). We report step counts (tool calls) by default.

%% file: sections/baselines.tex

\input{tables/main_table}

\section{Baseline Approches}\label{sec:baselines}

\paragraph{BM25 MLLM Agent {\normalfont{(Textual Retrieval, Visual Reasoning).}}} We propose an iterative system that couples text-based retrieval provided as a search tool with a MLLM, allowing the agent to formulate search queries and analyze rendered page images to answer questions (Appendix~\ref{appendix:search-agent-baseline}).

\paragraph{Claude Agent with Semtools {\normalfont{(CLI-Based Semantic Search).}}}
We propose a solution based on Claude Agents SDK, which operates through Unix-style tools such as \textit{parse} (converts PDFs to MD) and  
\textit{search} (performs semantic search), combined with regular bash commands, enabling flexible query reformulation 
and document exploration (Appendix~\ref{app:semtools}). 

\paragraph{Recursive Language Models {\normalfont{(Programmatic Context Decomposition).}}} A task-agnostic approach which enables LLMs to programmatically examine and recursively process the input \cite{zhang2025recursivelanguagemodels}. The document collection is exposed as a variable, allowing the LLM to write code processing it with sub-LLM calls (Appendix~\ref{app:rlm}).

\paragraph{MDocAgent {\normalfont{(Collaborative Agents).}}} 
Integrates textual and visual cues through a fixed five-stage pipeline of specialized agents. By employing parallel text and image retrieval, it coordinates General, Critical, Text, and Image agents to extract and analyze key information, which is then consolidated by a Summarizing agent \cite{han2025mdocagentmultimodalmultiagentframework}.

\paragraph{Managed RAG Services {\normalfont{(Blackbox Retrieval).}}}
An industry-standard reference point, relying on ``RAG-as-a-Service'' solutions. 
These include \textit{Gemini File Search} and \textit{OpenAI Assistants File Search} described in Appendix~\ref{app:managed}.

\paragraph{M3DocRAG {\normalfont{(Visual Retrieval).}}} Proposed by \citet{cho2024m3docragmultimodalretrievalneed} system with vision-aware retriever that encodes document pages as images, allowing it to capture visual cues (e.g., charts, layout). Results are then fed into a MLLM.

\paragraph{HEAVEN {\normalfont{(Hybrid Visual Retrieval).}}} Multi-vector retrieval proposed by \citet{kim2025hybrid}. Uses DSE~\cite{ma2024unifyingmultimodalretrievaldocument} to efficiently retrieve candidate pages, and then re-ranks candidates using ColQwen2.5~\cite{faysse2024colpaliefficientdocumentretrieval}. 
Retrieved pages are passed to a MLLM (see Appendix~\ref{app:heaven}).

\paragraph{ColBERTv2 + LLaMA {\normalfont{(Text-Only Late Interaction).}}}
A simple open-source baseline with late-interaction retrieval \cite{santhanam2022colbertv2effectiveefficientretrieval}. Relevant pages are retrieved and fed to Llama 3 8B model, following \citet{han2025mdocagentmultimodalmultiagentframework} setup.

\paragraph{Human Performance}.\label{sec:human}
To contextualize model performance, we collected human baselines annotations on all test questions (full interface details and annotator instructions are provided in Appendix~\ref{app:human-baseline}).

In the first setup, annotators used the same search engine as our \textit{BM25 MLLM Agent} baseline. The interface logged complete interaction trajectories: every search query, page view, and timestamp. This enables direct comparison not only of accuracy but also of search efficiency between humans and LLM agents. 

Additionally, we tested humans with oracle retrieval, eliminating the effect of an imperfect search tool on results, and reducing task cognitive load.

%% file: tables/main_table.tex
\begin{table*}[t!]
    \caption{\textbf{Main evaluation results on \benchmarknameshort{}.} Agentic systems consistently outperform their static RAG counterparts, yet an 18\% oracle gap reveals that retrieval---not reasoning---remains the primary bottleneck. We report aggregate Accuracy ($\pm$ confidence intervals) alongside specific multi-hop reasoning subsets (X-Page and X-Doc). Attribution is measured via Page F1 and Doc F1 to assess grounding fidelity. The Kuiper statistic ($\downarrow$ is better) quantifies effort calibration; it is excluded for Non-Agentic systems as they operate with fixed computational budgets.\\\textit{\colorbox[HTML]{E8A0BF}{Pink} indicates higher Kuiper values (worse; $\downarrow$ better). \colorbox[HTML]{84C6EB}{Blue} indicates higher performance (better).} \textbf{Bold} denotes the best performing model in Agentic and Non-Agentic systems for each subset.}
    \label{tab:main_leaderboard}
    \centering
    \small
    \setlength{\tabcolsep}{7pt}
    \renewcommand{\arraystretch}{1.1}
    \setlength{\aboverulesep}{0pt}
    \setlength{\belowrulesep}{0pt}
    \begin{tabular}{l c c c c c c}
    \toprule
    \noalign{\vspace{2pt}}
    \textbf{Model / Framework} & \textbf{Accuracy} & \textbf{X-Page} & \textbf{X-Doc} & \textbf{Page F1} & \textbf{Doc F1} & \textbf{Kuiper} $\downarrow$ \\[2pt]
    \midrule
    \noalign{\vspace{2pt}}
    \multicolumn{7}{l}{\textit{\textbf{Non-Agentic Systems}}} \\[2pt]
    \midrule
    Gemini 3 Pro \textsubscript{File Search} & \cellcolor[HTML]{84C6EB} \textbf{78.6 \textcolor{snowdark}{\scriptsize $\pm$ 2.2}} & \cellcolor[HTML]{86C7EB} \textbf{74.1 \textcolor{snowdark}{\scriptsize $\pm$ 3.6}} & \cellcolor[HTML]{83C6EA} \textbf{75.0 \textcolor{snowdark}{\scriptsize $\pm$ 3.6}} & \cellcolor[HTML]{82C6EA} \textbf{70.1 \textcolor{snowdark}{\scriptsize $\pm$ 2.0}} & \cellcolor[HTML]{5FB5E5} \textbf{94.2 \textcolor{snowdark}{\scriptsize $\pm$ 1.0}} & -- \\
    Gemini 2.5 Flash \textsubscript{File Search} & \cellcolor[HTML]{92CDED} 71.8 \textcolor{snowdark}{\scriptsize $\pm$ 2.4} & \cellcolor[HTML]{9DD2EF} 61.3 \textcolor{snowdark}{\scriptsize $\pm$ 4.1} & \cellcolor[HTML]{87C8EB} 73.0 \textcolor{snowdark}{\scriptsize $\pm$ 3.7} & \cellcolor[HTML]{AAD8F1} 52.2 \textcolor{snowdark}{\scriptsize $\pm$ 2.2} & \cellcolor[HTML]{7AC2E9} 80.9 \textcolor{snowdark}{\scriptsize $\pm$ 1.8} & -- \\
    M3DocRAG & \cellcolor[HTML]{A7D6F0} 61.6 \textcolor{snowdark}{\scriptsize $\pm$ 2.6} & \cellcolor[HTML]{D2EAF7} 31.0 \textcolor{snowdark}{\scriptsize $\pm$ 3.9} & \cellcolor[HTML]{CEE8F7} 35.0 \textcolor{snowdark}{\scriptsize $\pm$ 4.0} & \cellcolor[HTML]{86C7EB} 68.2 \textcolor{snowdark}{\scriptsize $\pm$ 2.1} & \cellcolor[HTML]{77C0E8} 82.6 \textcolor{snowdark}{\scriptsize $\pm$ 1.7} & -- \\
    GPT-5.2 (2024-08) \textsubscript{HEAVEN} & \cellcolor[HTML]{B9DFF3} 52.9 \textcolor{snowdark}{\scriptsize $\pm$ 2.7} & \cellcolor[HTML]{C5E4F5} 38.9 \textcolor{snowdark}{\scriptsize $\pm$ 4.1} & \cellcolor[HTML]{ACD9F1} 53.0 \textcolor{snowdark}{\scriptsize $\pm$ 4.2} & \cellcolor[HTML]{B3DCF2} 48.4 \textcolor{snowdark}{\scriptsize $\pm$ 2.2} & \cellcolor[HTML]{A0D3EF} 62.3 \textcolor{snowdark}{\scriptsize $\pm$ 2.2} & -- \\
    GPT-5.2 (2025-12) \textsubscript{File Search} & \cellcolor[HTML]{BFE1F4} 50.0 \textcolor{snowdark}{\scriptsize $\pm$ 2.7} & \cellcolor[HTML]{C3E3F5} 39.5 \textcolor{snowdark}{\scriptsize $\pm$ 4.1} & \cellcolor[HTML]{E4F3FA} 23.0 \textcolor{snowdark}{\scriptsize $\pm$ 3.5} & \cellcolor[HTML]{E0F0F9} 28.5 \textcolor{snowdark}{\scriptsize $\pm$ 2.0} & \cellcolor[HTML]{94CEED} 68.5 \textcolor{snowdark}{\scriptsize $\pm$ 2.1} & -- \\
    GPT-5 (2025-08) \textsubscript{File Search} & \cellcolor[HTML]{C0E2F4} 49.6 \textcolor{snowdark}{\scriptsize $\pm$ 2.7} & \cellcolor[HTML]{C9E6F6} 36.4 \textcolor{snowdark}{\scriptsize $\pm$ 4.0} & \cellcolor[HTML]{E1F1FA} 25.0 \textcolor{snowdark}{\scriptsize $\pm$ 3.6} & \cellcolor[HTML]{DEEFF9} 29.3 \textcolor{snowdark}{\scriptsize $\pm$ 2.0} & \cellcolor[HTML]{98CFEE} 66.6 \textcolor{snowdark}{\scriptsize $\pm$ 2.1} & -- \\
    GPT-4o (2024-08) \textsubscript{HEAVEN} & \cellcolor[HTML]{C2E3F5} 48.6 \textcolor{snowdark}{\scriptsize $\pm$ 2.7} & \cellcolor[HTML]{D0E9F7} 32.2 \textcolor{snowdark}{\scriptsize $\pm$ 3.9} & \cellcolor[HTML]{CAE7F6} 37.0 \textcolor{snowdark}{\scriptsize $\pm$ 4.0} & \cellcolor[HTML]{BEE1F4} 43.2 \textcolor{snowdark}{\scriptsize $\pm$ 2.2} & \cellcolor[HTML]{A7D6F0} 59.2 \textcolor{snowdark}{\scriptsize $\pm$ 2.2} & -- \\
    GPT-5 Mini (2025-08) \textsubscript{File Search} & \cellcolor[HTML]{C2E3F5} 48.5 \textcolor{snowdark}{\scriptsize $\pm$ 2.7} & \cellcolor[HTML]{CFE9F7} 32.8 \textcolor{snowdark}{\scriptsize $\pm$ 3.9} & \cellcolor[HTML]{DFF0F9} 26.0 \textcolor{snowdark}{\scriptsize $\pm$ 3.7} & \cellcolor[HTML]{DEF0F9} 29.0 \textcolor{snowdark}{\scriptsize $\pm$ 2.0} & \cellcolor[HTML]{96CFEE} 67.3 \textcolor{snowdark}{\scriptsize $\pm$ 2.1} & -- \\
    ColBERTv2 + Llama-3.1-8B & \cellcolor[HTML]{D3EBF8} 40.2 \textcolor{snowdark}{\scriptsize $\pm$ 2.6} & \cellcolor[HTML]{DFF0F9} 23.7 \textcolor{snowdark}{\scriptsize $\pm$ 3.5} & \cellcolor[HTML]{DFF0F9} 26.0 \textcolor{snowdark}{\scriptsize $\pm$ 3.7} & \cellcolor[HTML]{BEE1F4} 43.4 \textcolor{snowdark}{\scriptsize $\pm$ 2.2} & \cellcolor[HTML]{B6DDF3} 52.0 \textcolor{snowdark}{\scriptsize $\pm$ 2.2} & -- \\
    \midrule
    \noalign{\vspace{2pt}}
    \multicolumn{7}{l}{\textit{\textbf{Agentic Systems}}} \\[2pt]
    \midrule
    Gemini 3 Pro \textsubscript{BM25 Agent} & \cellcolor[HTML]{7CC3E9} \textbf{82.2 \textcolor{snowdark}{\scriptsize $\pm$ 2.0}} & \cellcolor[HTML]{93CDED} \textbf{66.8 \textcolor{snowdark}{\scriptsize $\pm$ 3.9}} & \cellcolor[HTML]{87C8EB} 73.0 \textcolor{snowdark}{\scriptsize $\pm$ 3.7} & \cellcolor[HTML]{6FBDE7} 78.5 \textcolor{snowdark}{\scriptsize $\pm$ 1.8} & \cellcolor[HTML]{67B9E6} 90.2 \textcolor{snowdark}{\scriptsize $\pm$ 1.3} & \cellcolor[HTML]{F7E1EB} 25.8 \\
    Claude Sonnet 4.5 (2025-09) \textsubscript{BM25 Agent} & \cellcolor[HTML]{80C4EA} 80.6 \textcolor{snowdark}{\scriptsize $\pm$ 2.1} & \cellcolor[HTML]{93CDED} \textbf{66.8 \textcolor{snowdark}{\scriptsize $\pm$ 3.9}} & \cellcolor[HTML]{76C0E8} \textbf{82.0 \textcolor{snowdark}{\scriptsize $\pm$ 3.2}} & \cellcolor[HTML]{6EBCE7} \textbf{79.1 \textcolor{snowdark}{\scriptsize $\pm$ 1.8}} & \cellcolor[HTML]{61B6E5} \textbf{93.0 \textcolor{snowdark}{\scriptsize $\pm$ 1.1}} & \cellcolor[HTML]{F4D4E2} 35.1 \\
    GPT-5 (2025-08) \textsubscript{BM25 Agent} & \cellcolor[HTML]{86C7EB} 77.7 \textcolor{snowdark}{\scriptsize $\pm$ 2.2} & \cellcolor[HTML]{9FD3EF} 60.1 \textcolor{snowdark}{\scriptsize $\pm$ 4.1} & \cellcolor[HTML]{85C7EB} 74.0 \textcolor{snowdark}{\scriptsize $\pm$ 3.7} & \cellcolor[HTML]{79C1E9} 74.2 \textcolor{snowdark}{\scriptsize $\pm$ 2.0} & \cellcolor[HTML]{6FBDE7} 86.5 \textcolor{snowdark}{\scriptsize $\pm$ 1.5} & \cellcolor[HTML]{EEBCD2} 52.6 \\
    Gemini 3 Pro \textsubscript{RLM} & \cellcolor[HTML]{8ECBEC} 73.8 \textcolor{snowdark}{\scriptsize $\pm$ 2.3} & \cellcolor[HTML]{93CDED} \textbf{66.8 \textcolor{snowdark}{\scriptsize $\pm$ 3.9}} & \cellcolor[HTML]{94CEED} 66.0 \textcolor{snowdark}{\scriptsize $\pm$ 3.9} & \cellcolor[HTML]{86C7EB} 69.1 \textcolor{snowdark}{\scriptsize $\pm$ 2.1} & \cellcolor[HTML]{67B9E6} 89.8 \textcolor{snowdark}{\scriptsize $\pm$ 1.4} & \cellcolor[HTML]{F8E5ED} \textbf{22.9} \\
    Claude Agent \textsubscript{Semtools} & \cellcolor[HTML]{90CCED} 72.6 \textcolor{snowdark}{\scriptsize $\pm$ 2.4} & \cellcolor[HTML]{9CD1EE} 62.0 \textcolor{snowdark}{\scriptsize $\pm$ 4.0} & \cellcolor[HTML]{9FD3EF} 60.0 \textcolor{snowdark}{\scriptsize $\pm$ 4.1} & \cellcolor[HTML]{ADD9F1} 51.1 \textcolor{snowdark}{\scriptsize $\pm$ 2.2} & \cellcolor[HTML]{68BAE6} 89.5 \textcolor{snowdark}{\scriptsize $\pm$ 1.4} & \cellcolor[HTML]{F3D0DF} 37.9 \\
    Claude 4.5 Sonnet (2025-09) \textsubscript{RLM} & \cellcolor[HTML]{95CEED} 70.5 \textcolor{snowdark}{\scriptsize $\pm$ 2.4} & \cellcolor[HTML]{96CFEE} 65.0 \textcolor{snowdark}{\scriptsize $\pm$ 4.0} & \cellcolor[HTML]{8FCBEC} 69.0 \textcolor{snowdark}{\scriptsize $\pm$ 3.9} & \cellcolor[HTML]{8BCAEC} 66.5 \textcolor{snowdark}{\scriptsize $\pm$ 2.1} & \cellcolor[HTML]{6BBBE7} 88.9 \textcolor{snowdark}{\scriptsize $\pm$ 1.5} & \cellcolor[HTML]{F2CADB} 42.3 \\
    Claude Haiku 4.5 (2025-10) \textsubscript{BM25 Agent} & \cellcolor[HTML]{99D0EE} 68.2 \textcolor{snowdark}{\scriptsize $\pm$ 2.5} & \cellcolor[HTML]{B4DDF2} 48.0 \textcolor{snowdark}{\scriptsize $\pm$ 4.2} & \cellcolor[HTML]{96CFEE} 65.0 \textcolor{snowdark}{\scriptsize $\pm$ 4.0} & \cellcolor[HTML]{7EC4EA} 72.0 \textcolor{snowdark}{\scriptsize $\pm$ 2.0} & \cellcolor[HTML]{6BBBE7} 88.2 \textcolor{snowdark}{\scriptsize $\pm$ 1.4} & \cellcolor[HTML]{EFBFD4} 50.7 \\
    GPT-5.2 (2025-12) \textsubscript{BM25 Agent} & \cellcolor[HTML]{9AD1EE} 67.8 \textcolor{snowdark}{\scriptsize $\pm$ 2.5} & \cellcolor[HTML]{AEDAF1} 51.6 \textcolor{snowdark}{\scriptsize $\pm$ 4.2} & \cellcolor[HTML]{A9D7F1} 55.0 \textcolor{snowdark}{\scriptsize $\pm$ 4.1} & \cellcolor[HTML]{88C8EB} 67.6 \textcolor{snowdark}{\scriptsize $\pm$ 2.1} & \cellcolor[HTML]{74BFE8} 83.7 \textcolor{snowdark}{\scriptsize $\pm$ 1.7} & \cellcolor[HTML]{EAABC6} 64.8 \\
    GPT-5 Mini (2025-08) \textsubscript{BM25 Agent} & \cellcolor[HTML]{9CD1EE} 66.9 \textcolor{snowdark}{\scriptsize $\pm$ 2.5} & \cellcolor[HTML]{B4DDF2} 48.0 \textcolor{snowdark}{\scriptsize $\pm$ 4.2} & \cellcolor[HTML]{B6DDF3} 48.0 \textcolor{snowdark}{\scriptsize $\pm$ 4.2} & \cellcolor[HTML]{88C8EB} 67.6 \textcolor{snowdark}{\scriptsize $\pm$ 2.1} & \cellcolor[HTML]{77C0E8} 82.4 \textcolor{snowdark}{\scriptsize $\pm$ 1.7} & \cellcolor[HTML]{E8A0BF} 73.2 \\
    GLM-4.6V \textsubscript{BM25 Agent} & \cellcolor[HTML]{9ED2EF} 66.1 \textcolor{snowdark}{\scriptsize $\pm$ 2.5} & \cellcolor[HTML]{C8E5F6} 37.1 \textcolor{snowdark}{\scriptsize $\pm$ 4.0} & \cellcolor[HTML]{8DCAEC} 70.0 \textcolor{snowdark}{\scriptsize $\pm$ 3.8} & \cellcolor[HTML]{8BCAEC} 65.9 \textcolor{snowdark}{\scriptsize $\pm$ 2.1} & \cellcolor[HTML]{6EBCE7} 86.6 \textcolor{snowdark}{\scriptsize $\pm$ 1.5} & \cellcolor[HTML]{EFBED3} 51.4 \\
    GPT-5.2 (2025-12) \textsubscript{RLM} & \cellcolor[HTML]{A2D4EF} 64.2 \textcolor{snowdark}{\scriptsize $\pm$ 2.6} & \cellcolor[HTML]{A8D7F0} 55.3 \textcolor{snowdark}{\scriptsize $\pm$ 4.1} & \cellcolor[HTML]{A7D6F0} 56.0 \textcolor{snowdark}{\scriptsize $\pm$ 4.1} & \cellcolor[HTML]{88C8EB} 67.6 \textcolor{snowdark}{\scriptsize $\pm$ 2.1} & \cellcolor[HTML]{74BFE8} 83.7 \textcolor{snowdark}{\scriptsize $\pm$ 1.7} & \cellcolor[HTML]{F6DBE7} 30.0 \\
    MDocAgent & \cellcolor[HTML]{A2D4F0} 63.8 \textcolor{snowdark}{\scriptsize $\pm$ 2.6} & \cellcolor[HTML]{C8E5F6} 37.1 \textcolor{snowdark}{\scriptsize $\pm$ 4.0} & \cellcolor[HTML]{C3E3F5} 41.0 \textcolor{snowdark}{\scriptsize $\pm$ 4.1} & \cellcolor[HTML]{89C9EB} 67.1 \textcolor{snowdark}{\scriptsize $\pm$ 2.1} & \cellcolor[HTML]{78C1E9} 82.1 \textcolor{snowdark}{\scriptsize $\pm$ 1.7} & \cellcolor[HTML]{F7DFEA} 27.1 \\
    Qwen3-VL (235B-A22B-Thinking) \textsubscript{BM25 Agent} & \cellcolor[HTML]{AAD8F1} 60.3 \textcolor{snowdark}{\scriptsize $\pm$ 2.6} & \cellcolor[HTML]{C9E6F6} 36.4 \textcolor{snowdark}{\scriptsize $\pm$ 4.0} & \cellcolor[HTML]{A9D7F1} 55.0 \textcolor{snowdark}{\scriptsize $\pm$ 4.1} & \cellcolor[HTML]{9CD1EE} 58.6 \textcolor{snowdark}{\scriptsize $\pm$ 2.2} & \cellcolor[HTML]{7BC2E9} 80.5 \textcolor{snowdark}{\scriptsize $\pm$ 1.8} & \cellcolor[HTML]{F4D2E1} 36.6 \\
    GPT-4.1 (2025-04) \textsubscript{BM25 Agent} & \cellcolor[HTML]{AAD8F1} 60.0 \textcolor{snowdark}{\scriptsize $\pm$ 2.6} & \cellcolor[HTML]{BEE1F4} 42.5 \textcolor{snowdark}{\scriptsize $\pm$ 4.1} & \cellcolor[HTML]{BDE1F4} 44.0 \textcolor{snowdark}{\scriptsize $\pm$ 4.1} & \cellcolor[HTML]{90CCEC} 64.1 \textcolor{snowdark}{\scriptsize $\pm$ 2.1} & \cellcolor[HTML]{76C0E8} 82.8 \textcolor{snowdark}{\scriptsize $\pm$ 1.7} & \cellcolor[HTML]{F2C9DB} 43.2 \\
    Gemini 2.5 Flash \textsubscript{BM25 Agent} & \cellcolor[HTML]{ADD9F1} 58.5 \textcolor{snowdark}{\scriptsize $\pm$ 2.6} & \cellcolor[HTML]{D4EBF8} 30.4 \textcolor{snowdark}{\scriptsize $\pm$ 3.8} & \cellcolor[HTML]{A7D6F0} 56.0 \textcolor{snowdark}{\scriptsize $\pm$ 4.1} & \cellcolor[HTML]{96CFEE} 61.0 \textcolor{snowdark}{\scriptsize $\pm$ 2.2} & \cellcolor[HTML]{7EC4EA} 78.8 \textcolor{snowdark}{\scriptsize $\pm$ 1.8} & \cellcolor[HTML]{F0C5D7} 46.5 \\
    GPT-5 Nano (2025-08) \textsubscript{BM25 Agent} & \cellcolor[HTML]{AEDAF1} 58.2 \textcolor{snowdark}{\scriptsize $\pm$ 2.6} & \cellcolor[HTML]{CFE9F7} 32.8 \textcolor{snowdark}{\scriptsize $\pm$ 3.9} & \cellcolor[HTML]{CEE8F7} 35.0 \textcolor{snowdark}{\scriptsize $\pm$ 4.0} & \cellcolor[HTML]{97CFEE} 60.9 \textcolor{snowdark}{\scriptsize $\pm$ 2.2} & \cellcolor[HTML]{77C1E9} 82.2 \textcolor{snowdark}{\scriptsize $\pm$ 1.7} & \cellcolor[HTML]{EFC0D4} 49.8 \\
    Qwen3-VL (8B-Thinking) \textsubscript{BM25 Agent} & \cellcolor[HTML]{C5E4F5} 47.3 \textcolor{snowdark}{\scriptsize $\pm$ 2.7} & \cellcolor[HTML]{EBF6FB} 17.0 \textcolor{snowdark}{\scriptsize $\pm$ 3.1} & \cellcolor[HTML]{BDE1F4} 44.0 \textcolor{snowdark}{\scriptsize $\pm$ 4.1} & \cellcolor[HTML]{B5DDF2} 47.6 \textcolor{snowdark}{\scriptsize $\pm$ 2.2} & \cellcolor[HTML]{92CDED} 69.4 \textcolor{snowdark}{\scriptsize $\pm$ 2.1} & \cellcolor[HTML]{EFBFD4} 50.2 \\
    GLM-4.6V Flash \textsubscript{BM25 Agent} & \cellcolor[HTML]{C7E5F6} 46.0 \textcolor{snowdark}{\scriptsize $\pm$ 2.7} & \cellcolor[HTML]{E9F5FB} 18.2 \textcolor{snowdark}{\scriptsize $\pm$ 3.2} & \cellcolor[HTML]{B8DEF3} 47.0 \textcolor{snowdark}{\scriptsize $\pm$ 4.2} & \cellcolor[HTML]{DFF0F9} 28.9 \textcolor{snowdark}{\scriptsize $\pm$ 2.0} & \cellcolor[HTML]{B7DEF3} 51.5 \textcolor{snowdark}{\scriptsize $\pm$ 2.2} & \cellcolor[HTML]{F7DFE9} 27.5 \\
    GPT-4.1 Nano (2025-04) \textsubscript{BM25 Agent} & \cellcolor[HTML]{FFFFFF} 19.5 \textcolor{snowdark}{\scriptsize $\pm$ 2.1} & \cellcolor[HTML]{FFFFFF} 6.1 \textcolor{snowdark}{\scriptsize $\pm$ 2.0} & \cellcolor[HTML]{FFFFFF} 9.0 \textcolor{snowdark}{\scriptsize $\pm$ 2.4} & \cellcolor[HTML]{E2F1FA} 27.6 \textcolor{snowdark}{\scriptsize $\pm$ 2.0} & \cellcolor[HTML]{CEE8F7} 40.2 \textcolor{snowdark}{\scriptsize $\pm$ 2.2} & \cellcolor[HTML]{F6DDE8} 28.6 \\
    \midrule
    \noalign{\vspace{2pt}}
    \multicolumn{7}{l}{\textit{\textbf{Human Performance}}} \\[2pt]
    \midrule
    Human \textsubscript{Oracle Retriever} & \cellcolor[HTML]{59B3E4} \textbf{99.4 \textcolor{snowdark}{\scriptsize $\pm$ 0.4}} & \cellcolor[HTML]{59B3E4} \textbf{100.0} & \cellcolor[HTML]{59B3E4} \textbf{98.0 \textcolor{snowdark}{\scriptsize $\pm$ 1.2}} & -- & -- & -- \\
    Human \textsubscript{BM25 Agent} & \cellcolor[HTML]{7CC3E9} 82.2 \textcolor{snowdark}{\scriptsize $\pm$ 2.0} & \cellcolor[HTML]{7DC3E9} 79.6 \textcolor{snowdark}{\scriptsize $\pm$ 3.4} & \cellcolor[HTML]{89C9EB} 72.0 \textcolor{snowdark}{\scriptsize $\pm$ 3.7} & \cellcolor[HTML]{6DBCE7} 79.3 \textcolor{snowdark}{\scriptsize $\pm$ 1.8} & \cellcolor[HTML]{60B6E5} 93.4 \textcolor{snowdark}{\scriptsize $\pm$ 1.1} & \cellcolor[HTML]{FAF0F5} \textbf{14.6} \\
    \bottomrule
    \end{tabular}
    \renewcommand{\arraystretch}{1.0}
    \setlength{\aboverulesep}{0.4ex}
    \setlength{\belowrulesep}{0.65ex}
    \end{table*}

%% file: sections/experiments.tex

\section{Results and Analysis}\label{sec:experiments}


\paragraph{Simple Agentic Systems Can Outperform Strong, Static RAG.}
The best-performing system, \textit{Gemini 3 Pro BM25 MLLM Agent}, achieves an accuracy of 82.2\%, representing a substantial improvement over its optimized non-agentic counterpart, \textit{Gemini 3 Pro File Search} (78.6\%). This trend holds also for all models from GPT family, confirming that the iterative planning capability is beneficial in \benchmarknameshort{}. The only exception is Gemini 2.5 Flash, which performs exceptionally well with Google's managed RAG.

\paragraph{Specialized Solutions Punch Above Their Weight.}
In the class of 8B parameter backbones,  M3DocRAG and MDocAgent achieve accuracy above 60\%, rivaling larger commercial models, and  significantly outperforming \textit{ColBERTv2 + Llama~3} ($40.2\%$) and \textit{Qwen3-VL 8B BM25 MLLM Agent} ($47.3\%$). This performance gap highlights the potential of domain-specific innovations.

\begin{figure}[h]
    \centering
    \includegraphics[width=0.9\linewidth]{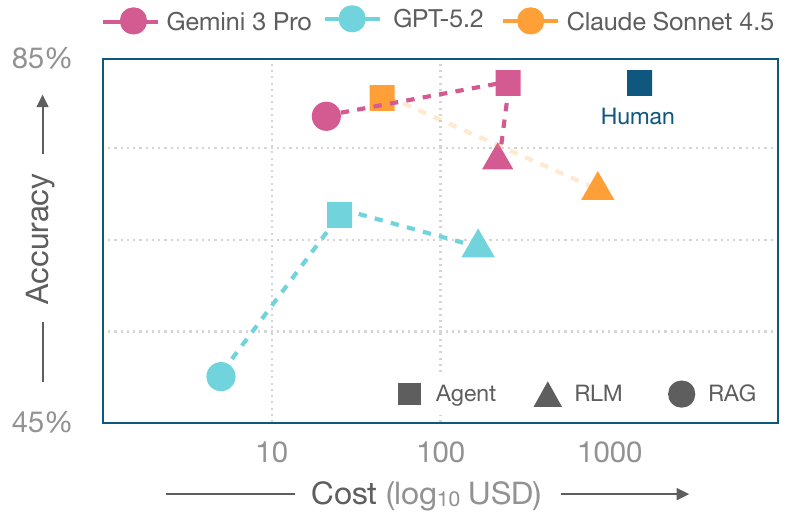}
    \caption{\textbf{Performance and cost of running test set inference.} Leading models with Managed RAG, compared to employing them in BM25 MLLM Agent or RLM setup.}
    \label{fig:accuracy_cost}
\end{figure}

\paragraph{Retrieval Constraints are Essential for Cost-Effective Reasoning.}
Incorporating inference cost in analysis (Figure~\ref{fig:accuracy_cost}) reveals that the constraints imposed by search tools and RAG pipelines are beneficial. While RLMs offer theoretical flexibility, their lack of constraints leads to inefficient information processing without performance gains. For example, the \textit{Claude Sonnet 4.5 RLM} processed over $270$ million input tokens---incurring a cost of $\$850$---yet failed to match the accuracy of its \textit{BM25 MLLM} counterpart.

\paragraph{Agents Achieve Superior Page-Level Attribution.}
While managed RAGs often achieve high \docf{}, they struggle with precise localization, as evidenced by lower \pagef{} scores. Agentic systems offer better page-level attribution. High \docf{} can mask a ``last-mile'' failure at the page level, justifying the diagnostic value of both.

\paragraph{Calibration is a Distinct Axis from Accuracy.}
Kuiper varies widely and is not monotonic in answer quality:
for example, \emph{GPT-5 BM25 Agent} reaches 77.7\% accuracy but exhibits substantially worse calibration
(Kuiper 52.6) than \emph{Gemini 3 Pro BM25 Agent} (82.2\%, Kuiper 25.8).

\begin{figure*}[t]
    \centering
    \includegraphics[width=\textwidth]{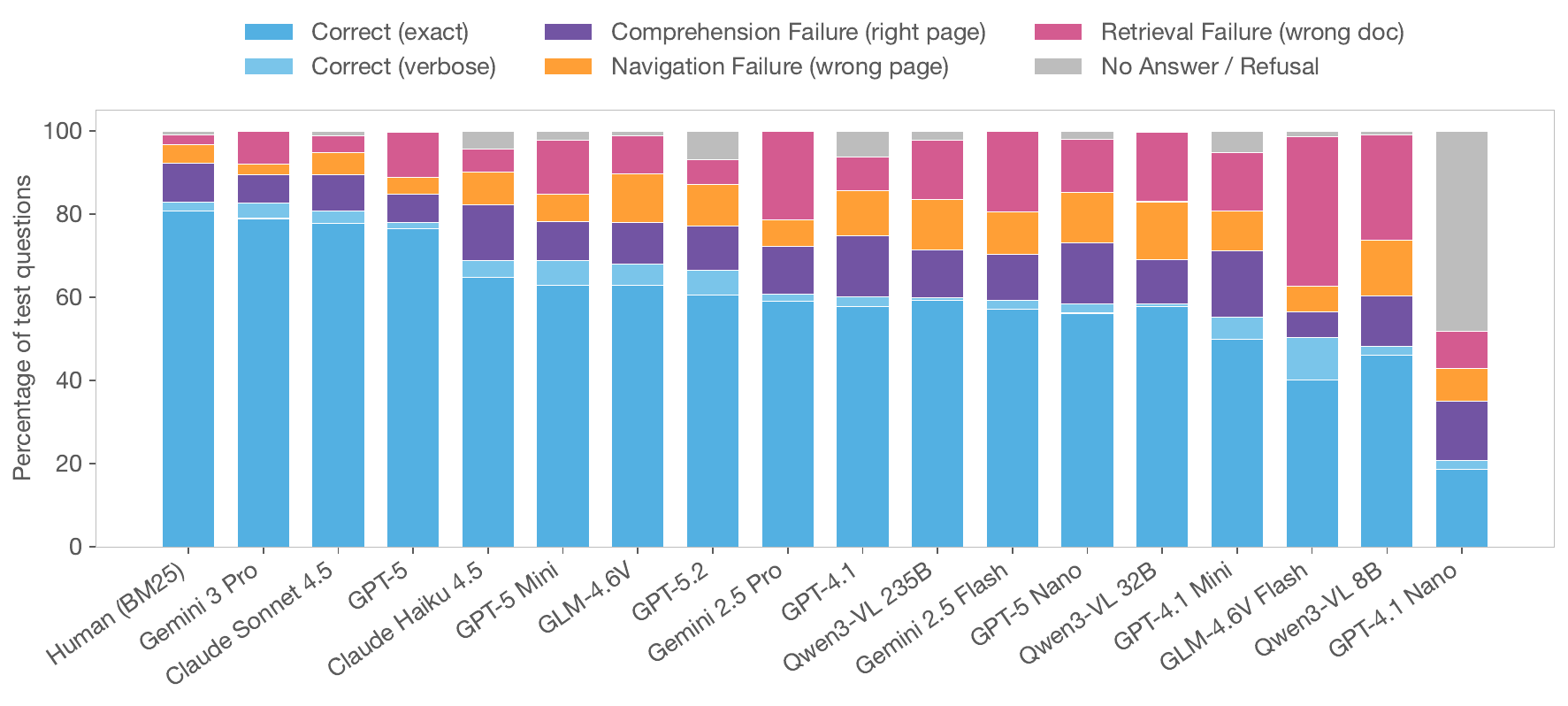}
    \caption{\textbf{Error decomposition across all BM25 MLLM agents}, ordered by accuracy. Each bar decomposes a system's test predictions into correct (exact and verbose) and four error types. Weaker models are dominated by refusals and retrieval failures, while stronger models shift toward comprehension errors---suggesting that retrieval is largely solved for top systems and answer extraction remains the bottleneck.}
    \label{fig:error-decomposition-main}
\end{figure*}

\subsection{Search Dynamics and Error Taxonomy}\label{sec:search-dynamics}

We analyze errors of the BM25 MLLM Agent family, which has the most models evaluated (Appendix~\ref{subsec:failure-analysis}).

\paragraph{Failure Modes Differ Qualitatively Across Models.} 
Among 3{,}273 agent errors, retrieval failures (wrong document) account for 35.7\%, followed by comprehension failures (right page, wrong answer, 28.8\%), navigation failures (right document, wrong page, 23.0\%), and refusals (12.6\%). However, these proportions vary drastically by model (Figure~\ref{fig:error-decomposition-main}): Gemini~2.5~Pro fails predominantly at retrieval (21.4\% of all predictions), while Claude~Sonnet~4.5 retrieves well (4.0\% retrieval failure) but struggles more with comprehension (8.6\%). GPT-4.1~Nano is dominated by refusals (48.2\%), indicating premature search abandonment. Detailed per-system profiles, including a retrieval-vs.-comprehension scatter (Figure~\ref{fig:ret-vs-comp}), are in Appendix~\ref{subsec:failure-analysis}.

\paragraph{Query Reformulation Magnitude Predicts Success.}
When an initial search fails, effective agents try a substantially \emph{different} query rather than a minor rephrasing.
We quantify this by embedding all search queries and measuring cosine drift between consecutive queries (Appendix~\ref{subsec:reformulation-analysis}). Top-performing systems reformulate more aggressively: Claude~Sonnet~4.5 has the highest mean drift per step (0.38), while GPT-4.1~Nano barely changes its queries (drift 0.10). Accuracy correlates strongly with reformulation magnitude.

\paragraph{Semantic Discontinuity, Not Physical Distance, Drives Failure.} 
Concerning multi-hop queries, the physical ``page gap'' between pieces of evidence is irrelevant to performance. Difficulty is determined by \emph{semantic distance}: accuracy drops by 38 percentage points when evidence spans conceptually dissimilar contexts (Appendix~\ref{app:multihop}).

\paragraph{Errors Reflect Genuine Misunderstanding.}
Only 5\% of correct predictions are over-verbose concerning our specification (Section~\ref{sec:accuracy}), indicating that when systems find the right evidence, they almost always provide the requested \textit{Extractive} answer we require. Among errors, 87.4\% involve a concrete retrieval, navigation, or comprehension failure---not hallucination or refusal---confirming that the benchmark primarily tests information-seeking ability.

\subsection{Human-Agent Comparative Analysis}\label{sec:human-agent}

\paragraph{Human-Agent Performance Gap.}
Despite the impressive capabilities of frontier models, humans with access to oracle retrieval perform at least 18\% better. 
The "Oracle Gap" reveals that nearly 18\% of the benchmark remains unsolved due to retrieval bottlenecks. The plateau at 80\% represents a limitation of current search capability, not a saturation of the benchmark's reasoning difficulty. This is consistent with the fact that, given the same BM25 search tool, humans and Gemini 3 Pro achieve comparable performance. 

\paragraph{Same Accuracy, Different Competencies.}
Although humans and Gemini~3~Pro both reach ${\sim}82\%$ accuracy, pairwise item agreement is remarkably low (Cohen's $\kappa=0.24$; Figure~\ref{fig:kappa-heatmap} in Appendix~\ref{subsec:item-agreement}): they succeed on \emph{different} questions. Of 107 disagreement items, 54 are solved only by humans and 53 only by the model. Human-specific failures are dominated by comprehension errors (64\%), reflecting attention fatigue on complex extractions, while model-specific failures split evenly between retrieval (43\%) and comprehension (43\%). In contrast, model--model pairs at similar accuracy levels agree substantially more ($\kappa \approx 0.43$), sharing systematic weaknesses. This complementarity suggests that hybrid human--agent pipelines could exceed the accuracy ceiling of either alone.

\begin{figure}[h!]
    \includegraphics[width=0.93\linewidth]{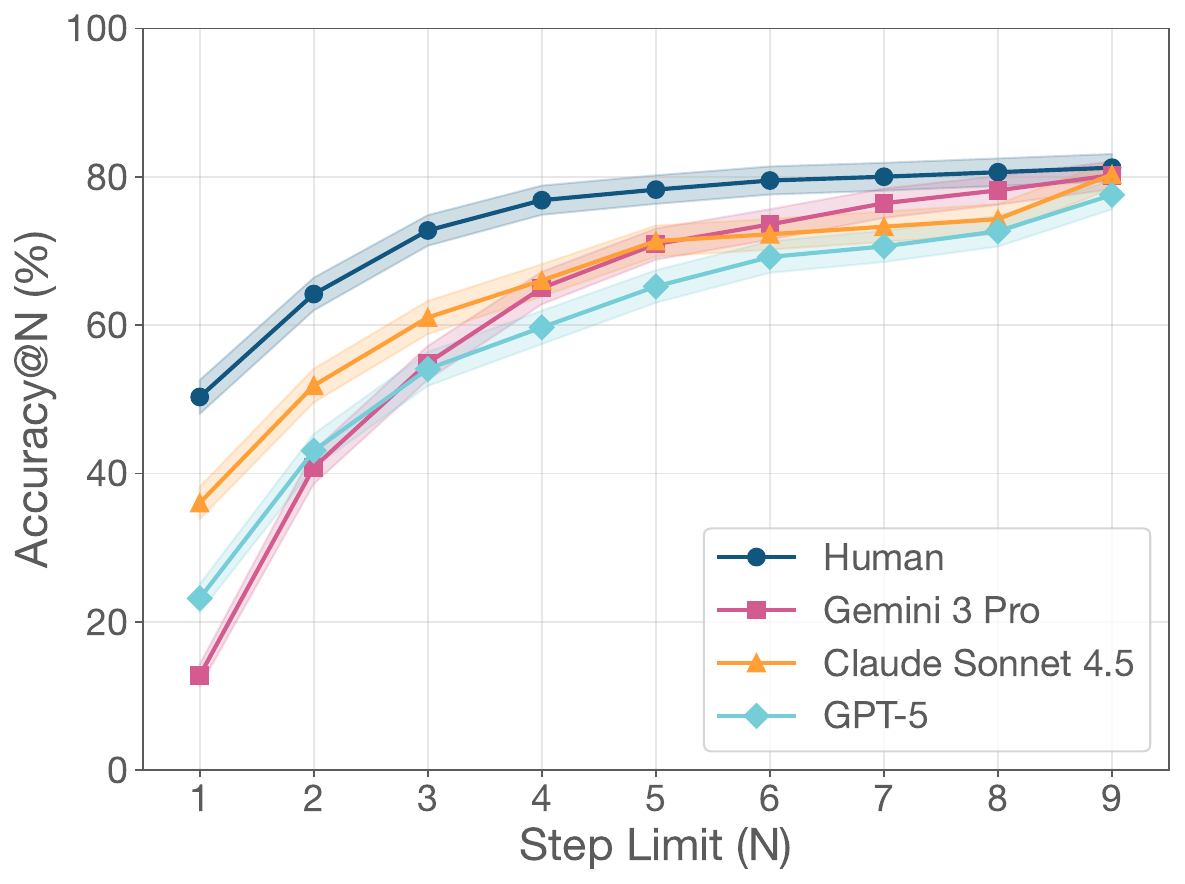}
    \caption{\textbf{Accuracy as a function of step limit ($N$) for BM25 MLLM agents.} While frontier agents eventually match human accuracy given a large budget ($N=9$), they suffer from a severe "Cold Start" efficiency gap. Humans achieve 50\% accuracy on their first query, whereas Gemini 3 Pro starts at only ~12\%, requiring an aggressive, high-cost recovery strategy to reach parity.}
    \label{fig:steps}
\end{figure}

\paragraph{Illusion of Infinite Budget.} Figure~\ref{fig:steps} plots the cumulative accuracy against the step limit $N$. While frontier agents eventually converge near the retrieval tool's theoretical ceiling ($\sim$80\%), they suffer from a severe ``Cold Start'' disparity. Human annotators demonstrate strong zero-shot strategic calibration, achieving $\sim$50\% accuracy on their very first query. In contrast, Gemini 3 Pro starts at only $\sim$12\%, relying on a steep, compute-intensive recovery. The ``Oracle Gap'' of 17.2\% combined with this cold-start inefficiency proves that MADQA remains unsolved for \textit{efficient} agents.

\paragraph{Perfect Retrieval Eliminates Human Reasoning Errors.}
Humans occasionally succumbed to ``negation blindness,'' temporal confusion, or role conflation (e.g., mistaking a signer for an approver). Detailed examples of these genuine human mistakes, distinct from annotation noise, are provided in Appendix~\ref{app:human_errors}. Interestingly, these problems almost disappear entirely with perfect retrieval---when the annotator is focusing on the right place in the document. 

\paragraph{Humans Calibrate Effort Better.}\looseness=-1
Humans achieve a \kuiper{} statistic of 14.6---below every agent system, whose scores range from 22.9 (Gemini 3 Pro RLM) to 73.2 (GPT-5 Mini BM25 Agent).
When initial queries fail, humans quickly change strategy. Agents, in contrast, often persist through minor query reformulations and spend compute on problems they ultimately fail to solve.


\begin{figure}[h]
    \centering
    \includegraphics[width=0.9\linewidth]{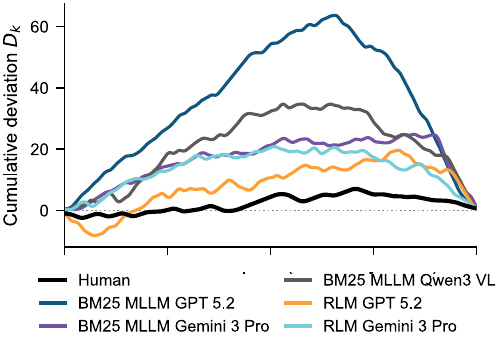}
    \caption{\textbf{Kuiper calibration curves} measuring effort-accuracy alignment. Human annotators remain well-calibrated across effort levels. RLM agents exhibit calibration closer to humans.}
    \label{fig:kuiper_curves}
\end{figure}

\paragraph{Response Time Inversely Correlates with Accuracy.} Humans answered questions in a median time of 2 minutes (mean 3.3 minutes), with 50\% of responses falling between 1--4 minutes. Response time inversely relates to accuracy: questions answered in under one minute achieve 86\%, while those requiring over ten minutes---68\%.


%% file: sections/conclusion.tex

\section{Conclusion}

Even when frontier MLLM agents can answer challenging document-grounded questions, they expend substantial effort without reliably recognizing \textit{if} and \textit{what} additional exploration is beneficial.

To support the shift toward calibrated, efficient reasoning, we release \benchmarknameshort{} alongside open-source implementations of our strong agentic baselines. Furthermore, Appendix~\ref{app:solution-design-space} characterizes a broad design space of architectural ``toggles,'' encouraging the benchmark's use for evaluating diverse strategies we did not explicitly consider.

Our primary findings suggest two concrete directions. First, \emph{episodic memory} could help agents learn corpus-specific terminology and document structure across queries. Second, \emph{reinforcement learning} with search tool feedback could significantly improve exploration policies.

As the community progresses, we will adapt our future evaluation campaigns to target new bottlenecks, ensuring the benchmark continues to serve as a discriminative signal for frontier capabilities.

%% file: sections/limitations.tex
\section{Limitations}
\paragraph{Coverage and Representativeness.}
\benchmarknameshort{} targets multi-step, grounded reasoning over heterogeneous PDFs, but the corpus is limited to English-language documents and is predominantly U.S.-centric in content, formatting conventions, and institutional terminology. As a result, performance on MADQA may not transfer to non-English settings or to regions with distinct document genres, layouts, and administrative practices.

Moreover, documents are sourced from publicly available repositories, which may underrepresent private, proprietary, or highly domain-specific enterprise document collections (e.g., internal policies, regulated clinical workflows, bespoke procurement forms).

\paragraph{Public Documents and Training Data Exposure.}
Because documents in the corpus are public, some content may have been observed during model pretraining. While we analyze and quantify answer guessability to bound this effect, we cannot guarantee the complete elimination of memorization advantages, and such exposure may differentially benefit certain model families.

Consequently, reported scores should be interpreted as a mixture of document-grounded competence and (to a limited extent) prior exposure for a subset of items.

\paragraph{Annotation Granularity and Task Scope.}
We provide minimal evidence at the \emph{page} level rather than at finer granularity (e.g., bounding boxes, table cells, or token spans). This choice supports reliable annotation at scale and aligns with how retrieval systems operate. Still, it limits the benchmark's ability to diagnose fine-grained grounding failures (e.g., extracting the wrong field within a form) when the correct page is retrieved.

In addition, \benchmarknameshort{} focuses on attributed question answering; it does not directly evaluate other critical document-workflow capabilities such as long-horizon planning across many tasks, document editing, data entry into external systems, or end-to-end automation with real-world side effects.

\paragraph{Evaluation Methodology and Residual Noise.}
We rely on an LLM-as-a-judge to score answer correctness, calibrated to human judgments, but this evaluation is still imperfect. It does not include systematic human verification for every prediction. Despite calibration, rare formatting edge cases and ambiguous surface forms can lead to residual measurement error.

Furthermore, correctness and attribution are scored separately (answer accuracy vs.\ Page/Doc F1), which means a system can be credited for a correct answer even when its citations are incomplete or noisy; conversely, a system can be penalized for imperfect localization even when it has effectively identified the proper document.

These design choices reflect the benchmark’s emphasis on both \emph{task completion} and \emph{grounded navigation}, but they may not match every deployment objective.

\paragraph{Effort as a Proxy for Uncertainty.}
Our calibration analysis follows the behavioral assumption that \emph{greater observed effort} (e.g., more search/tool iterations) reflects higher \emph{implicit uncertainty} or difficulty, and we quantify whether agents allocate effort in a way that correlates with success.

However, the specific operationalization of effort as \emph{step count} is inherently system-dependent: a "step" can bundle different amounts of computation, retrieved context, or latency depending on tool design and model interface.

As a result, Kuiper scores are most interpretable when comparing methods under a consistent agent/tooling regime. They should ideally be reported alongside complementary cost measures (tokens, wall-clock time, or dollar cost) when these differ across systems.

\paragraph{Sensitivity of Ranking to Budget.}\looseness=-1 Our evaluation imposes a fixed upper bound on agent steps (typically $N=10$), but system rankings are susceptible to this parameter. As shown in our efficiency profiling (Section~\ref{sec:human-agent}), the relative order of frontier models inverts depending on the allowed budget: for instance, while Gemini 3 Pro dominates the leaderboard at convergence, it underperforms both humans and Claude Sonnet 4.5 in the $N \le 3$ regime.

Consequently, the reported benchmark scores should be interpreted as ``asymptotic performance'' under a generous compute budget. Applications requiring real-time responsiveness may find that the ``best'' model on the leaderboard is suboptimal for constrained inference settings, necessitating a shift from scalar accuracy metrics to budget-aware evaluation curves.

\paragraph{Robustness, Safety, and Responsible Use.}
The benchmark evaluates benign documents and does not measure robustness to adversarial inputs (e.g., prompt-injection instructions embedded in PDFs, deceptive layouts, or malicious formatting intended to derail tool-using agents). In real deployments, such threats are central, and firm performance on \benchmarknameshort{} should not be taken as evidence of security against document-based attacks.

Finally, portions of the corpus include public records that may contain personal or sensitive information. \benchmarknameshort{} is intended for research and evaluation, and users should follow applicable policies for handling, redacting, and redistributing sensitive content.

%% file: appendix/appendix-main.tex

\newpage
\appendix
\onecolumn

\renewcommand{\partname}{}
\renewcommand{\thepart}{}
\part{Appendix}


\etocsettocstyle{\setlength{\parskip}{1pt}}{}
\etocsetnexttocdepth{3}
\localtableofcontents
\clearpage

\input{sections/reproducibility}
\input{appendix/benchmark-overview}

\input{appendix/corpus}

\input{appendix/annotation}

\input{appendix/subset-selection}

\input{appendix/metrics}

\input{appendix/baselines-details}

\input{appendix/failure-analysis}

\input{appendix/reproducibility}
\input{appendix/solution-space}

\input{appendix/related-works}

%% file: appendix/benchmark-overview.tex
\clearpage

\section{Benchmark Specification}

\subsection{Dataset Card}
\label{app:dataset_card}

\textbf{Dataset Summary.} The \benchmarkname{} Benchmark is a comprehensive dataset designed to evaluate agentic systems, RAG pipelines, and multimodal models on their ability to perform multi-step reasoning over collections of visually rich PDF documents. This dataset enforces a closed-world assumption where systems must navigate, retrieve, and reason across multiple pages or documents (e.g., cross-referencing annual reports, legal contracts, and government forms).

\begin{table}[h]
\centering
\caption{Summary of dataset statistics and splits.}
\label{tab:card_stats}
\small
\begin{tabular}{@{}ll@{}}
\toprule
\textbf{Metric} & \textbf{Value} \\
\midrule
\textbf{Total Questions} & \numquestions{} \\
\textbf{Total Documents} & \numdocs{} (sourced from DocumentCloud) \\
\textbf{Total Pages} & \numpages{} (median 5, mean 23.3, max 859 per doc) \\
\textbf{Total Tokens} & 12.2M (assuming Qwen3 tokenizer) \\
\textbf{Multi-hop Questions} & \multihoppercent{} (8.3\% cross-page, 9.0\% cross-document) \\
\textbf{Visual Necessity} & 58\% require layout/table/artifact understanding \\
\textbf{Domains} & \numdomains{} high-level domains (\numcategories{} categories) \\
\textbf{Language} & English \\
\bottomrule
\end{tabular}
\end{table}

\textbf{Data Splits.} The dataset is partitioned using \textit{Classical Test Theory} to ensure the test set retains high discrimination power while preserving hard questions, solved by none of the current models. Test: \testsize{} samples (inputs released; labels hidden for leaderboard evaluation). Development: \devsize{} samples (released with ground truth). Train: annotations released to facilitate RL-based optimization ($\sim$\trainsize{} samples).

\paragraph{Curation Rationale.} Existing benchmarks often lack the complexity to evaluate agentic capabilities such as planning and multi-step retrieval. This benchmark was created to bridge the gap between ``chat-with-PDF'' tasks and realistic document automation workflows involving heterogeneous layouts and long contexts.

\paragraph{Source Data.} Documents were manually curated from DocumentCloud, specifically looking for clusters of related documents (e.g., sequential reports, invoices, legal filings) to enable cross-document reasoning. The corpus covers multiple domains, including Financial, Legal, Government, Commercial, and Personal domains.

\paragraph{Annotation Process.} The benchmark required over 1,200 hours of professional annotation work. 100\% of questions and answers are human-generated by trained annotators who were restricted from using external knowledge. All answers are strictly grounded in the provided PDFs. A multi-stage pipeline was employed: (1) a pilot phase where 20 candidate annotators were evaluated on identical documents, (2) automated verification using GPT-5 with oracle evidence to flag potential errors, (3) manual review by domain experts from the authors' institution, and (4) final corrections based on baseline model and human performance analysis.

\paragraph{Biases and Limitations.} The majority of documents originate from the United States. The dataset is in English only. Guessability analysis shows that 9--15\% of answers (depending on the model) can be correctly guessed without document access, with approximately 3\% attributable to random chance on binary questions and the remainder to training data contamination from public documents.

\paragraph{Personal and Sensitive Information.}
The corpus references public records, which may contain personally identifiable information. The benchmark itself releases only questions, answers, and document identifiers---not the documents themselves. Users should exercise appropriate care when handling retrieved documents and follow applicable institutional guidelines for working with sensitive content.

\paragraph{Licensing and Data Access.} All human-generated question-answer pairs and evidence mappings are released under the CC BY-NC 4.0 license. Additionally, the benchmark provides a curated index of documents hosted on DocumentCloud by third-party organizations (e.g., news outlets, government agencies, non-profits). The authors do not own, host, or control DocumentCloud or the referenced documents.

\subsection{Desired Properties}
\label{app:problem_statement}
We formally define \textit{\taskname{}}, a task requiring systems to navigate, retrieve, reason over, and aggregate information from heterogeneous document collections.\\

\begin{definition}[Document Collection]
Let $\mathcal{C} = \{D_1, D_2, \dots, D_N\}$ be a corpus of $N$ multi-page PDF documents. Each document $D_i = (p_{i,1}, p_{i,2}, \dots, p_{i,|D_i|})$ is an ordered sequence of pages, where each page $p_{i,j}$ comprises visual content (layout, tables, figures) and textual content $\mathcal{T}(p_{i,j})$ representing its token sequence.
\end{definition}

\begin{definition}[\tasknameshort{}]
Given a corpus $\mathcal{C}$ and a natural language query $q$, the task is to produce:
\begin{enumerate}
    \item An answer $a = (t_1, t_2, \dots, t_k)$ as a sequence of tokens, and
    \item A minimal evidence set $\mathcal{E} = \{p_{i_1,j_1}, \dots, p_{i_m,j_m}\} \subseteq \mathcal{P}(\mathcal{C})$, where $\mathcal{P}(\mathcal{C}) = \bigcup_{i=1}^{N} \{p_{i,j}\}_{j=1}^{|D_i|}$ denotes all pages in the corpus, and the evidence set contains $m$ pages from potentially distinct documents. 
\end{enumerate}
\end{definition}

The task is characterized by six formal properties that distinguish it from standard document QA:

\paragraph{Desired Property 1: Extractive.} Answer tokens are drawn from the evidence pages rather than generated abstractly:
\begin{equation}
\forall t_\ell \in a: \exists\, p \in \mathcal{E} \text{ such that } t_\ell \in \mathcal{T}(p)
\end{equation}
This permits answers spanning multiple pages or documents, but requires lexical grounding.

\paragraph{Desired Property 2: Multi-Hop.} The evidence set may comprise multiple disjoint pages requiring aggregation:
\begin{equation}
|\mathcal{E}| \geq 1, \quad \text{with } |\mathcal{E}| > 1 \text{ for multi-hop questions}
\end{equation}
Multi-hop questions further decompose into \textit{cross-page} (evidence within one document: $\exists\, D_i: \mathcal{E} \subseteq D_i$) and \textit{cross-document} (evidence spanning documents: $\nexists\, D_i: \mathcal{E} \subseteq D_i$).

\paragraph{Desired Property 3: Closed-World.} The answer must be derivable solely from $\mathcal{C}$, independent of parametric world knowledge $\theta$ encoded in the model:
\begin{equation}
a = f(\mathcal{C}, q) \quad \text{where } f \text{ uses no knowledge from } \theta \setminus \mathcal{C}
\end{equation}
Any answer leveraging external facts constitutes a hallucination.

\paragraph{Desired Property 4: Grounded Attribution.} The answer must be faithfully attributed to the evidence set:
\begin{equation}
\textsc{Entails}(\mathcal{E}, a) = \texttt{True} \quad \land \quad \mathcal{E} \text{ is minimal}
\end{equation}
where minimality requires that no proper subset $\mathcal{E}' \subset \mathcal{E}$ suffices to derive $a$.

\paragraph{Desired Property 5: Agentic.} The task requires iterative retrieval and reasoning that cannot be solved in a single forward pass:
\begin{equation}
\nexists\, \text{query } q' : \textsc{Retrieve}(q', \mathcal{C}) \supseteq \mathcal{E} \text{ in one step}
\end{equation}
This necessitates \textit{planning} (decomposing $q$ into sub-queries), \textit{navigation} (iterating retrieval based on intermediate findings), and \textit{aggregation} (synthesizing partial answers across $\mathcal{E}$).

Unlike standard RAG benchmarks where relevant contexts are often provided or easily retrievable via lexical overlap, our formulation imposes that (1) information is often encoded in non-textual modalities or layout structures (see Figure~\ref{fig:hero}), (2) single retrieval steps are insufficient, and (3) the system must function under the closed-world assumption.

\jvl{\paragraph{Desired Property 6: Visual.} The evidence contains information encoded in non-textual modalities—such as spatial layout, table structure, figures, or graphical elements—that is necessary for deriving $a$:
\begin{equation}
\nexists\, a': \textsc{Entails}(\mathcal{T}(\mathcal{E}), a') \land \textsc{Equiv}(a', a)
\end{equation}
where $\mathcal{T}(\mathcal{E})$ denotes the linearized textual content (e.g., OCR output) stripped of visual structure. This property distinguishes tasks requiring genuine multimodal comprehension from those solvable via text extraction alone.}

\subsection{Benchmark Design Decisions}

We structure the development of this benchmark around  design decisions that distinguish agentic reasoning from standard QA. Table~\ref{tab:design_space} summarizes this design space, highlighting our specific choices.

\input{appendix/design-space-table}

\paragraph{Rationale for Page-Level Evidence.}\label{sec:bbox_rationale}

Unlike benchmarks with bounding-box or region-level annotations (e.g., ViDoRe v3, FinRAGBench-V), we annotate at page level. Our pilot study showed high inter-annotator agreement at this granularity, which also aligns with how humans navigate documents and how retrieval systems operate. For agentic evaluation, page-level evidence suffices to verify correct navigation without introducing annotation noise; finer localization can be delegated to downstream grounding models.


%% file: appendix/design-space-table.tex

{
\small
\setlength{\tabcolsep}{6pt}
\renewcommand{\arraystretch}{1.5}

\begin{longtable}{p{0.26\linewidth} p{0.33\linewidth} p{0.33\linewidth}}

\caption{\textbf{Design space for agentic document benchmark.} We map the broader landscape of document evaluation choices (center) against the specific instantiation in our benchmark (right), emphasizing strict PDF-grounding, multi-hop reasoning, and efficiency-aware metrics.}
\label{tab:design_space}\\
\toprule
\textbf{Axis} & \textbf{Choices in the Design Space} & \textbf{Our Decision in this Benchmark} \\
\midrule
\endfirsthead

\multicolumn{3}{c}%
{{\textit {Table \thetable\ (continued).} Design space for agentic document benchmarks and our instantiation.}}\\
\toprule
\textbf{Axis} & \textbf{Choices in the Design Space}  & \textbf{Our Decision in this Benchmark} \\
\midrule
\endhead

\midrule
\multicolumn{3}{r}{\footnotesize\textit{Continued on next page}}\\
\endfoot

\bottomrule
\endlastfoot

Corpus modality \& source
& Single web pages; HTML dumps; synthetic PDFs; scanned documents only; born-digital only; mixture of images and text; private vs.\ public corpora.
& Real-world, publicly hostable multi-page PDFs sourced from DocumentCloud; mixture of scanned and digital documents; benchmark definition is agnostic to whether systems use images, OCR, or layout-aware models. \\

Domain \& layout diversity
& Single domain (e.g., finance); few related domains; broad open-domain; homogeneous layouts (forms only); mixed narrative + tables; highly heterogeneous visual structure.
& Broad coverage of real-world domains (financial, legal, government, commercial, personal) with heterogeneous layouts, as detailed in \S\ref{app:domains}. \\

Document granularity
& Single isolated document; small set of thematically related documents; large corpus/collection per query; streaming documents.
& Each query is posed over a \emph{collection} of PDFs, encouraging cross-document and cross-page reasoning rather than single-document lookup. \\

\midrule
Grounding assumptions
& World-knowledge allowed; mixed (doc + world); purely corpus-grounded; open-ended generation without evidence requirement.
& Strictly PDF-grounded: by construction, every question is answerable using the provided documents alone; use of external world knowledge is unnecessary and treated as a source of hallucination. \\

Question source \& semantics
& Fully human-authored; LLM-generated; human-edited LLM generations; templated questions; questions that can be answered from a single span vs.\ questions requiring multi-hop reasoning.
& Fully human-authored questions created directly over the PDFs, with explicit control to ensure $\sim$20\% multi-hop cases (cross-page or cross-document) and strict PDF-grounding. \\

Answer format
& Free-form natural language; single spans; multiple spans; sets/lists; numeric or boolean values; open-ended rationales.
& Short, concrete answers: spans, booleans, lists, and numeric values. Answers are represented as small sets of strings (possibly multi-element for list questions). No rationales are required in the official metric. \\

Evidence granularity
& No evidence labels; document-level only; page-level; finer-grained regions (bounding boxes, table cells, tokens); full rationale chains.
& Minimal \emph{page-level} evidence sets, including cross-document and cross-page hops. For multi-hop questions, the annotation captures the full reasoning path by tagging all pages contributing to the final answer. We deliberately chose page-level over bounding-box annotations (see \S\ref{sec:bbox_rationale}). \\

Multi-hop definition
& No explicit notion of ``hop''; only single-context questions; soft multi-hop (multiple relevant snippets but not required); explicit hops with annotated paths.
& Multi-hop questions explicitly require combining information across pages or documents. We annotate the minimal set of pages required to solve the question and use this to define our principled splits and multi-hop subset. \\
Difficulty control
& No explicit notion of difficulty; manual tagging; heuristics (e.g., answer length); psychometric methods for subset selection; no hard items preserved.
& Difficulty estimated from model responses using Classical Test Theory. We select dev/test subsets to preserve discrimination and include a dedicated pool of hard items to avoid premature saturation. \\
\midrule
Answer correctness metrics
& Exact-match accuracy; token-level F1; ROUGE/bleu; normalized edit distance (ANLS); LLM-as-a-judge semantic scoring; numeric tolerance-aware metrics; per-type custom metrics.
& Task-optimized LLM-as-a-Judge. We found standard string\--matching metrics (like ANLS*) too strict. Our judge is calibrated to maximize agreement with humans and focuses on semantic equivalence rather than surface formatting. \\

Attribution \& retrieval metrics
& No grounding metrics; recall@k on documents; MRR/NDCG based on retrieval logs; F1 over documents; F1 over pages; span-level overlap; learned entailment / attribution scores.
& Two F1-style metrics on the final citations: \emph{\docf{}} (did the system identify the correct documents?) and \emph{\pagef{}} (did it cite the minimal set of gold pages?). \\

Faithfulness metrics
& Separate textual entailment models; LLM judges checking that answers are supported; combined answer+evidence scores; human audits on subsets.
& We rely on the benchmark design (PDF-only solvability) and \pagef{} as a proxy for groundedness: a correct answer with low \pagef{} is flagged as unfaithful. \\

Efficiency metrics
& Wall-clock latency; FLOPs or energy; number of retrieved documents; number of tool calls; number of model invocations; input/output token counts.
& Step Counts (number of tool calls), in our analyses supplemented by Inference Cost (USD). We penalize architectures that incur catastrophic overhead for marginal accuracy gains. \\

Calibration / process metrics
& Standard ECE/Brier score using explicit confidences; risk–coverage curves; abstain rates; trajectory-level metrics (e.g., number of retries vs.\ success).
& The \emph{\kuiper{}} statistic, derived from cumulative difference curves, to quantify \emph{effort calibration}. It measures whether an agent's decision to invest more compute (steps) correlates with success, or if it suffers from diminishing returns in long trajectories. \\
\midrule
Data splits \& training signal
& Evaluation-only benchmark; public train/dev, hidden test; additional unlabeled corpora; RL-ready logs; challenge sets only.
& Public train and dev splits with full annotations, plus a CTT-selected held-out test set served via a test server. We explicitly design the train split to support RL and prompt-search on realistic agent traces, while keeping the test set hidden. \\

Contamination \& governance
& No contamination checks; ad-hoc manual checks; explicit URL/document-level overlap analysis; rotating hidden tests; no leaderboard policy vs.\ strict disclosure.
& We commit to semantic versioning of the dataset and metrics, and a hidden test server with periodic refreshes. Submissions must disclose models, training data statements, and compute budgets. \\

\end{longtable}
}

%% file: appendix/corpus.tex
\clearpage
\section{Document Corpus}

\subsection{Domains and Categories}\label{app:domains}

The benchmark relies on real-world PDFs sourced from DocumentCloud. Unlike datasets focused solely on academic papers or financial reports, our corpus includes high variability in layout, OCR quality, and visual density. Table~\ref{tab:category_stats_full} lists the primary document categories present in the test set.

\input{appendix/category-stats-table}


\subsection{Layout Element Density Heatmap}\label{app:layout_elemt_density}

To visualize the distribution of layout elements across document categories, we construct a heatmap with per-element z-score normalization. 

For each document, we use the Granite-Docling MLLM \cite{livathinos2025docling} to extract layout elements from PDF pages. The model identifies various element types including tables, figures, lists, headers, text blocks, checkboxes, and other structural components.

For each document category $c$ and element type $e$, we compute the element density as the average number of elements per page:
\begin{equation}
    d_{c,e} = \frac{\sum_{p \in P_c} n_{p,e}}{|P_c|}
\end{equation}
where $P_c$ is the set of all pages in category $c$, and $n_{p,e}$ is the count of element type $e$ on page $p$.

To enable meaningful comparison across element types with different baseline frequencies, we apply column-wise (per-element) z-score normalization. For each element type $e$, we compute:
\begin{equation}
    z_{c,e} = \frac{d_{c,e} - \mu_e}{\sigma_e}
\end{equation}
where $\mu_e = \frac{1}{|C|}\sum_{c \in C} d_{c,e}$ is the mean density across all categories, and $\sigma_e$ is the corresponding standard deviation. This normalization centers each element's distribution at zero, allowing us to identify categories with unusually high or low presence of specific elements regardless of the element's overall frequency.

The resulting z-scores are displayed using a diverging colormap where cyan indicates below-average density ($z < 0$), white indicates average density ($z \approx 0$), and pink indicates above-average density ($z > 0$). Cell annotations show the original density values $d_{c,e}$ (elements per page), while the color intensity reflects the magnitude of deviation from the element-specific mean. Element labels include the corpus-wide mean $\mu_e$ for reference.





\begin{figure}[hp] 
    \includegraphics[width=0.85\linewidth,trim={2.5cm 2cm 3.6cm 3cm},clip]{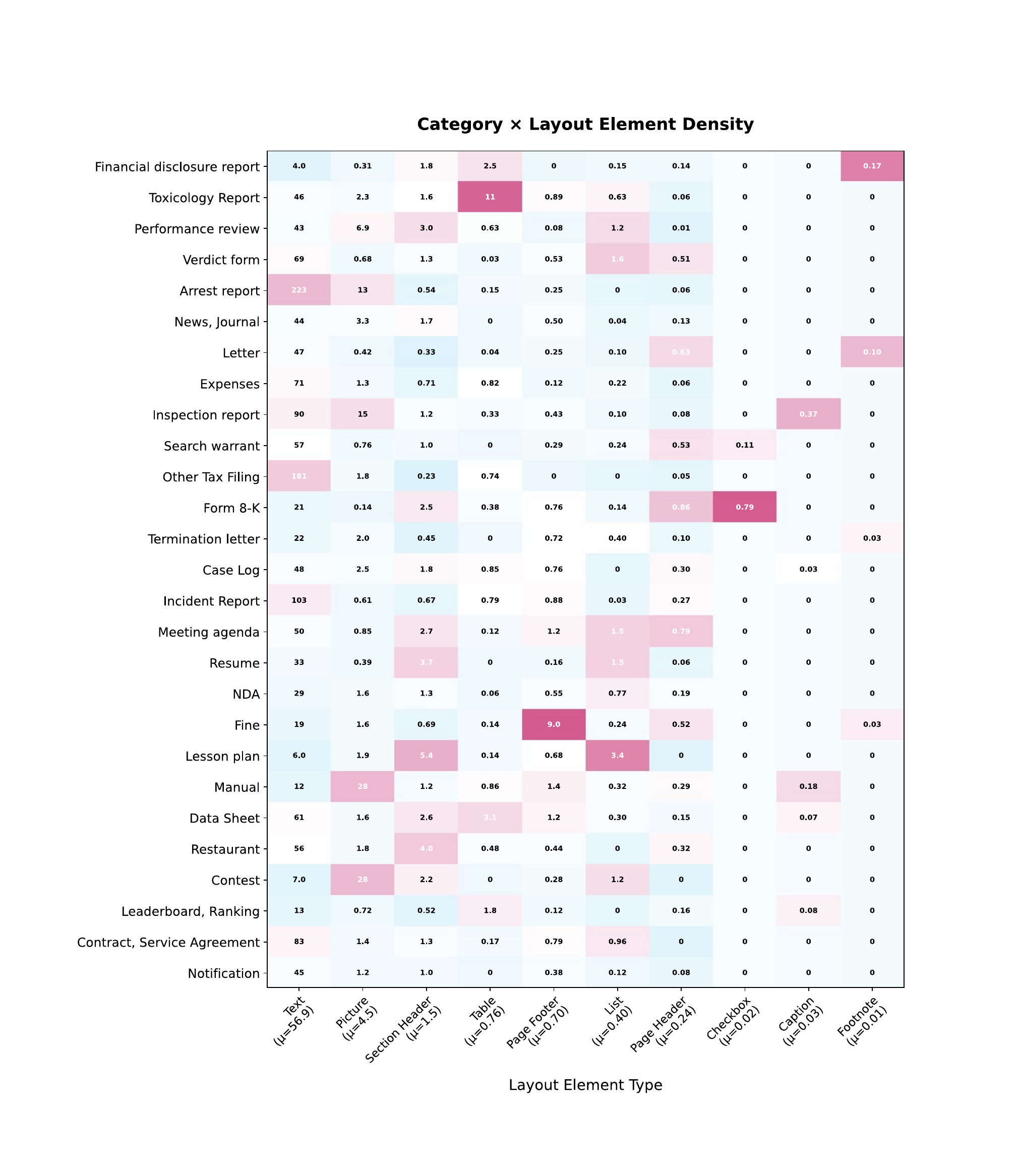}
    
    \caption{\textbf{Per-category layout element density.} Cell values indicate average elements per page; colors show z-score normalized deviation from the element-type mean. \textit{(Continued on next page...)}}
    \label{fig:category_layout_heatmap_detailed}
\end{figure}


\begin{figure}[hp] 
    \ContinuedFloat 
    \includegraphics[width=0.85\linewidth,trim={0.4cm 1cm 0.8cm 2cm},clip]{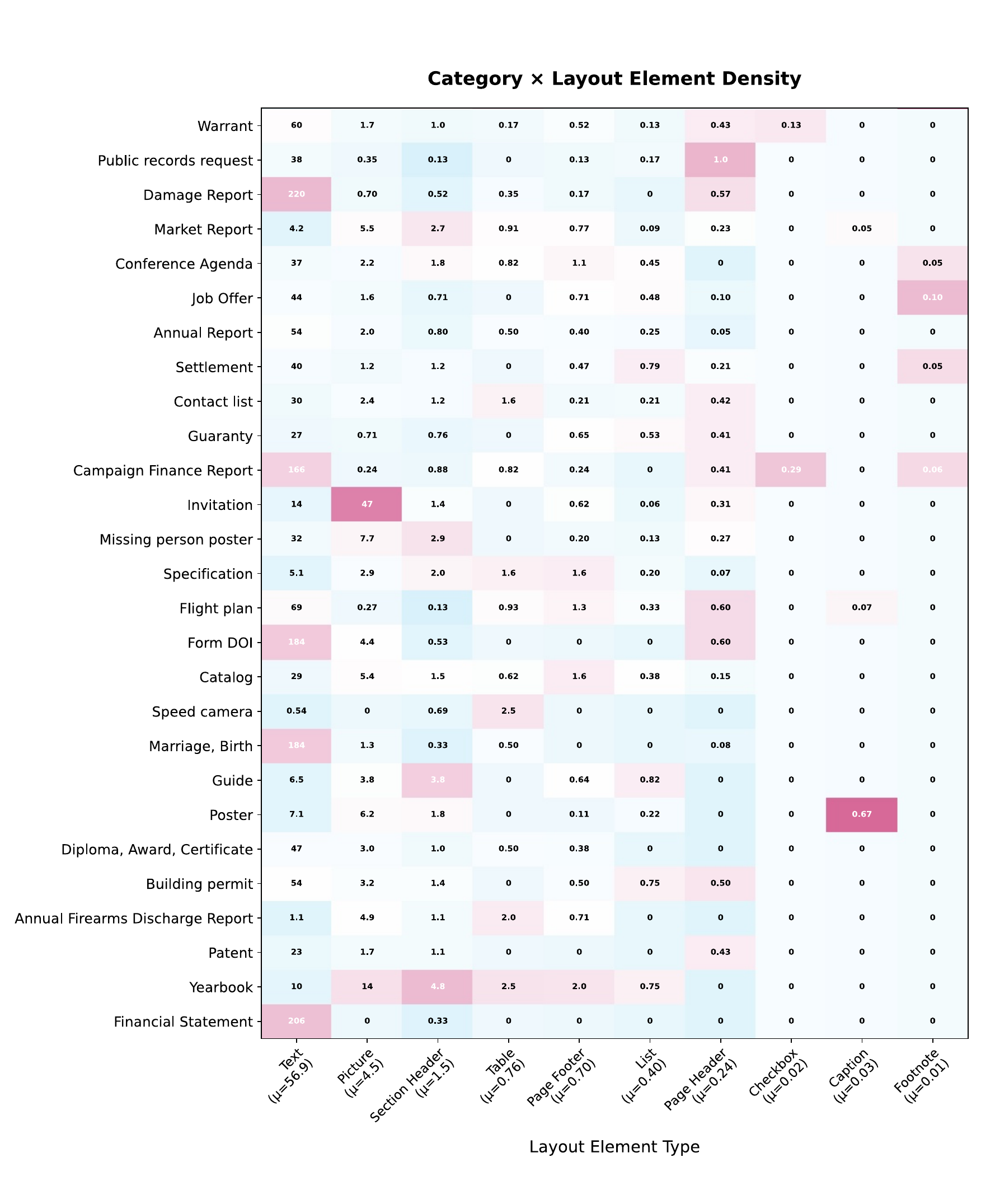}

    \caption[]{\textbf{Per-category layout element density} \textit{(Continued).} Categories such as \emph{Yearbook} and \emph{News, Journal} exhibit high picture density, while \emph{990 Form} and \emph{Budget} show elevated table presence. Text-heavy categories like \emph{Settlement} and \emph{Patent} display high text block counts but minimal visual elements. This heterogeneity ensures the benchmark tests model robustness across diverse layout structures.}
\end{figure}


\clearpage

%% file: appendix/category-stats-table.tex
{\small\begin{longtable}{lrrrrr}
\caption{\textbf{Detailed statistics per document category.} Categories are sorted by document count. The corpus exhibits high variance in document length, from single-page posters to 859-page expense document.}
\label{tab:category_stats_full} \\
\toprule
\textbf{Category} & \textbf{Docs} & \textbf{Pages} & \textbf{Mean} & \textbf{Med.} & \textbf{Max} \\
\midrule
\endfirsthead

\multicolumn{6}{c}{\tablename\ \thetable\ (continued)} \\
\toprule
\textbf{Category} & \textbf{Docs} & \textbf{Pages} & \textbf{Mean} & \textbf{Med.} & \textbf{Max} \\
\midrule
\endhead

\midrule
\multicolumn{6}{r}{\textit{Continued on next page}} \\
\endfoot

\bottomrule
\endlastfoot

    Annual Report & 30 & 1,754 & 58.5 & 33 & 222 \\
    990 Form & 20 & 1,890 & 94.5 & 36 & 697 \\
    Verdict form & 20 & 143 & 7.2 & 3 & 46 \\
    News, Journal & 20 & 446 & 22.3 & 10 & 80 \\
    Expenses & 20 & 914 & 45.7 & 2 & 859 \\
    Lesson plan & 20 & 247 & 12.3 & 10 & 46 \\
    Toxicology Report & 19 & 77 & 4.1 & 4 & 7 \\
    Inspection report & 19 & 227 & 11.9 & 4 & 102 \\
    Performance review & 19 & 87 & 4.6 & 4 & 9 \\
    Termination letter & 18 & 50 & 2.8 & 2 & 10 \\
    Financial disclosure report & 17 & 111 & 6.5 & 7 & 9 \\
    Financial Statement & 16 & 794 & 49.6 & 46 & 163 \\
    Form 8-K & 16 & 165 & 10.3 & 3 & 85 \\
    Incident Report & 15 & 131 & 8.7 & 3 & 40 \\
    Other Tax Filing & 15 & 208 & 13.9 & 6 & 70 \\
    Guide & 15 & 422 & 28.1 & 20 & 124 \\
    Market Report & 15 & 435 & 29.0 & 13 & 139 \\
    Arrest report & 15 & 65 & 4.3 & 4 & 10 \\
    Sustainability report & 15 & 595 & 39.7 & 41 & 54 \\
    Search warrant & 14 & 148 & 10.6 & 8 & 27 \\
    Climate action plan & 14 & 550 & 39.3 & 37 & 80 \\
    Missing person poster & 14 & 15 & 1.1 & 1 & 2 \\
    Meeting agenda & 14 & 44 & 3.1 & 2 & 11 \\
    Form DOI & 14 & 15 & 1.1 & 1 & 2 \\
    Catalog & 13 & 573 & 44.1 & 34 & 248 \\
    Public records request & 13 & 23 & 1.8 & 2 & 4 \\
    Contact list & 13 & 224 & 17.2 & 2 & 177 \\
    Campaign Finance Report & 13 & 399 & 30.7 & 33 & 84 \\
    Resume & 13 & 31 & 2.4 & 2 & 3 \\
    Warrant & 12 & 125 & 10.4 & 5 & 41 \\
    Notification & 12 & 25 & 2.1 & 2 & 6 \\
    Fine & 11 & 88 & 8.0 & 2 & 57 \\
    Restaurant & 11 & 48 & 4.4 & 3 & 12 \\
    Yearbook & 11 & 1,046 & 95.1 & 69 & 248 \\
    Data Sheet & 11 & 77 & 7.0 & 4 & 21 \\
    Contract, Service Agreement & 11 & 76 & 6.9 & 4 & 18 \\
    Audit report & 10 & 624 & 62.4 & 48 & 164 \\
    Annual Firearms Discharge Report & 10 & 720 & 72.0 & 78 & 102 \\
    Manual & 10 & 463 & 46.3 & 15 & 318 \\
    Patent & 10 & 132 & 13.2 & 14 & 31 \\
    Employee handbook & 10 & 624 & 62.4 & 61 & 103 \\
    Letter & 10 & 75 & 7.5 & 6 & 23 \\
    Settlement & 10 & 670 & 67.0 & 8 & 533 \\
    Budget & 10 & 958 & 95.8 & 31 & 330 \\
    Building permit & 10 & 115 & 11.5 & 2 & 44 \\
    Invitation & 10 & 16 & 1.6 & 1 & 4 \\
    Leaderboard, Ranking & 10 & 25 & 2.5 & 1 & 8 \\
    Poster & 10 & 10 & 1.0 & 1 & 1 \\
    Damage Report & 10 & 242 & 24.2 & 3 & 205 \\
    Contest & 10 & 25 & 2.5 & 2 & 5 \\
    Crop report & 10 & 218 & 21.8 & 19 & 40 \\
    Biography & 10 & 30 & 3.0 & 2 & 11 \\
    Guaranty & 9 & 79 & 8.8 & 7 & 22 \\
    Case Log & 9 & 43 & 4.8 & 4 & 10 \\
    Conference Agenda & 9 & 32 & 3.6 & 3 & 9 \\
    Marriage, Birth & 9 & 10 & 1.1 & 1 & 4 \\
    NDA & 9 & 54 & 6.0 & 4 & 23 \\
    Job Offer & 8 & 21 & 2.6 & 2 & 7 \\
    Specification & 8 & 302 & 37.8 & 12 & 139 \\
    Speed camera & 6 & 13 & 2.2 & 1 & 5 \\
    Diploma, Award, Certificate & 6 & 8 & 1.3 & 1 & 3 \\
    Flight plan & 5 & 15 & 3.0 & 1 & 8 \\
    SEC filing & 4 & 827 & 206.8 & 189 & 274 \\
\end{longtable}}

%% file: appendix/annotation.tex
\clearpage
\section{Annotations and Human Baseline}


\subsection{Annotation Guidelines}\label{app:guidelines}

Annotators were tasked with creating Question-Answer (QA) pairs grounded strictly in the provided PDF collections. The core constraints were:
\begin{itemize}
    \item Questions must be unanswerable without the documents (e.g., forbidding generic facts like ``What is the capital of Poland?'').
    \item Questions must be unambiguous without self-referential location cues.
    \begin{itemize}
        \item \xmark{} {Bad:} ``What is the title of \textit{this} document?'' (Ambiguous in a collection).
        \item \xmark{} {Bad:} ``What is the email on \textit{page 4}?'' (Trivializes retrieval).
        \item \cmark{} {Good:} ``What is the email address of Site Director William Wood?'' (Assumes uniqueness in the corpus).
    \end{itemize}
    \item The question must be answerable using the text or visual elements present in the PDFs.
\end{itemize}

For every question, annotators provided the \textit{Minimal Evidence Set}, i.e., the specific file names and page numbers required to answer the question. If a question requires comparing two menus to find the ``most expensive spaghetti,'' pages from \textit{both} menus must be listed as evidence, even if the answer comes from only one. If evidence appears multiple times (duplication), only the most plausible or first occurrence is noted.

Annotators were instructed to ensure $\sim$20\% of questions required \textit{Multiple Pieces of Evidence}. These questions fall into specific reasoning categories:

\begin{enumerate}
    \item \textit{Bridging.} The answer to part A is required to find part B.
    \begin{quote}
        {Example:} ``What is the national bird of the nation that has a negative carbon footprint?'' (Requires finding the nation with the footprint in Doc A, then finding its bird in Doc B).
    \end{quote}
    \item \textit{Comparison.} Aggregating values across documents.
    \begin{quote}
        {Example:} ``Which country has won more soccer World Cups: Argentina or Brazil?''
    \end{quote}
    \item \textit{Common Properties.} Intersection of attributes.
    \begin{quote}
        {Example:} ``Who is the only person to win an Olympic medal and a Nobel prize?''
    \end{quote}
\end{enumerate}


\input{appendix/human-study}

\input{sections/examples}

\clearpage
\section{Principled Splits Creation}
\label{app:subset-selection}

This appendix details the Classical Test Theory (CTT) methodology used to split the complete evaluation dataset ($N=2250$) into the Test ($N=\testsize$) and Development ($N=\devsize$) subsets (the Train set consists of questions rejected from the Development and Test sets).
To ensure the subset is not biased toward a single architecture, we computed item statistics using predictions from 16 distinct models (Table~\ref{tab:models-performance}). These models cover a diverse range of providers, sizes, and capabilities.

\begin{table}[h]
\centering\footnotesize
\caption{Models used for Item Response Analysis (Sorted by Performance)}
\label{tab:models-performance}
\begin{tabular}{clllc}
\toprule
\textbf{Rank} & \textbf{Provider} & \textbf{Model Name} & \textbf{Model ID} & \textbf{\anlsstar{}} \\
\midrule
1 & Google & Gemini 3 Pro Preview & \texttt{gemini-3-pro-preview} & 0.688 \\
2 & Anthropic & Claude 4.5 Sonnet & \texttt{claude-sonnet-4-5-20250929} & 0.681 \\
3 & OpenAI & GPT-5 & \texttt{gpt-5-2025-08-07} & 0.647 \\
4 & Zhipu AI & GLM-4.6V & \texttt{zai-org/GLM-4.6V} & 0.616 \\
5 & Anthropic & Claude 4.5 Haiku & \texttt{claude-haiku-4-5-20251001} & 0.601 \\
6 & Alibaba & Qwen3-VL-235B Thinking & \texttt{Qwen/Qwen3-VL-235B-A22B-Thinking} & 0.598 \\
7 & Alibaba & Qwen3-VL-32B Thinking & \texttt{Qwen/Qwen3-VL-32B-Thinking} & 0.594 \\
8 & Google & Gemini 2.5 Pro & \texttt{gemini-2.5-pro} & 0.590 \\
9 & Google & Gemini 2.5 Flash & \texttt{gemini-2.5-flash} & 0.576 \\
10 & OpenAI & GPT-5 Mini & \texttt{gpt-5-mini-2025-08-07} & 0.550 \\
11 & OpenAI & GPT-5 Nano & \texttt{gpt-5-nano-2025-08-07} & 0.548 \\
12 & Alibaba & Qwen3-VL-8B Thinking & \texttt{Qwen/Qwen3-VL-8B-Thinking} & 0.530 \\
13 & Alibaba & Qwen3-VL-235B Instruct & \texttt{Qwen/Qwen3-VL-235B-A22B-Instruct} & 0.516 \\
14 & Zhipu AI & GLM-4.6V Flash & \texttt{zai-org/GLM-4.6V-Flash} & 0.444 \\
15 & OpenAI & GPT-4.1 Nano & \texttt{gpt-4.1-nano-2025-04-14} & 0.302 \\
16 & Alibaba & Qwen3-VL-32B Instruct & \texttt{Qwen/Qwen3-VL-32B-Instruct} & 0.300 \\
\bottomrule
\end{tabular}
\end{table}

\begin{figure}[b]
    \centering
    \includegraphics[width=0.9\linewidth]{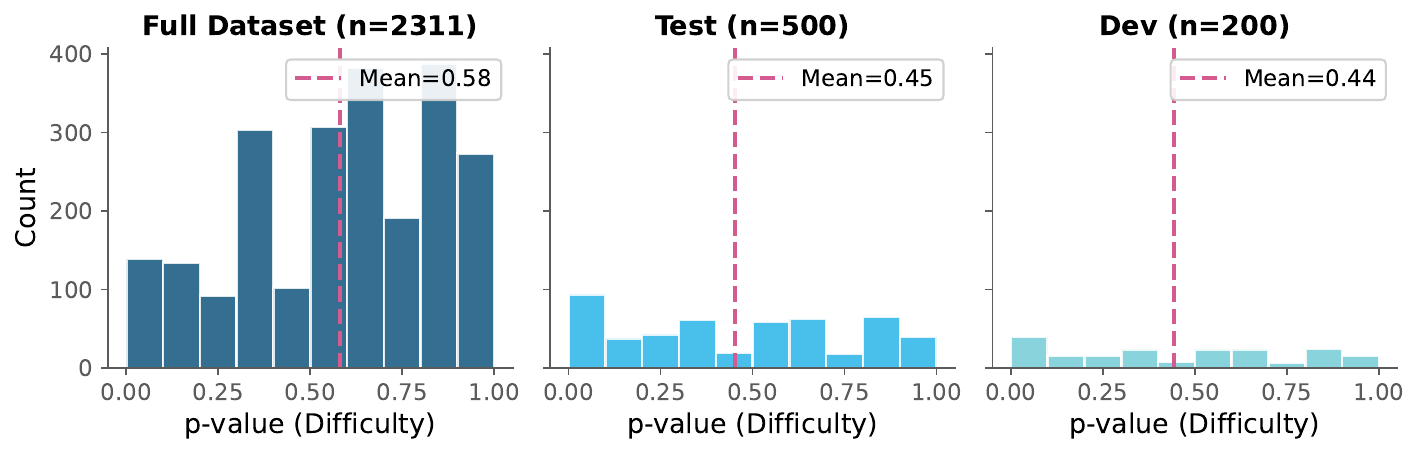}
    \caption{\textbf{Distribution of item difficulties} ($p$-values) in the Full Dataset vs. the Selected Test Subset. The selection strategy boosts the density of ``Hard'' and ``Sentinel'' items to preventing saturation.}
    \label{fig:p_value_dist}
\end{figure}

For each item $j$ in the dataset, we calculate two key metrics:

\begin{itemize}
    \item \textbf{Item Difficulty ($p_j$)} Defined as the mean accuracy of all models on item $j$. A lower $p$-value indicates a harder question:
    $p_j = \frac{1}{M} \sum_{i=1}^{M} X_{ij}$. (Note that ``$p$-value'' is used here to denote is distinct from the statistical $p$-value used in hypothesis testing to measure significance. We are consistent with how \citet{crocker1986introduction} uses the term.)
    \item \textbf{Item Discrimination ($r_{pbis}$)}
    We use the point-biserial correlation to measure how well item $j$ distinguishes between strong and weak models. To prevent autocorrelation, we use the \textit{corrected} total score, subtracting the item's own score from the total sum:$r_{pbis, j} = \text{Corr}(X_{\cdot, j}, S_{-j})  \text{where} \quad S_{-j} = \sum_{k \neq j} X_{\cdot, k}$.
\end{itemize}

Our selection strategy balances the need for high signal (high discrimination) with the need for future headroom (hard items). We select items with $p \le 0.1$ (solved by $\le$ 10\% of models) to form the Sentinel Pool. These items typically have low variance (and thus low discrimination metrics) but are crucial for measuring frontier capability. The remaining budget is filled by items with $p > 0.1$. We stratify these items into 9 uniform difficulty bins (e.g., $0.1 < p \le 0.2$, $0.2 < p \le 0.3$, etc.). Within each bin, we greedily select items with the highest discrimination ($r_{pbis}$) to maximize the subset's predictive power.

%% file: appendix/human-study.tex

\subsection{Human Baseline Collection}
\label{app:human-baseline}

To establish a meaningful upper bound on benchmark performance and enable behavioral comparison with LLM-based agents, we collected human baseline annotations on the full test set (\testsize{} questions). Unlike typical human evaluation setups where annotators have unrestricted access to documents, our human participants used \emph{the same retrieval interface} as our agentic baselines---a keyword-based search engine. This design choice allows us to compare not only answer accuracy but also retrieval strategies and search efficiency between humans and AI agents.

\begin{figure}[h]
    \centering
    \frame{\includegraphics[width=0.6\textwidth]{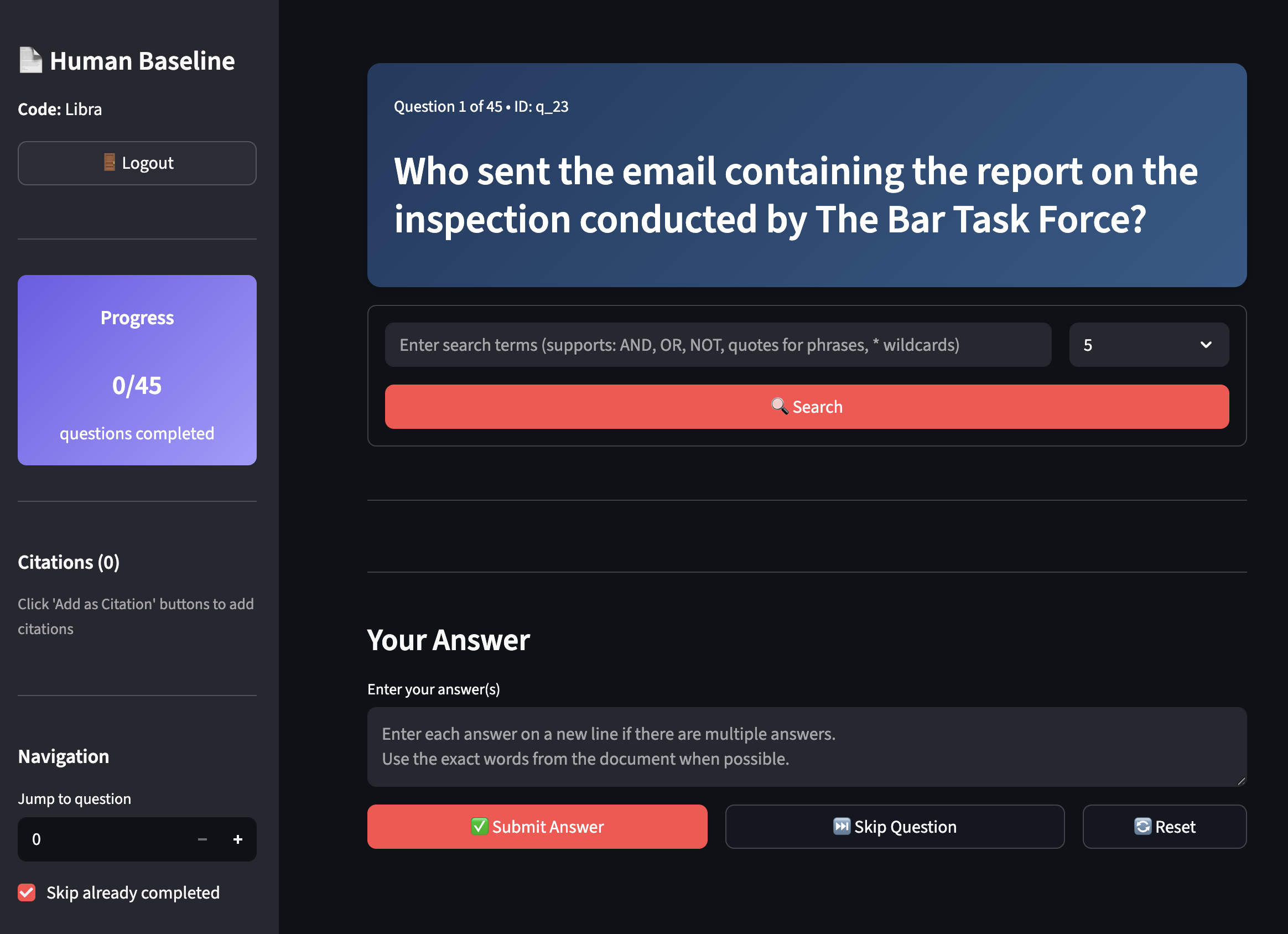}}
    \caption{\textbf{Main annotation interface.} The question appears at the top (blue box), followed by the search bar. Search results display as interactive PDF pages with navigation controls. The sidebar shows progress, selected citations, and search history.}
    \label{fig:human-main}
\end{figure}

We developed a custom web application using Streamlit that provides annotators with:

\begin{itemize}
    \item A Whoosh-based\footnote{\url{https://github.com/whoosh-community/whoosh}} BM25 search engine operating at the page level, supporting Boolean operators (AND, OR, NOT), phrase matching with quotes, and wildcard searches (Figure~\ref{fig:human-main}).
    \item An interactive PDF viewer with page navigation (first, previous, next, last, jump-to-page) allowing annotators to explore documents beyond the initial search results. 
    \item Ability to mark specific pages as evidence supporting the answer.
\end{itemize}

A key feature of our interface is comprehensive trajectory logging. For each question, we record: (1) Start and end timestamps, enabling per-question time measurement. (2) Every search query with timestamp, number of results returned, and the specific pages retrieved. (3) All document navigation actions (viewing pages, scrolling through documents) with timestamps. (4) Final evidence pages selected by the annotator.

This detailed logging enables analysis beyond accuracy, including average number of searches per question, time spent per question, retrieval precision (whether humans find evidence pages faster than agents), search query patterns and reformulation strategies.

\paragraph{Annotation Protocol.} The \testsize{} test questions were divided into 20 batches of 25 questions each. Each batch was assigned a constellation code name (Aquarius, Orion, Phoenix, etc.) for anonymized tracking. Annotators received the following instructions:

\begin{enumerate}
    \item Read the question carefully.
    \item Use the search interface to find relevant document pages. The search operates on OCR-extracted text and supports:
    \begin{itemize}
        \item Multiple keywords: \textit{mountain car} finds pages containing both words
        \item Boolean operators: \textit{mountain AND car}, \texttt{mountain OR car}
        \item Exact phrases: \textit{``annual report''} matches the exact phrase
        \item Wildcards: \textit{car*} matches ``car'', ``cars'', ``carriage'', etc.
    \end{itemize}
    \item Navigate within documents to find supporting evidence. Search results show individual pages, but annotators can browse the full document.
    \item Mark relevant pages as citations using the ``Add as Citation'' button.
    \item Enter the answer in the text box. For questions with multiple answers, enter each on a separate line.
\end{enumerate}

The human baseline interface was intentionally designed to match the capabilities available to our agentic systems. This controlled setup enables fair comparison of retrieval effectiveness and answer accuracy between humans and LLM-based agents operating under identical constraints.

%% file: sections/examples.tex



\clearpage
\twocolumn

\noindent
\begin{tabular}[!c]{ @{} m{0.49\textwidth} @{\hspace{0.05\textwidth}} m{0.49\textwidth} @{} }
    \begin{minipage}{\linewidth}
        \vspace{8em}
        \subsection{Sample Questions}\label{app:examples} 
        \textcolor{black}{We present a few examples to illustrate the reasoning complexity required by \benchmarknameshort{}.
        Success on our benchmark requires identifying relevant documents among hundreds of distractors and switching fluidly between modalities—reading handwriting, interpreting color-coded tables, and aggregating data across disjoint files.}\vspace{2.5em}
    \end{minipage}
    \begin{minipage}{\linewidth}
    \centering
        \frame{\includegraphics[width=0.9\linewidth]{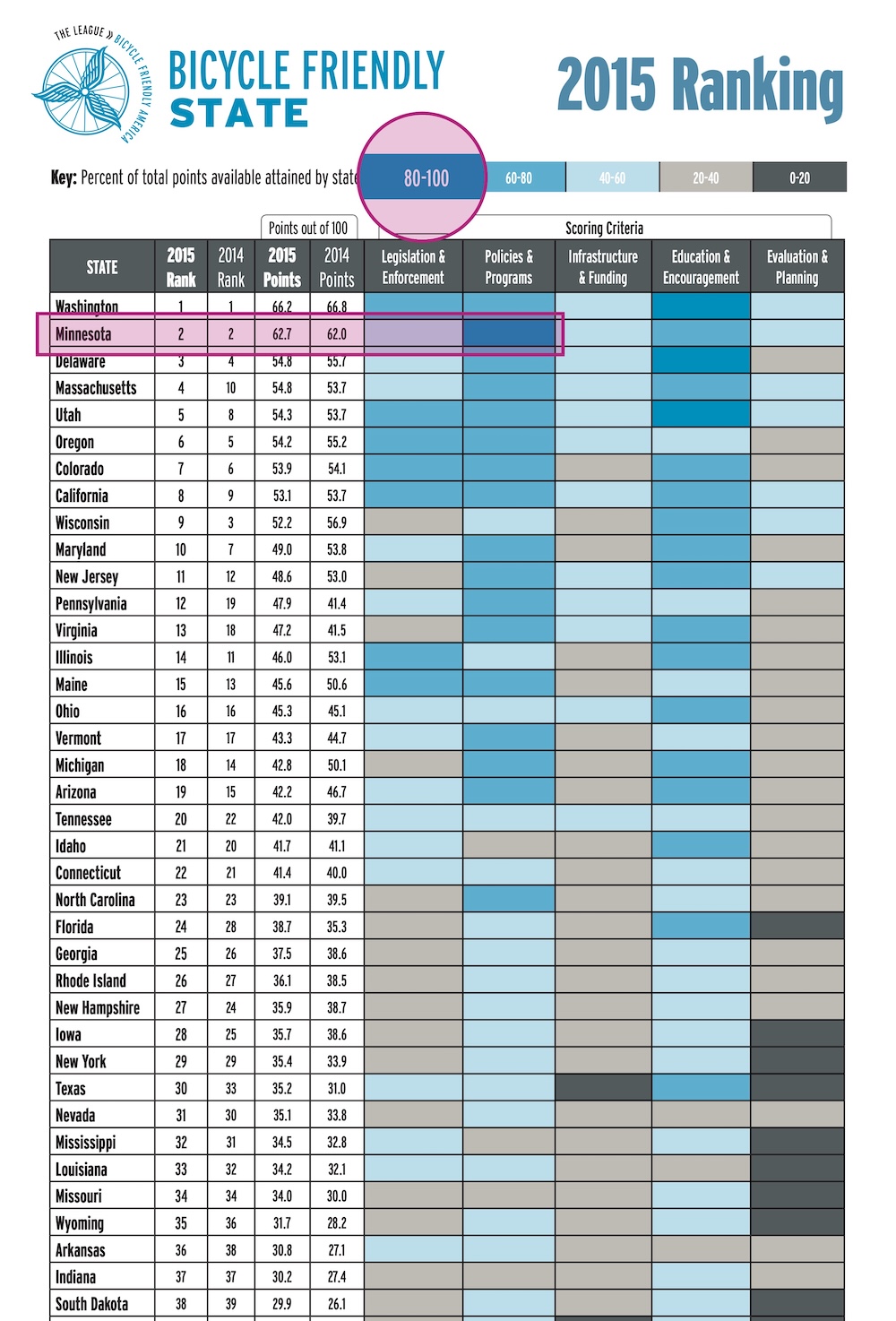}}
        \captionof*{figure}{\textbf{Which state had the highest score in terms of Policies \& Programs in the Bicycle Friendly State 2015 Ranking?} The agent must identify the correct ``2015 Ranking'' document among similar reports or yearbooks. The answer ``Minnesota'' cannot be found via text search because the value is encoded purely through color intensity (dark blue indicating ``80-100'').} 
        \label{fig:example_ranking}
    \end{minipage}
    &
    \begin{minipage}{\linewidth}
        \centering
        \frame{\includegraphics[width=0.8\linewidth]{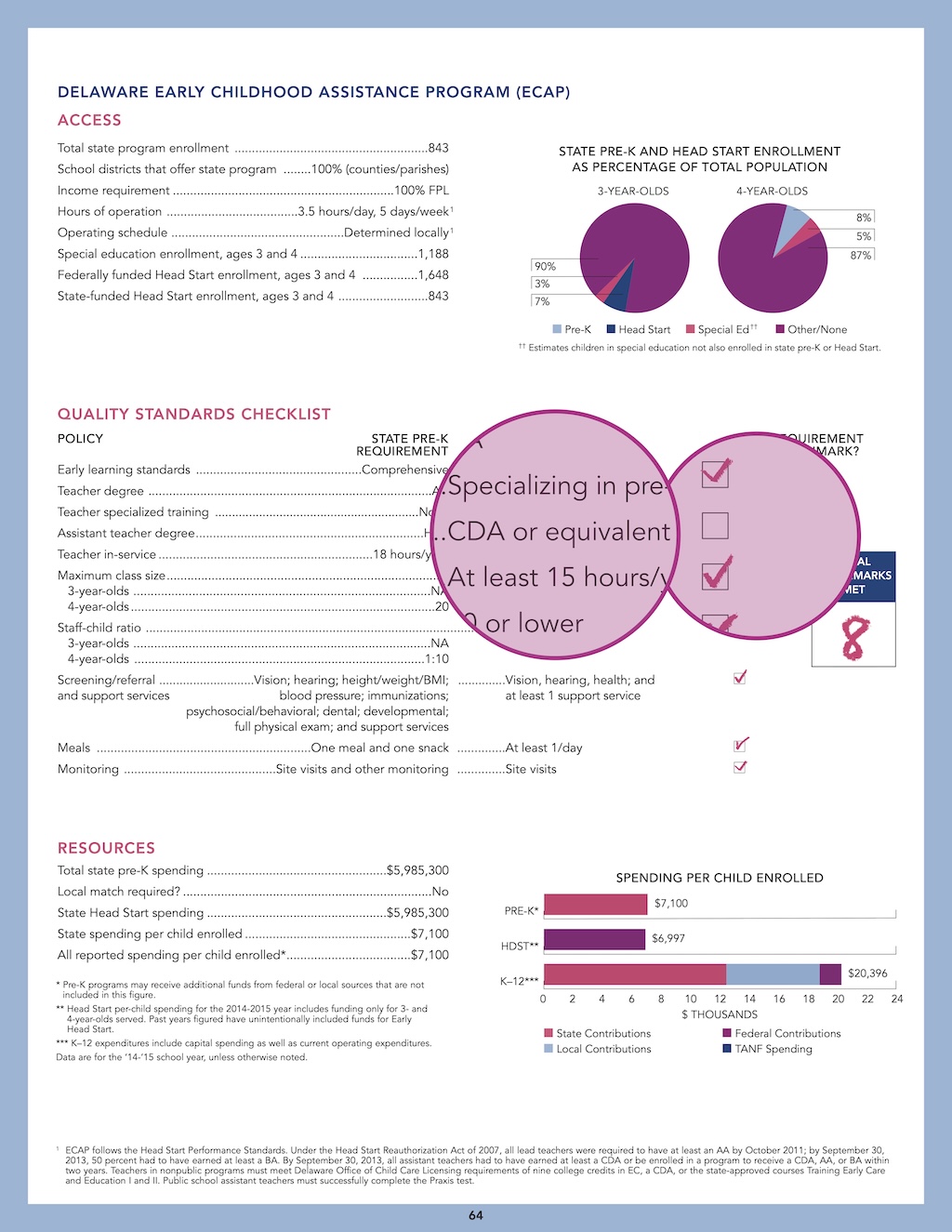}}
        \vspace{1em}\\
        \frame{\includegraphics[width=0.8\linewidth]{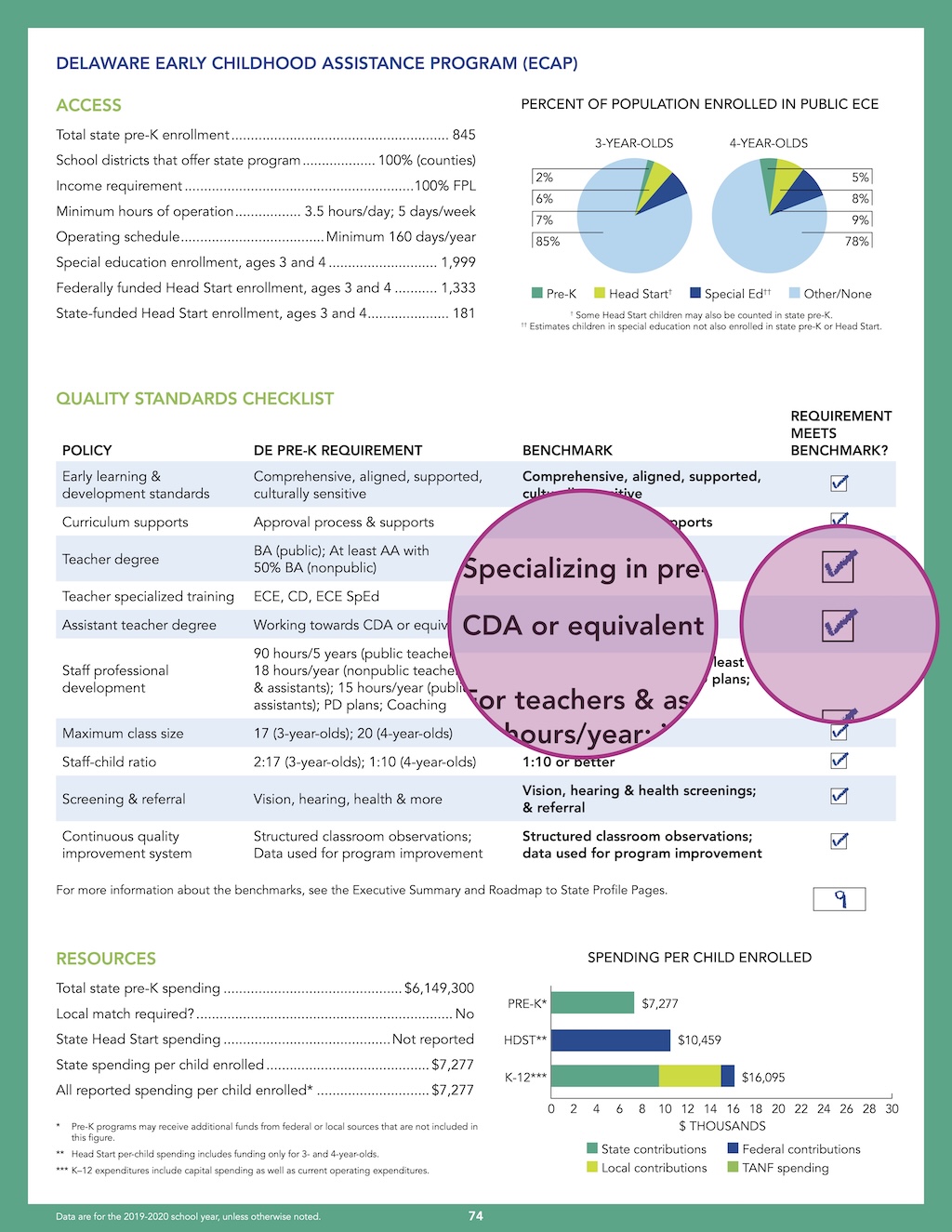}}
        \captionof*{figure}{\textbf{Which quality standards benchmark(s) for preschool programs did Delaware meet in the 2019-2020 school year that it did not meet in 2014-2015?} The agent must retrieve two State Preschool Yearbook editions from different years, locate the relevant benchmarks tables in each, interpret the associated checkboxes to determine which standards were met, and compare them to identify newly met standards.}
        \label{fig:example_standards}


    
    

    \end{minipage}
\end{tabular}

\clearpage

\begin{figure}[h]
    \vspace{3em}
    \centering
    \frame{\includegraphics[width=0.8\linewidth]{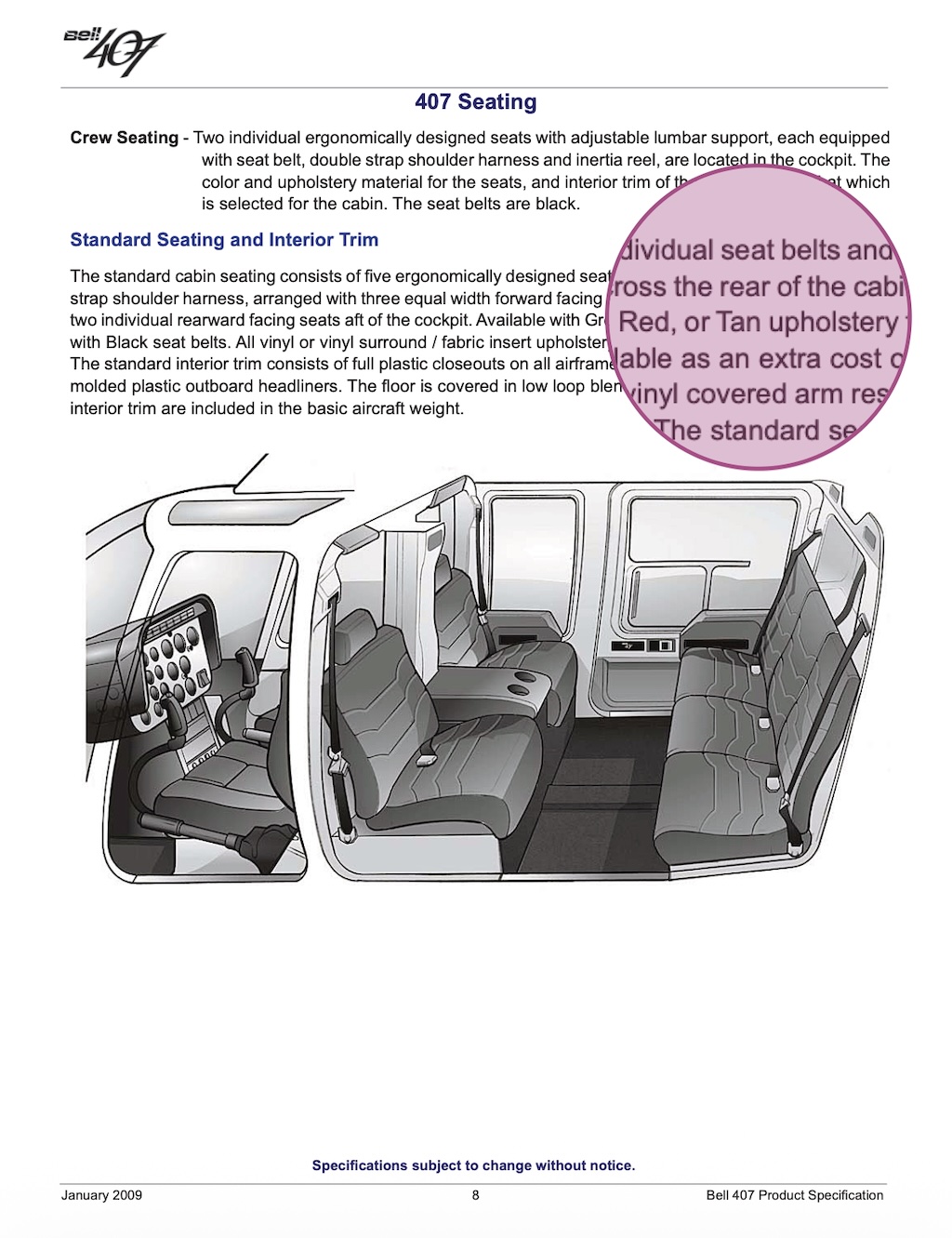}}
    \vspace{1em}\\
    \frame{\includegraphics[width=0.8\linewidth]{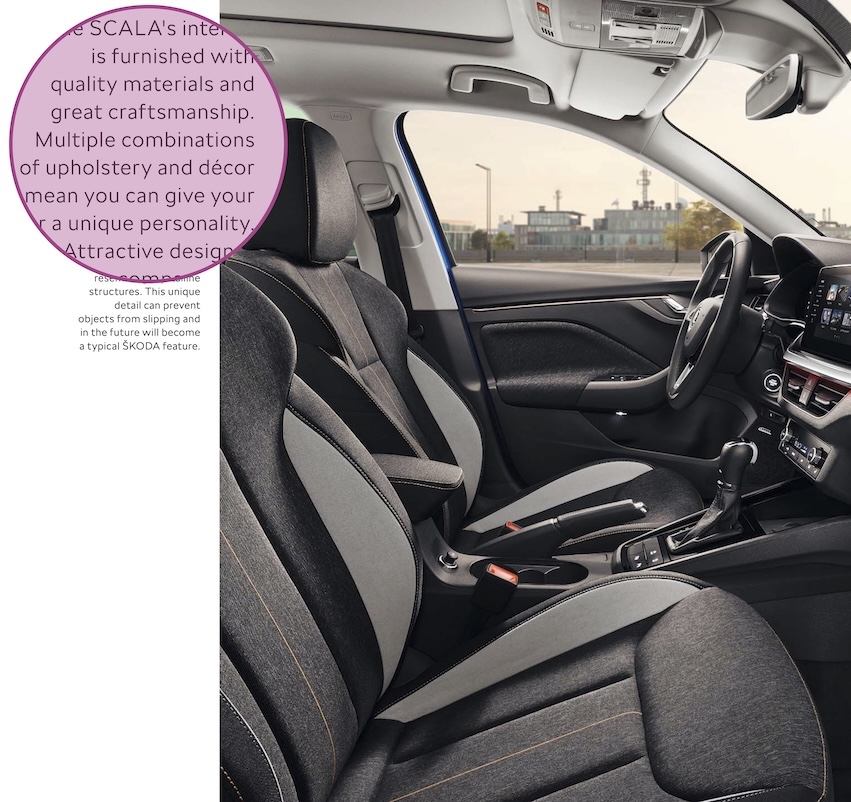}}
    \caption*{\textbf{Which vehicles offer multiple upholstery designs?} The agent must retrieve separate product brochures from a large collection and compare interior specification tables across documents to identify which vehicles list more than one upholstery option.}
    \label{fig:upholstery}
\end{figure}

\label{fig:example_piano}

\begin{figure}[h]
    \centering
    \frame{\includegraphics[width=0.8\linewidth]{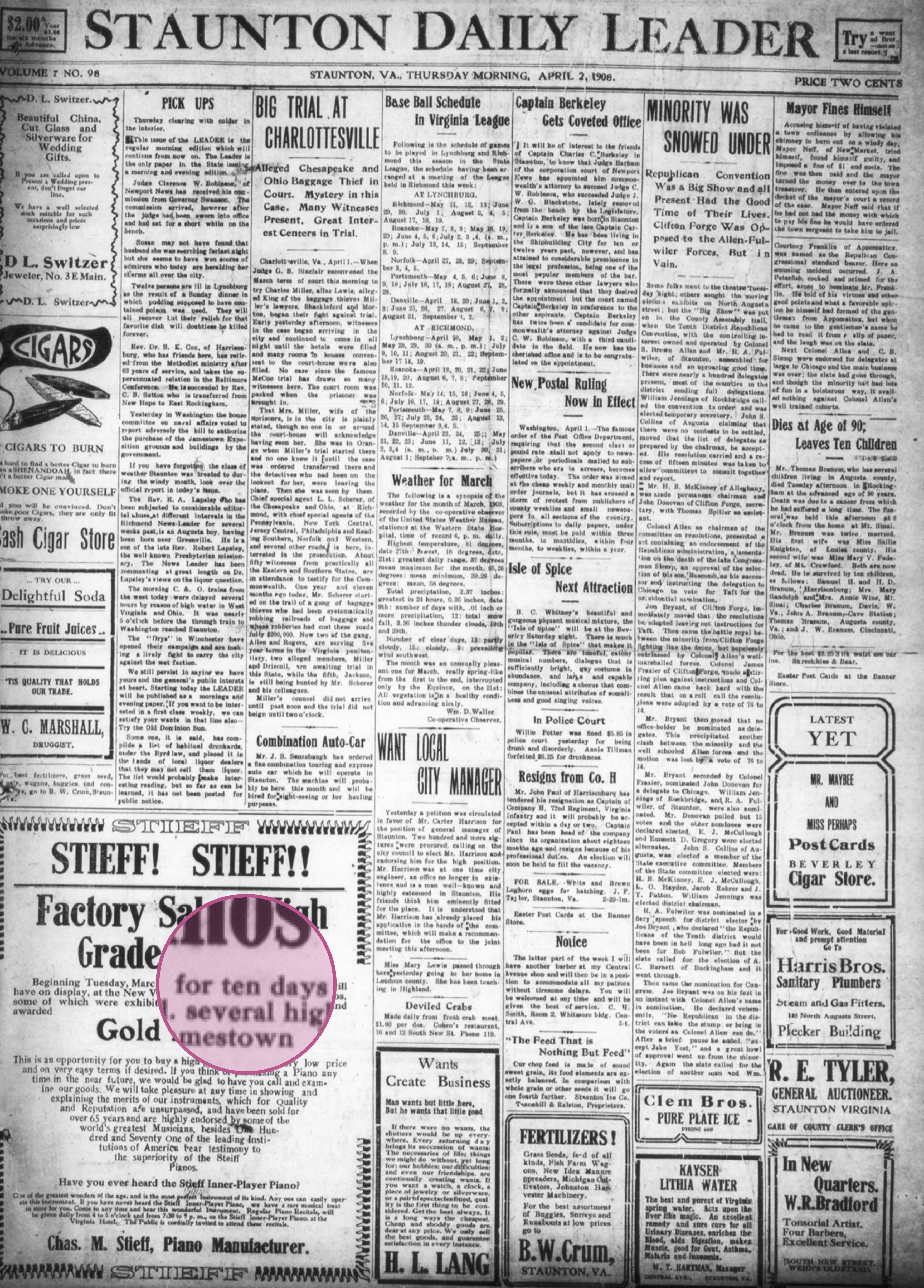}}
    \caption*{\textbf{How many days was the sale of pianos at the New Virginia Hotel planned to last?} The target information is buried within a dense, multi-column newspaper layout, likely surrounded by hundreds of unrelated articles and advertisements in the corpus. Standard chunking algorithms may fail to correctly segment the advertisement column from adjacent news stories.}
    \label{fig:example_sale}
\end{figure}

\begin{figure}[h]
    \centering
    \frame{\includegraphics[width=0.8\linewidth]{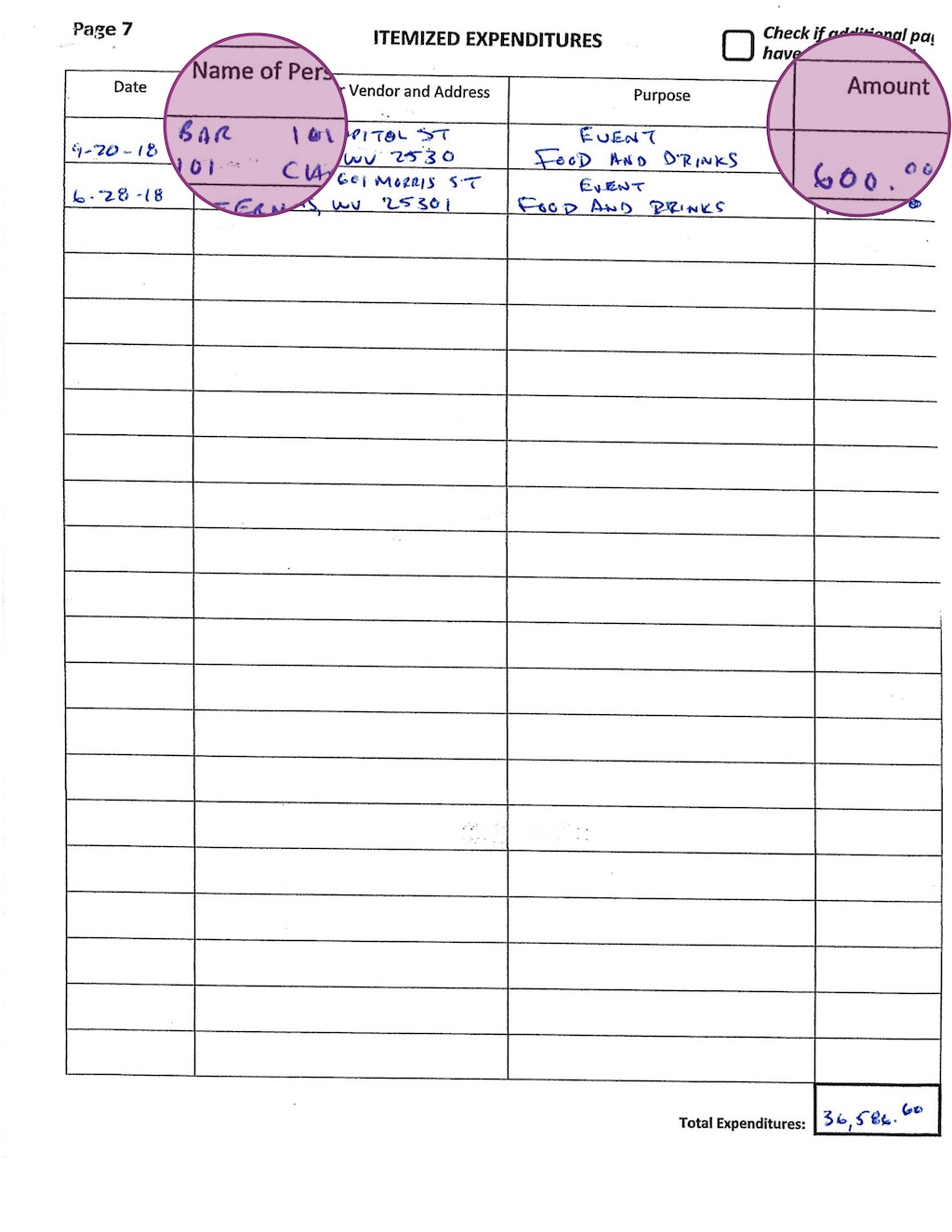}}
    \caption*{\textbf{How much was spent on Food and drinks at Bar 101 during the event?} The model must first determine that this information is in one of the financial statements within the collection. Once retrieved, the task involves reading handwritten entries in a table to extract the correct monetary value.}
    \label{fig:example_bar}
\end{figure}

\clearpage\onecolumn


%% file: appendix/subset-selection.tex
\clearpage
\section{Construct Validity Analyses}

\subsection{Lexical Overlap vs. Reasoning}

\label{app:lexical-solvability}

To verify that \benchmarknameshort{} cannot be solved through simple keyword matching (i.e., ``Ctrl+F'' strategies), we conducted a systematic n-gram analysis across the entire corpus.

For each question, we extract n-grams ($n \in \{1, 2, 3\}$) from the question text after removing stopwords and common question words (e.g., ``what,'' ``which,'' ``how,'' ``describe''). We then search the full corpus (\numpages{} pages) for pages containing these n-grams and measure: (1) \textit{Pages matched.} How many corpus pages contain at least one question n-gram. (2) \textit{Precision.} What fraction of matched pages are gold evidence pages. (3) \textit{Recall.} What fraction of gold evidence pages are retrieved.



\begin{table}[h]
\centering
\small
 \caption{\textbf{Lexical solvability analysis.} Unigrams and bigrams match too many pages (low precision), while trigrams miss half the questions entirely (low recall). No n-gram size enables effective Ctrl+F-based retrieval.} \label{tab:lexical-solvability}
\begin{tabular}{lrrr}
\toprule
\textbf{N-gram} & \textbf{Median Pages} & \textbf{Precision} & \textbf{Recall} \\
\midrule Unigram & 4,125 & 0.03\% & 99\% \\
Bigram  & 24    & 2.6\%  & 86\% \\
Trigram & 1     & 1.9\%  & 51\% \\
\bottomrule \end{tabular}\end{table}


This vocabulary gap between questions and evidence confirms that solving \benchmarknameshort{} requires semantic understanding to bridge the lexical mismatch, not just pattern matching. An effective system must reason about what information is needed and formulate appropriate search strategies---the core capability our benchmark is designed to evaluate.

\subsection{Parametric Knowledge vs. Grounding}
\label{app:guessability}


\begin{prompt}[t]
\begin{tcolorbox}[promptbox, title={\small\textbf{Guessing Prompt}}]
  \begin{tcolorbox}[psec-role]
    \vspace{4pt}
    {\small You are participating in an answer prediction experiment. You will be given a question about a document, but you will NOT see the document itself.}
    \vspace{4pt}
    {\small \textbf{Your task:} PREDICT or GUESS what the answer might be, based solely on the question text.}
  \end{tcolorbox}

  \begin{tcolorbox}[psec-warning]
    {\small\textbf{\textcolor{valencia!60!black}{Critical Rules}}}
    {\small
      \begin{enumerate}[leftmargin=*, topsep=4pt, itemsep=4pt]
        \item \textbf{You MUST always provide an answer} -- never refuse or say you cannot answer.
        \item \textbf{For yes/no questions:} Pick one randomly (Yes or No).
        \item \textbf{For specific values (dates, names, numbers):} Make your best guess based on common patterns.
        \item \textbf{For multiple formats:} Guess the most likely format.
        \item \textbf{For list-type questions:} Provide 1--3 plausible items.
      \end{enumerate}
    }
  \end{tcolorbox}

  \begin{tcolorbox}[psec-output]
    {\small\textbf{\textcolor{snowstar!60!black}{Output Format}}}
    {\small
      \begin{itemize}[label={-}, leftmargin=*, topsep=4pt, itemsep=2pt]
        \item Return a \textbf{list of values}.
        \item If there is a single answer, return a one-element list.
        \item If there are multiple items/entities, return multiple elements.
        \item Use as few words as possible.
        \item \textbf{Do NOT explain your reasoning} -- just provide the answer.
      \end{itemize}
    }
  \end{tcolorbox}

  \begin{tcolorbox}[psec-role]
    {\small\textbf{\textcolor{snowgray}{Remember}}}
    \vspace{2pt}
    {\small This is a guessing experiment. There is no "wrong" answer - we want to see what predictions are possible from the question alone.}
  \end{tcolorbox}
\end{tcolorbox}
\caption{\textbf{Guessing prompt} for parametric knowledge analysis. Models are instructed to predict answers without document evidence to measure training data contamination.}
\label{prompt:guessing}
\end{prompt}


\begin{prompt}[t]
\begin{tcolorbox}[promptbox, title={\small\textbf{Question Classification Prompt}}]
  \begin{tcolorbox}[psec-role]
    \vspace{4pt}
    {\small You are classifying questions for an analysis of answer guessability.}
  \end{tcolorbox}

  \begin{tcolorbox}[psec-steps]
    {\small\textbf{\textcolor{snowdark}{Categories}}}
    \vspace{4pt}
    {\small Classify each question into \textbf{ONE} of these categories:}
    {\small
      \begin{enumerate}[leftmargin=*, topsep=4pt, itemsep=6pt]
        \item \textbf{``yes\_no''} -- Questions that have Yes or No as the answer.
        \begin{itemize}[label={-}, leftmargin=*, topsep=0pt, itemsep=0pt]
          \item \textit{Examples:} "Does X offer Y?", "Is X greater than Y?", "Did X happen?"
        \end{itemize}
        \item \textbf{``binary\_choice''} -- Questions asking to choose between exactly 2 options mentioned in the question.
        \begin{itemize}[label={-}, leftmargin=*, topsep=0pt, itemsep=0pt]
          \item \textit{Examples:} "Which is larger: A or B?", "Is it X or Y?", "Did they choose option A or option B?"
          \item \textit{Note:} The two options must be explicitly stated in the question text.
        \end{itemize}
        \item \textbf{``other''} -- Everything else.
        \begin{itemize}[label={-}, leftmargin=*, topsep=0pt, itemsep=0pt]
          \item Questions about specific facts, names, dates, numbers, values.
          \item Questions about people, organizations, locations.
          \item Questions asking "who", "what", "when", "where", "how many", etc.
          \item Any question that would require looking up specific information.
        \end{itemize}
      \end{enumerate}
    }
  \end{tcolorbox}

  \begin{tcolorbox}[psec-output]
    {\small\textbf{\textcolor{snowstar!60!black}{Thinking Process}}}
    {\small
      Think step by step:
      \begin{enumerate}[leftmargin=*, topsep=4pt, itemsep=2pt]
        \item Does the answer have to be Yes or No? $\to$ \textbf{"yes\_no"}
        \item Does the question explicitly list 2 choices to pick from? $\to$ \textbf{"binary\_choice"}
        \item Otherwise $\to$ \textbf{"other"}
      \end{enumerate}
    }
  \end{tcolorbox}

  \begin{tcolorbox}[psec-input]
    {\small\textbf{\textcolor{snowpurple}{Input Task}}}
    \vspace{4pt}
    {\small
      Question: \texttt{\{question\}}\\[2pt]
      Gold answer: \texttt{\{answer\}}\\[4pt]
      \textbf{Classify this question:}
    }
  \end{tcolorbox}
\end{tcolorbox}
\caption{\textbf{Question classification prompt.} Used to categorize correctly guessed answers into yes/no, binary choice, or memorization to distinguish random chance from training data contamination.}
\label{prompt:question-classification}
\end{prompt}

Before attributing model performance to document understanding, we must establish what fraction of answers can be ``guessed'' without any document evidence. This baseline captures two phenomena: (1) random chance on constrained answer types, and (2) training data contamination where models have memorized facts from public documents seen during pre-training.

We prompted six frontier models to answer all test questions \emph{without} showing any document text or images (Prompt~\ref{prompt:guessing}). The prompt explicitly instructed models to guess even when uncertain, selecting randomly for yes/no or binary choice questions. We then classified each correctly guessed answer into three categories using GPT-5-mini (Prompt~\ref{prompt:question-classification}):

\begin{itemize}
    \item \textit{Yes/No.} Questions with binary yes/no answers (50\% random baseline)
    \item \textit{Other Binary Choice.} Questions explicitly listing two options, e.g., ``Which is larger: A or B?'' (50\% random baseline)  
    \item \textit{Memorization.} All other questions requiring recall of specific facts, names, dates, or values
\end{itemize}

\begin{table}
\centering\footnotesize
\caption{\textbf{Answer guessability analysis across frontier models.} ``Correct'' shows questions answered correctly without evidence. Categories show the breakdown of correct guesses by question type. ``Random'' estimates the expected contribution from 50\% chance on yes/no and binary questions; ``Contam.'' shows pure memorization questions (facts, names, values) as percentage of test set.
}
\label{tab:guessability}
\begin{tabular}{lcccccc}
\toprule
\textbf{Model} & \textbf{Correct} & \textbf{Yes/No} & \textbf{Binary} & \textbf{Memorized} & \textbf{Random} & \textbf{Contam.} \\
\midrule
Claude Haiku 4.5   & 45 (9.1\%)  & 16 (35.6\%) & 16 (35.6\%) & 13 (28.9\%) & 3.2\% & 2.6\% \\
Claude Sonnet 4.5  & 53 (10.7\%) & 16 (30.2\%) & 13 (24.5\%) & 24 (45.3\%) & 2.9\% & 4.8\% \\
Gemini 2.5 Flash   & 48 (9.7\%)  & 13 (27.1\%) & 14 (29.2\%) & 21 (43.8\%) & 2.7\% & 4.2\% \\
Gemini 2.5 Pro     & 65 (13.1\%) & 15 (23.1\%) & 17 (26.2\%) & 33 (50.8\%) & 3.2\% & 6.7\% \\
GPT-5              & 75 (15.2\%) & 19 (25.3\%) & 15 (20.0\%) & 41 (54.7\%) & 3.4\% & 8.3\% \\
GPT-5 Mini         & 47 (9.5\%)  & 14 (29.8\%) & 13 (27.7\%) & 20 (42.6\%) & 2.7\% & 4.0\% \\
\midrule
\textbf{Average}   & 55.5 (11.2\%) & --- & --- & --- & 3.0\% & 5.1\% \\
\bottomrule
\end{tabular}
\end{table}

As shown in Table~\ref{tab:guessability}, approximately 3\% of questions would be answered correctly by chance alone, stemming from 50\% accuracy on yes/no and binary choice questions. However, the actual correct rate on these question types exceeds 50\%, indicating that models also \emph{know} many yes/no and binary answers from training data.

The ``Memorized'' column shows questions requiring specific factual recall (names, dates, values). However, total contamination is higher: some yes/no and binary questions are also answered from knowledge, not chance. We estimate total knowledge-based answering at approximately 8\% (total guessability minus random baseline), with GPT-5 exhibiting the highest contamination at 11.8\% (15.2\% total $-$ 3.4\% random).

The 9--15\% guessability (varying by model) represents an upper bound on ``free'' performance---answers obtainable without document understanding. Decomposing this: $\sim$3\% from random chance on constrained answer types, and $\sim$8\% from training data contamination. When models achieve 85\%+ accuracy with document evidence, the additional 70+ percentage points represent genuine document comprehension beyond prior knowledge.

\paragraph{Examples of Memorized Facts.} The following questions were answered correctly by GPT-5 purely from training data:
\begin{itemize}
    \item ``The 17,082 metric tons of SO2 emissions reported by TVA in 2020 were 99\% below the peak level from which year?'' $\rightarrow$ {1977}
    \item ``In what year was Chicken Annie's Original established?'' $\rightarrow$ {1934}
    \item ``Who was the Special Inspector General for Afghanistan Reconstruction in 2014?'' $\rightarrow$ {John F. Sopko}
    \item ``What is the fee that the U.S. Department of Treasury charges for debt collection?'' $\rightarrow$ {30\%}
    \item ``What does MOTCA stand for (Afghan Ministries)?'' $\rightarrow$ {Ministry of Transport and Civil Aviation}
\end{itemize}

These examples demonstrate that public documents (government reports, court filings, annual reports) in our benchmark were likely indexed on the web and incorporated into model training data.

\subsection{Visual Perception}
\label{sec:visual_modality_taxonomy}

To validate Property~6 (Visual) and understand what types of visual comprehension our benchmark requires, we developed a taxonomy for classifying questions based on whether visual modality is beneficial for answering them, assuming the correct evidence pages have been retrieved. Our taxonomy comprises five primary categories, each representing a distinct type of visual requirement:

\begin{enumerate}
    \item {\textit{Free Text.}} Answer is present in free-flowing narrative text. Clean OCR extraction would be sufficient; no table, form, or layout comprehension required.
    
    \item {\textit{Tabular.}} Requires understanding tabular relationships—correlating rows with columns, comparing values across rows, or aggregating data from multiple cells. The key requirement is comprehending \emph{table relationships}, not merely reading structured text.
    
    \item {\textit{Structured.}} Requires navigating structured forms (permits, licenses, certificates, applications) to locate specific fields. Distinguished from tables by the label-value pair structure rather than row-column correlation.
    
    \item {\textit{Other Spatial.}} Requires understanding spatial positioning on the page (e.g., ``at the bottom of the page,'' ``in the header''). Applied only when position matters beyond what table/form structure captures.
    
    \item {\textit{Artifacts.}} Visual artifacts not captured above, including: handwritten text, signatures, checkbox/tick mark status, stamps and seals, charts and diagrams, figures and images.
\end{enumerate}

We employed \textit{Gemini 3 Flash} to classify each question (Prompt~\ref{prompt:modality-classifier}), providing the model with the question text, gold-standard answer, and rendered images of all gold evidence pages. The prompt instructed the model to determine which visual modalities would be beneficial for a human or AI to answer the question correctly. Each question receives a primary category and may receive secondary categories when multiple visual modalities are genuinely required (e.g., a question requiring both table comprehension and reading a handwritten annotation)

Table~\ref{tab:visual_modality} presents the distribution across our taxonomy. Notably, fewer than half (42.8\%) of questions can be answered from unstructured free-text alone, while the majority require some form of visual or structural comprehension.

\begin{table}
\centering\footnotesize
\caption{Visual modality requirements for answering \benchmarknameshort{} questions. Categories are mutually exclusive for primary assignment; some questions have secondary categories when multiple modalities apply.}
\label{tab:visual_modality}
\begin{tabular}{lrr}
\toprule
\textbf{Category} & \textbf{Count} & \textbf{\%} \\
\midrule
{Free Text} & 959 & 42.8 \\
{Structured} & 644 & 28.8 \\
{Tabular} & 471 & 21.0 \\
{Artifacts} & 150 & 6.7 \\
\quad{(of which: charts/diagrams)} & \emph{41} & \emph{1.8} \\
\quad{(handwriting/signatures)} & \emph{92} & \emph{4.1} \\
\quad{(images/photos)} & \emph{6} & \emph{0.3} \\
{Other Spatial} & 16 & 0.7 \\
\midrule
\textbf{Requires Visual/Structure} & \textbf{1,281} & \textbf{57.2} \\
\bottomrule
\end{tabular}
\end{table}

While most questions (87.7\%) require only a single visual modality, 12.3\% benefit from multiple. The most common combination is \textit{Structured} + \textit{Other Visual} (7.0\%), typically occurring when forms contain handwritten entries or checkbox fields. Only 1.5\% of questions require both \textit{Tabular} and \textit{Structured}, confirming that tables and forms represent distinct visual challenges.

These findings have several implications for system design:

\begin{itemize}
    \item \textit{Structure is pervasive.} Over 57\% of questions require understanding document structure, yet this structure is often \emph{implicit} in the visual layout rather than explicit in the text.
    
    \item \textit{Visual encoders are not strictly necessary.} PDF-to-markdown conversion (e.g., via document understanding models) represents a viable alternative to end-to-end visual encoding, though with potential information loss. Our benchmark enables measuring these trade-offs.
    
    \item \textit{Text is structured, not free-form.} While documents are highly text-intensive, the text is rarely unstructured. Models must either explicitly comprehend visual layout or successfully infer structural relationships from linearized text—a capability that varies significantly across architectures.
    
    \item \textit{Retrieval vs.\ reading.} This analysis isolates the \emph{reading} component by providing gold evidence pages. The impact of visual capabilities on \emph{retrieval} accuracy represents a complementary research direction, as visual document representations may help identify relevant pages that text-based retrieval misses.
\end{itemize}

This taxonomy and analysis complement the text-vs-image ablation study (Figure ~\ref{fig:visual_necessity}), which measures \emph{performance} differences, by providing a \emph{structural} understanding of why visual modality matters for specific question types.


\begin{prompt}[h]
\begin{tcolorbox}[promptbox, title={\small\textbf{Question Modality Classifier Prompt}}]
  \begin{tcolorbox}[psec-role]
    \vspace{4pt}
    {\small You are an expert at analyzing document VQA (Visual Question Answering) questions to determine whether visual modality is required or beneficial for answering them.}
  \end{tcolorbox}

  \begin{tcolorbox}[psec-steps]
    {\small\textbf{\textcolor{snowdark}{Category Definitions}}}
    {\small
      Classify the question into one or more categories based on these precise definitions:
      \begin{enumerate}[leftmargin=*, topsep=4pt, itemsep=6pt]
        \item \textbf{FREE\_TEXT} -- Answer is present in free-flowing text.
        \begin{itemize}[label={-}, leftmargin=*, topsep=0pt, itemsep=0pt]
          \item Answer is extracted directly from running/narrative text.
          \item No reference to tables, charts, forms, or visual elements.
          \item No need to understand document layout or spatial relationships.
          \item Linear reading comprehension; clean OCR text would be sufficient.
        \end{itemize}
        \item \textbf{TABLE\_STRUCTURE} -- Requires understanding TABLE relationships (rows/columns).
        \begin{itemize}[label={-}, leftmargin=*, topsep=0pt, itemsep=0pt]
          \item Answer requires correlating rows and columns in a TABLE.
          \item Need to understand which values belong to which headers.
          \item Requires comparing multiple rows (finding max/min, aggregating data).
          \item \textbf{NOT for forms:} Forms go to \texttt{STRUCTURED\_DATA\_EXTRACTION}.
        \end{itemize}
        \item \textbf{CHART\_DIAGRAM} -- Requires visual chart/diagram interpretation.
        \begin{itemize}[label={-}, leftmargin=*, topsep=0pt, itemsep=0pt]
          \item Data encoded visually: bar heights, line trends, pie slices, diagrams.
          \item Requires counting visual elements or extracting values from visual scales.
          \item Visual relationships not captured in text.
        \end{itemize}
        \item \textbf{TEXT\_FORMATTING} -- Requires seeing text formatting ONLY.
        \begin{itemize}[label={-}, leftmargin=*, topsep=0pt, itemsep=0pt]
          \item Identifying bold, italic, underlined, colored, or highlighted text.
          \item Font size or style differences (headings vs body).
          \item \textbf{NOT for:} handwriting, signatures, checkboxes $\to$ use \texttt{OTHER\_VISUAL}.
        \end{itemize}
        \item \textbf{IMAGE\_PHOTO} -- Requires actual image/photo content.
        \begin{itemize}[label={-}, leftmargin=*, topsep=0pt, itemsep=0pt]
          \item Asks about photo descriptions, captions, or content visible in photos.
          \item References specific images/figures containing visual information.
        \end{itemize}
        \item \textbf{SPATIAL\_LAYOUT} -- Requires spatial/layout understanding NOT covered by others.
        \begin{itemize}[label={-}, leftmargin=*, topsep=0pt, itemsep=0pt]
          \item \textbf{ONLY use if:} Location matters (top/bottom, left/right, margins, header/footer).
          \item \textbf{Special cases only:} "at the bottom of page", "in the sidebar".
          \item \textbf{NOT for tables/forms:} Use \texttt{TABLE\_STRUCTURE} or \texttt{STRUCTURED\_DATA...}
        \end{itemize}
        \item \textbf{STRUCTURED\_DATA\_EXTRACTION} -- Extraction from structured FORMS.
        \begin{itemize}[label={-}, leftmargin=*, topsep=0pt, itemsep=0pt]
          \item Source is a structured FORM (permit, license, application, financial form).
          \item Visual structure helps navigate the form and find the right field.
          \item Answer is typically a specific value from ONE form field (label:value pairs).
          \item \textbf{NOT for tables:} If comparing rows/columns, use \texttt{TABLE\_STRUCTURE}.
        \end{itemize}
        \item \textbf{OTHER\_VISUAL} -- Visual artifacts not covered above.
        \begin{itemize}[label={-}, leftmargin=*, topsep=0pt, itemsep=0pt]
          \item Handwritten text, signatures, checkboxes/tick marks.
          \item Stamps, seals, logos, mathematical notation, special symbols.
        \end{itemize}
      \end{enumerate}
    }
  \end{tcolorbox}
\end{tcolorbox}
\caption{\textbf{Question modality classifier prompt.} Classifies questions by visual requirements to quantify the importance of layout understanding, table comprehension, and visual artifacts.}
\label{prompt:modality-classifier}
\end{prompt}

\begin{prompt}[h]
\ContinuedFloat
\begin{tcolorbox}[promptbox-notitle]
  \begin{tcolorbox}[psec-input]
    {\small\textbf{\textcolor{snowpurple}{Input}}}
    \vspace{4pt}
    {\small
      Question: \texttt{\{question\}}\\[2pt]
      Answer: \texttt{\{answer\}}
      \vspace{4pt}
      \textbf{The gold evidence page images are provided below.} Examine them to determine whether visual modality is needed to answer the question.
    }
  \end{tcolorbox}

  \begin{tcolorbox}[psec-warning]
    {\small\textbf{\textcolor{valencia!60!black}{Guidance \& Priority}}}
    {\small
      \textbf{Multi-label Guidance:}
      \begin{itemize}[label={-}, leftmargin=*, topsep=2pt, itemsep=0pt]
        \item \textbf{Most questions should have ONLY ONE category} (the primary visual requirement).
        \item Only assign multiple categories if the question TRULY requires combining different modalities.
        \item \textbf{Table vs Form:} \texttt{TABLE\_STRUCTURE} + \texttt{STRUCTURED\_DATA} is extremely rare (requires both correlating rows AND extracting form fields).
        \item \textbf{Spatial:} Only use if position matters AND it is not covered by table/form structure.
      \end{itemize}
      \vspace{4pt}
      \textbf{Selection Priority:}
      \begin{enumerate}[leftmargin=*, topsep=2pt, itemsep=0pt]
        \item If it's a table with rows/columns to correlate $\to$ \textbf{TABLE\_STRUCTURE}.
        \item If it's a form with label:value pairs $\to$ \textbf{STRUCTURED\_DATA\_EXTRACTION}.
        \item If spatial position matters beyond structure $\to$ add \textbf{SPATIAL\_LAYOUT} (rare).
      \end{enumerate}
    }
  \end{tcolorbox}

  \begin{tcolorbox}[psec-role]
    {\small\textbf{\textcolor{snowgray}{Instructions}}}
    {\small
      \begin{itemize}[label={-}, leftmargin=*, topsep=4pt, itemsep=0pt]
        \item Judge based on whether visual modality would actually help answer the question.
        \item Choose the ONE primary visual requirement in most cases.
        \item Do not rely on keyword presence - analyze the actual task.
        \item Consider whether clean OCR text would be sufficient.
      \end{itemize}
    }
  \end{tcolorbox}

  \begin{tcolorbox}[psec-output]
    {\small\textbf{\textcolor{snowstar!60!black}{Output Schema}}}
    \vspace{4pt}
    {\small Respond with a JSON object:}
    \small
\begin{verbatim}
{
  "primary_category": "the single most important category",
  "all_categories": ["list", "of", "ALL", "applicable", "categories"],
  "confidence": 0.0-1.0,
  "reasoning": "brief explanation covering all assigned categories"
}
\end{verbatim}
  \end{tcolorbox}
\end{tcolorbox}
\caption[]{\textbf{Question modality classifier prompt.} \textit{(Continued)}}
\end{prompt}

%% file: appendix/metrics.tex
\clearpage
\section{Formal Definition of Metrics}
\subsection{Accuracy Metric}
\label{app:anls_star}
We evaluate answer accuracy using a hybrid metric that combines string-based matching with LLM-based semantic judgment, designed to be both strict on factual correctness and flexible regarding valid textual variations.

When answer yields no exact match in any of the alternative gold standard answers, we invoke an LLM judge to assess semantic correctness (Prompt~\ref{prompt:llm-judge}). Following the G-Eval framework \cite{liu2023geval}, the judge performs chain-of-thought evaluation across five criteria:
\begin{enumerate}
    \item \textit{Refusal detection.} Does the answer refuse or claim inability to respond?
    \item \textit{Content matching.} Does the core meaning match any ground truth variant?
    \item \textit{Critical errors.} Missing unit qualifiers, wrong entities, or type mismatches?
    \item \textit{Format issues.} List structure presented as comma-separated string?
    \item \textit{Verbosity.} Excessive explanation beyond direct answer?
\end{enumerate}

The judge outputs a three-way classification: \textit{Correct} (score = 1.0), semantically equivalent to ground truth; \textit{Partial} (score = 0.5), correct content with format/verbosity issues; \textit{Incorrect} (score = 0.0), wrong content, missing answer, or critical errors.
We use Gemini 2.5 Flash with function calling to ensure structured output.

\paragraph{Handling Multiple Valid Answers.}
Questions may have multiple valid ground truth representations. Let $\mathcal{G} = \{G^{(1)}, \dots, G^{(k)}\}$ be valid alternatives. The final score is: $
    \text{Score}(P) = \max_{k} \left( \text{Metric}(P, G^{(k)}) \right)
$
where Metric applies our two-stage evaluation pipeline.

\begin{prompt}[!ph]
\begin{tcolorbox}[promptbox, title={\small\textbf{LLM Judge Prompt}}]
  \begin{tcolorbox}[psec-role]
    \vspace{4pt}
    {\small You are evaluating answer correctness for a Document QA benchmark.}
  \end{tcolorbox}

  \begin{tcolorbox}[psec-input]
    {\small\textbf{\textcolor{snowpurple}{Input}}}
    \vspace{4pt}
    {\small
      Question: \texttt{\{question\}}\\[2pt]
      Predicted Answer: \texttt{\{predicted\}}\\[2pt]
      Gold Answer Variants: \texttt{\{gold\_variants\}}
    }
  \end{tcolorbox}

  \begin{tcolorbox}[psec-role]
    {\small\textbf{\textcolor{snowgray}{Evaluation Criteria}}}
    \vspace{4pt}
    {\small Evaluate the predicted answer based on the following definitions:}

    \begin{tcolorbox}[psec-correct]
      {\small\textbf{\textcolor{star!60!black}{Correct}}}
      \vspace{2pt}
      {\small Predicted answer is semantically equivalent to at least one gold variant. Minor format differences are acceptable.}
    \end{tcolorbox}

    \begin{tcolorbox}[psec-warning]
      {\small\textbf{\textcolor{valencia!60!black}{Partial}}}
      \vspace{2pt}
      {\small Predicted answer contains correct core information but has a significant format issue (e.g., list presented as comma-separated string when items are short/atomic) OR includes irrelevant additions.}
    \end{tcolorbox}

    \begin{tcolorbox}[psec-error]
      {\small\textbf{\textcolor{codepurple}{Incorrect}}}
      \vspace{2pt}
      {\small Predicted answer is factually wrong, missing, contains different information, or fails to answer the question type (e.g., no Yes/No for binary questions). Missing unit qualifiers that change magnitude (thousands, millions) are incorrect.}
    \end{tcolorbox}
  \end{tcolorbox}

  \begin{tcolorbox}[psec-steps]
    {\small\textbf{\textcolor{snowdark}{Evaluation Steps}}}
    \vspace{4pt}
    {\small Follow these steps in order:}
    {\small
      \begin{enumerate}[leftmargin=*, itemsep=2pt, topsep=4pt]
        \item \textbf{Check for refusal:} Does the answer refuse or claim inability to answer? If yes $\to$ incorrect.
        \item \textbf{Compare content:} Does the predicted answer match the core meaning of any gold variant? If content is wrong or different $\to$ incorrect.
        \item \textbf{Check critical errors:} (any of these $\to$ incorrect)
        \begin{itemize}[label={-}, leftmargin=*, topsep=0pt]
          \item Missing scale qualifiers (e.g., "50" vs "\$50 million").
          \item Binary questions without explicit Yes/No.
          \item Wrong entity/value (different person, company, number).
          \item Partial list with wrong items mixed in.
        \end{itemize}
        \item \textbf{Check format:} (only if content is correct)
        \begin{itemize}[label={-}, leftmargin=*, topsep=0pt]
          \item Gold expects list but Predicted is string $\to$ partial.
        \end{itemize}
        \item \textbf{Check verbosity:} (only if content is correct)
        \begin{itemize}[label={-}, leftmargin=*, topsep=0pt]
          \item \textbf{Correct:} Extra qualifiers, relevant context, clarifying phrases.
          \item \textbf{Partial:} Unrequested details, over-specific precision.
          \item \textbf{Incorrect:} Multi-sentence responses, full paragraphs, conversational preambles.
        \end{itemize}
      \end{enumerate}
    }
  \end{tcolorbox}

  \begin{tcolorbox}[psec-role]
    {\small\textbf{\textcolor{snowgray}{Instructions}}}
    \vspace{4pt}
    {\small
      Based on your step-by-step analysis, provide your final judgment.
      \vspace{4pt}
      After your reasoning, you MUST call \texttt{submit\_judgment} with your final decision.
    }
  \end{tcolorbox}
\end{tcolorbox}
\caption{\textbf{LLM judge prompt} for semantic answer correctness evaluation. The judge performs chain-of-thought assessment across five criteria: refusal detection, content matching, critical errors, format issues, and verbosity.}
\label{prompt:llm-judge}
\end{prompt}

The evaluation prompt was developed iteratively: starting from a vanilla LLM judge, we conducted multiple rounds of human review on 100 stratified samples, identifying systematic disagreements and refining the prompt to address edge cases—particularly around list formatting, acceptable verbosity levels, and unit qualifier handling. This calibration process improved human-LLM agreement from 82\% to 90\% on held-out samples before final specificity/sensitivity measurement.


\paragraph{Bias Correction.} Following \citet{lee2025llmjudge}, we apply Rogan-Gladen correction to adjust for LLM judge bias. Based on a 200-sample human evaluation, we measured sensitivity $q_1 = 0.980$ (probability the LLM judges correct when human judges correct) and specificity $q_0 = 1.000$ (probability the LLM judges incorrect when human judges incorrect). The bias-adjusted score is computed as:
\begin{equation}
    \hat{\theta} = \frac{\hat{p} + q_0 - 1}{q_0 + q_1 - 1}
\end{equation}
where $\hat{p}$ is the raw LLM judgment score. Confidence intervals account for both the test sample variance and calibration uncertainty.

\subsection{Retrieval and Attribution Metrics}
\label{app:retrieval_metrics}

To quantify the agent's ability to locate and attribute relevant information sources, we employ the F1 score evaluated at two distinct levels of granularity: Page-level and Document-level.

For a given question $i$, let $\mathcal{R}_i$ denote the set of unique evidence units (pages or documents) cited by the agent in its final response, and let $\mathcal{G}_i$ denote the minimal set of ground truth evidence units required to answer the question.

We first calculate the Precision ($\text{P}_i$) and Recall ($\text{R}_i$) for the instance:

\begin{equation}
    \text{P}_i = \frac{|\mathcal{R}_i \cap \mathcal{G}_i|}{|\mathcal{R}_i|}
    \quad , \quad
    \text{R}_i = \frac{|\mathcal{R}_i \cap \mathcal{G}_i|}{|\mathcal{G}_i|}
\end{equation}

We handle edge cases as follows:
\begin{itemize}
    \item If $|\mathcal{R}_i| = 0$ (the agent cites nothing), then $\text{P}_i = 0$ and $\text{F1}_i = 0$.
    \item Since our benchmark guarantees solvability, $|\mathcal{G}_i| \ge 1$ for all valid questions.
\end{itemize}

The F1 score for the $i$-th question is the harmonic mean of precision and recall:

\begin{equation}
    \text{F1}_i = 
    \begin{cases} 
    2 \cdot \frac{\text{P}_i \cdot \text{R}_i}{\text{P}_i + \text{R}_i} & \text{if } (\text{P}_i + \text{R}_i) > 0 \\
    0 & \text{otherwise}
    \end{cases}
\end{equation}

The final benchmark score is reported as the arithmetic mean over all $N$ test samples:
\begin{equation}
    \text{Score} = \frac{1}{N} \sum_{i=1}^{N} \text{F1}_i
\end{equation}

We apply this formulation at two levels to diagnose specific bottlenecks in the agentic workflow:

\paragraph{\pagef{}.}
The elements of the sets $\mathcal{R}_i$ and $\mathcal{G}_i$ are unique page identifiers defined as tuples $(d, p)$, where $d$ is the document ID and $p$ is the page index. This metric enforces strict grounding: an agent is penalized if it correctly identifies the document but cites the wrong page (e.g., citing the Table of Contents instead of the specific clause on page 42).

\paragraph{\docf{}.}
The elements of the sets are reduced to unique document identifiers $d$. This metric evaluates the retrieval system's ability to locate the correct file within the corpus, disregarding intra-document navigation errors. Comparing \docf{} against \pagef{} allows us to isolate ``last-mile'' navigation failures from fundamental retrieval failures.

\subsection{Efficiency and Calibration Metrics}
\label{app:calibration_metrics}

To evaluate whether agentic systems effectively allocate computational resources (steps) in proportion to problem difficulty, we employ a non-parametric calibration metric based on the Cumulative Difference method \cite{kloumann2024cumulativedifferencespairedsamples}. Let the evaluation set consist of $N$ samples. For each sample $i$, we observe a tuple $(s_i, y_i)$, where:
\begin{itemize}
    \item $s_i \in \mathbb{N}$ represents the \textit{effort}, quantified as the number of discrete steps (e.g., tool calls, search actions) taken by the agent.
    \item $y_i \in \{0, 1\}$ represents the \textit{outcome}, defined as the binary correctness of the final answer (based on Accuracy thresholding).
\end{itemize}

Let $\bar{y} = \frac{1}{N} \sum_{i=1}^{N} y_i$ denote the global average accuracy of the agent on the benchmark.

We first order the test samples by ascending effort. Let $\pi$ be a permutation of indices $\{1, \dots, N\}$ such that the step counts are non-decreasing:
\begin{equation}
    s_{\pi(1)} \le s_{\pi(2)} \le \dots \le s_{\pi(N)}
\end{equation}

The Cumulative Difference sequence, $D = (D_0, D_1, \dots, D_N)$, measures the accumulating deviation of the agent's performance from its mean performance as effort increases. It is defined recursively:
\begin{align}
    D_0 &= 0 \\
    D_k &= \sum_{j=1}^{k} (y_{\pi(j)} - \bar{y}) \quad \text{for } k=1, \dots, N
\end{align}

The trajectory of $D_k$ reveals conditional performance regimes. Positive slope indicates a sub-population where the local accuracy exceeds the global mean $\bar{y}$. This typically occurs in low-step regimes for well-calibrated agents (easy problems are solved efficiently). Negative slope indicates a sub-population where local accuracy is below $\bar{y}$. This typically occurs in high-step regimes, where the agent expends significant effort without achieving success.

To summarize the calibration quality into a scalar metric, we utilize the \kuiper{} statistic $K$. This statistic measures the total range of the cumulative deviations, capturing the severity of non-uniform performance across the effort spectrum.

\begin{equation}
    K = \max_{0 \le k \le N} (D_k) - \min_{0 \le k \le N} (D_k)
\end{equation}

A lower $K$ value indicates ``effort-invariant'' performance, where the probability of correctness $P(y=1 | s)$ is independent of the number of steps taken. This implies the agent is equally reliable whether it takes 1 step or 20. A high $K$ value indicates distinct regimes of systematic over-performance and under-performance. For agentic systems, this usually manifests as a sharp decline in accuracy as step counts increase, highlighting an inability to recover from initial errors.

\subsection{Sensitivity of Kuiper Statistic to Effort Measure}
\label{app:effort-sensitivity}

We investigate whether the choice of effort metric materially affects the Kuiper calibration statistic. For each system, we compute Kuiper under four effort definitions where available: (1) steps/calls---the number of tool calls or LLM invocations, (2) tokens (total)---sum of input and output tokens, (3) tokens (generated)---output tokens only, and (4) execution time.

Table~\ref{tab:effort-correlation} reports pairwise Spearman rank correlations between effort measures. For models achieving $\geq 60\%$ accuracy, correlations between steps and token measures range from $\rho = 0.72$ to $\rho = 0.95$, indicating that sample orderings are largely preserved across definitions.

\begin{table}[h]
\centering
\caption{Spearman rank correlation ($\rho$) between effort measures across systems. Higher values indicate more similar sample orderings.}
\label{tab:effort-correlation}
\small
\begin{tabular}{lcccc}
\toprule
\textbf{System} & \textbf{Acc.} & \textbf{Steps vs Total} & \textbf{Steps vs Gen.} & \textbf{Total vs Gen.} \\
\midrule
BM25 GPT-5.2 & 67.8\% & 0.76 & 0.95 & 0.78 \\
BM25 Claude Sonnet & 80.6\% & 0.89 & 0.92 & 0.90 \\
BM25 GPT-4.1 & 60.0\% & 0.79 & 0.85 & 0.72 \\
RLM GPT-5.2 & 64.2\% & 0.83 & 0.74 & 0.80 \\
RLM Gemini-3-Pro & 73.8\% & 0.84 & 0.93 & 0.87 \\
RLM Claude Sonnet & 70.5\% & 0.91 & 0.85 & 0.89 \\
\bottomrule
\end{tabular}
\end{table}

Table~\ref{tab:kuiper-by-effort} shows the resulting Kuiper statistics under each effort definition. For the six systems above, Kuiper values vary by at most 20\% across metrics, confirming that the calibration assessment is robust to effort definition choice.

\begin{table}[h]
\centering
\caption{Kuiper statistic under different effort definitions. Missing entries (--) indicate unavailable measurements.}
\label{tab:kuiper-by-effort}
\small
\begin{tabular}{lccccc}
\toprule
\textbf{System} & \textbf{Acc.} & \textbf{Steps} & \textbf{Tokens (Total)} & \textbf{Tokens (Gen.)} & \textbf{Time} \\
\midrule
BM25 GPT-5.2 & 67.8\% & 64.1 & 53.3 & 59.9 & -- \\
BM25 Claude Sonnet & 80.6\% & 35.4 & 36.4 & 37.8 & -- \\
BM25 GPT-4.1 & 60.0\% & 42.0 & 35.9 & 36.2 & -- \\
RLM GPT-5.2 & 64.2\% & 28.4 & 31.2 & 21.1 & 22.3 \\
RLM Gemini-3-Pro & 73.8\% & 22.3 & 20.9 & 25.1 & 21.2 \\
RLM Claude Sonnet & 70.5\% & 42.7 & 38.1 & 34.3 & 30.8 \\
Human & 82.2 \% & 14.6 & -- & -- & 12.5 \\
\bottomrule
\end{tabular}
\end{table}

%% file: appendix/baselines-details.tex
\clearpage
\section{Baseline Implementation Details}

\subsection{BM25 MLLM Agent.}
\label{appendix:search-agent-baseline}

We implement a search-augmented agent baseline that combines text-based retrieval with vision-language model (VLM) reasoning. The agent iteratively searches a document collection and analyzes retrieved page images to answer questions.

We first construct a full-text search index from OCR-extracted text. Each document page is indexed with its source filename, page number, and OCR content using the Whoosh search library. The index supports boolean queries (AND, OR, NOT), phrase matching, and wildcard patterns.

The agent operates in an iterative loop, equipped with a \texttt{search\_documents} tool that accepts natural language queries and returns rendered images of the top-$k$ matching pages. The agent receives these images directly (not OCR text), enabling it to leverage the VLM's visual understanding capabilities for layout-sensitive documents such as tables, forms, and figures. Tool is defined using JSON Schema and passed to the model alongside the system prompt with the following description:

\begin{quote}
    \small
    Search document collection and return images of
    matching pages. Supports: terms and phrases (use quotes for
    exact match), boolean operators (AND, OR, NOT - AND is default),
    wildcards (* for multiple chars, ? for single char). Examples:
    'engine specifications', '"Bell 407" AND accessories',
    'Bell*', 'incorporation NOT date'.
\end{quote}

The agent produces structured outputs containing: (1) a list of answer strings, and (2) citations specifying the exact source file and page number for each piece of evidence. On the final iteration, the model is forced to provide an answer via constrained decoding (OpenAI) or tool forcing (Anthropic).

\begin{algorithm}[h]
\caption{Search Agent for Document QA}
\label{alg:search-agent}
\begin{algorithmic}[1]
\Require Question $q$, search index $\mathcal{I}$, VLM $\mathcal{M}$, max iterations $T$, top-$k$
\Ensure Answer $a$, citations $C$
\State $\textit{messages} \gets [\textsc{SystemPrompt}, q]$
\For{$t = 1$ to $T$}
    \If{$t = T$}
        \State $\textit{response} \gets \mathcal{M}(\textit{messages}, \textit{force\_answer}=\text{True})$
    \Else
        \State $\textit{response} \gets \mathcal{M}(\textit{messages}, \textit{tools}=[\texttt{search}, \texttt{answer}])$
    \EndIf
    \If{\textit{response} is \texttt{answer} tool call}
        \State \Return $\textit{response.answer}, \textit{response.citations}$
    \EndIf
    \If{\textit{response} is \texttt{search} tool call}
        \State $\textit{query} \gets \textit{response.query}$
        \State $\textit{results} \gets \textsc{Search}(\mathcal{I}, \textit{query}, k)$ \Comment{Returns (file, page) tuples}
        \State $\textit{images} \gets [\textsc{RenderPage}(f, p) \text{ for } (f, p) \in \textit{results}]$
        \State Append tool result and $\textit{images}$ to $\textit{messages}$
    \EndIf
\EndFor
\State \Return fallback answer from final response
\end{algorithmic}
\end{algorithm}

Default hyperparameters are $T=10$ maximum iterations and $k=5$ results per search. Document pages are rendered as PNG images at high resolution; for Anthropic's API, images exceeding the 5MB limit are progressively downscaled using Lanczos resampling. The system prompt is provided in Prompt~\ref{prompt:bm25-agent}.


\begin{prompt}[h]
\begin{tcolorbox}[promptbox, title={\small\textbf{BM25 MLLM Agent Prompt}}]
  \begin{tcolorbox}[psec-role]
    \vspace{4pt}
    {\small You are a document QA assistant with access to a search tool. The search tool returns images of document pages.}
  \end{tcolorbox}

  \begin{tcolorbox}[psec-warning]
    {\small\textbf{\textcolor{valencia!60!black}{Search Instructions}}}
    \vspace{4pt}
    {\small The answer is definitely in the documents. If search returns no results, try different terms (synonyms, abbreviations, rephrasing).}
  \end{tcolorbox}

  \begin{tcolorbox}[psec-output]
    {\small\textbf{\textcolor{snowstar!60!black}{Analysis \& Output}}}
    \vspace{4pt}
    {\small Once relevant pages are found, analyze images and provide:}
    {\small
      \begin{enumerate}[leftmargin=*, topsep=4pt, itemsep=4pt]
        \item \textbf{answer}: List of answer values.
        \begin{itemize}[label={-}, leftmargin=*, topsep=0pt, itemsep=0pt]
          \item Single answer $\to$ one-element list.
          \item Multiple items/entities $\to$ several elements.
          \item Use as few words as possible (exact document words preferred).
          \item Do not write full sentences.
        \end{itemize}
        \item \textbf{citations}: List of sources. Each citation must have:
        \begin{itemize}[label={-}, leftmargin=*, topsep=0pt, itemsep=0pt]
          \item \texttt{file}: Exact PDF filename shown in image (e.g., "1007969.pdf").
          \item \texttt{page}: The page number (integer).
        \end{itemize}
      \end{enumerate}
    }
  \end{tcolorbox}
\end{tcolorbox}
\caption{\textbf{BM25 MLLM Agent system prompt.} The agent receives this prompt alongside the tool definition for document search.}
\label{prompt:bm25-agent}
\end{prompt}

\subsection{Managed RAG Services}\label{app:managed}

We evaluate two proprietary RAG-as-a-Service offerings that abstract the document processing pipeline, allowing us to benchmark against industry-standard solutions.

\paragraph{Gemini File Search.}
Google's File Search API provides end-to-end managed retrieval. We upload all 796 PDF documents to a \emph{file search store}, where Google's infrastructure automatically handles chunking, embedding generation, and vector indexing. At query time, the Gemini model receives access to this store as a tool. Unlike agentic approaches with explicit search--read--reason loops, File Search operates as a \emph{single-shot} retrieval mechanism: one API call triggers retrieval, context injection, and answer generation atomically. The system returns grounding metadata indicating which document chunks were retrieved, though the internal retrieval strategy (query reformulation, re-ranking, chunk selection) remains opaque. We prompt the model to output answers in a structured format with explicit page-level citations, which we extract for evaluation.

\paragraph{OpenAI Assistants File Search.}
OpenAI's Assistants API provides a similar managed RAG capability through its \texttt{file\_search} tool. We upload documents to a \emph{vector store}, which handles chunking and embedding using OpenAI's proprietary models. An Assistant configured with the \texttt{file\_search} tool automatically retrieves relevant chunks when processing queries. The system operates via a threads/runs abstraction: each question creates a conversation thread, triggers a run, and returns results with inline citations. We prompt the model to include page-level citations in its structured output when identifiable from context.

For both services, we provide the complete document collection without any task-specific fine-tuning or prompt engineering beyond basic answer formatting instructions (Prompt~\ref{prompt:managed-rag}). This establishes a fair comparison point representing what practitioners would obtain ``out of the box'' from these commercial offerings.

\begin{prompt}[h]
\begin{tcolorbox}[promptbox, breakable, title={\small\textbf{Managed RAG Prompt}}]
  \begin{tcolorbox}[psec-role]
    \vspace{4pt}
    {\small You are a document QA assistant with access to a file search tool. The file search tool retrieves relevant content from a collection of PDF documents.}
  \end{tcolorbox}

  \begin{tcolorbox}[psec-warning]
    {\small\textbf{\textcolor{valencia!60!black}{Important}}}
    \vspace{2pt}
    {\small The answer to the question is definitely in the documents. Search carefully using different terms if needed.}
  \end{tcolorbox}

  \begin{tcolorbox}[psec-output]
    {\small\textbf{\textcolor{snowstar!60!black}{Output Format}}}
    \vspace{4pt}
    {\small When you have the answer, respond with a JSON object in this exact format:}
    \small
\begin{verbatim}
{
  "answer": ["answer value 1", "answer value 2", ...],
  "citations": [
    {"file": "exact_filename.pdf", "page": 1},
    {"file": "another_file.pdf", "page": 3}
  ]
}
\end{verbatim}
    \vspace{2pt}
    {\small \textbf{Where:}}
    \begin{itemize}[label={-}, leftmargin=*, topsep=0pt, itemsep=0pt]
      \item {\small \texttt{answer}: list of answer values (use as few words as possible, exact document wording preferred)}
      \item {\small \texttt{citations}: list of sources with exact PDF filename and page number}
    \end{itemize}
  \end{tcolorbox}
\end{tcolorbox}
\caption{\textbf{Managed RAG prompt} used with both Gemini File Search and OpenAI Assistants File Search services.}
\label{prompt:managed-rag}
\end{prompt}

\subsection{HEAVEN}
\label{app:heaven}


We use the original encoder configurations by \citet{kim2025hybrid}: DSE (\texttt{MrLight/dse-qwen2-2b-mrl-v1}), and ColQwen2.5 (\texttt{vidore/colqwen2.5-v0.2}). Document Layout Analysis uses DocLayout-YOLO for title region extraction. All hyperparameters follow paper defaults. The VLM prompt is shown in Prompt~\ref{prompt:heaven}.

\begin{table}
\caption{HEAVEN hyperparameters (paper defaults).}
\centering
\small
\begin{tabular}{lcc}
\toprule
\textbf{Parameter} & \textbf{Symbol} & \textbf{Value} \\
\midrule
Stage 1 candidates & $k_1$ & 200 \\
Final results & $k_2$ & 5 \\
VS-page reduction factor & $r$ & 15 \\
VS-page filter ratio & -- & 0.5 \\
Query filter ratio & $\rho$ & 0.25 \\
Stage 1 weight & $\alpha$ & 0.1 \\
Stage 2 weight & $\beta$ & 0.3 \\
\bottomrule
\end{tabular}
\label{tab:heaven-params}
\end{table}


\begin{prompt}[h]
\begin{tcolorbox}[promptbox, title={\small\textbf{HEAVEN VLM Prompt}}]
  \begin{tcolorbox}[psec-role]
    \vspace{4pt}
    {\small You are a document QA assistant. Answer the question based ONLY on the provided document images.}
  \end{tcolorbox}

  \begin{tcolorbox}[psec-input]
    {\small\textbf{\textcolor{snowpurple}{Input}}}
    \vspace{4pt}
    {\small
      Document sources: \texttt{\{context\}}\\[2pt]
      Question: \texttt{\{question\}}
    }
  \end{tcolorbox}

  \begin{tcolorbox}[psec-steps]
    {\small\textbf{\textcolor{snowdark}{Instructions}}}
    {\small
      \begin{enumerate}[leftmargin=*, itemsep=2pt, topsep=4pt]
        \item Carefully examine ALL provided document images.
        \item Find the specific information that answers the question.
        \item Respond with a JSON object in the exact format defined below.
      \end{enumerate}
    }
  \end{tcolorbox}

  \begin{tcolorbox}[psec-output, capture=minipage]
    {\small\textbf{\textcolor{snowstar!60!black}{Output Schema}}}
    \small
\begin{verbatim}
{
  "answer": ["answer value 1", "answer value 2", ...],
  "citations": [
    {"file": "exact_filename.pdf", "page": 1},
    {"file": "another_file.pdf", "page": 3}
  ]
}
\end{verbatim}
    \vspace{2pt}
    {\small \textbf{Where:}}
    \begin{itemize}[label={-}, leftmargin=*, topsep=0pt, itemsep=0pt]
      \item {\small \texttt{answer}: list of answer values (use as few words as possible, exact document wording preferred)}
      \item {\small \texttt{citations}: list of sources with exact PDF filename and page number (1-indexed)}
    \end{itemize}
  \end{tcolorbox}

  \begin{tcolorbox}[psec-warning]
    {\small\textbf{\textcolor{valencia!60!black}{Important}}}
    \vspace{2pt}
    {\small The answer to the question is definitely in the documents. Examine all provided pages carefully.}
  \end{tcolorbox}
\end{tcolorbox}
\caption{\textbf{HEAVEN VLM prompt.} After hybrid retrieval returns top-$k$ pages, this prompt instructs the VLM to extract answers with citations.}
\label{prompt:heaven}
\end{prompt}

\subsection{Claude Agent with Semtools}
\label{app:semtools}

We implement an agentic baseline using the Claude Agents SDK integrated with 
\texttt{semtools}, a suite of CLI tools designed for semantic document processing. The agent has access to three composable Unix-style utilities:

\begin{enumerate}
    \item \textit{parse}: Converts non-text formats (e.g., PDF) into Markdown 
    using LlamaParse as the backend. Outputs file paths to \texttt{stdout}, enabling pipeline composition. 
    \item \textit{search}: Performs semantic keyword search over text files using static 
    embeddings. Supports configurable context windows (\texttt{--n-lines}), top-$k$ retrieval 
    (\texttt{--top-k}), and distance thresholding (\texttt{--max-distance}). The tool operates on 
    \texttt{stdin} or explicit file lists, returning matching passages with surrounding context.
    
    \item \textit{workspace}: Manages persistent embedding caches for document collections. 
    Once a workspace is activated (\texttt{export SEMTOOLS\_WORKSPACE=name}).
\end{enumerate}

The agent uses Claude's \texttt{claude\_code} system prompt preset, granting it bash execution capabilities. 
Tool permissions allow \texttt{Bash}, 
\texttt{Read}, \texttt{Glob}, \texttt{Grep}, and \texttt{Search} operations. The agent produces 
structured JSON output containing the answer, page-level citations, and the search history.

This paradigm allows flexible, iterative exploration but relies on the agent's ability to formulate 
effective bash pipelines and interpret unstructured search results. The user prompt and CLAUDE.md configuration are shown in Prompt~\ref{prompt:claude-agent}.


\begin{prompt}[h]
\begin{tcolorbox}[promptbox, title={\small\textbf{Claude Agent User Prompt}}]
  \begin{tcolorbox}[psec-input]
    \vspace{4pt}
    {\small
      Try to answer the question based on the PDFs at \texttt{\{PDF\_PATH\}}.
      All of them have been converted to Markdown and can be found at \texttt{\{MARKDOWN\_PATH\}}.
      \vspace{4pt}\\
      \textbf{Question:} \texttt{\{question\}}
      \vspace{4pt}\\
      Provide the answer along with citations (file names and page number) and the search queries you used.
    }
  \end{tcolorbox}
\end{tcolorbox}

\begin{tcolorbox}[promptbox, title={\small\textbf{Claude Agent CLAUDE.md}}]
  \begin{tcolorbox}[psec-role]
    {\small\textbf{\textcolor{snowgray}{Environment Setup}}}
    \vspace{4pt}
    {\small Use \texttt{uv} alongside the virtual environment when running Python scripts. When requiring environment variables, use:}
    \small
\begin{verbatim}
uv run --env-file keys.env main.py
\end{verbatim}
  \end{tcolorbox}

  \begin{tcolorbox}[psec-steps]
    {\small\textbf{\textcolor{snowdark}{Augmented CLI Tooling}}}
    {\small
      If executing bash commands, you have three very helpful utilities installed:
      \begin{itemize}[label={-}, leftmargin=*, topsep=4pt, itemsep=2pt]
        \item \textbf{parse}: Converts any non-grep-able format into markdown. Outputs a filepath for a converted markdown file for every input file to stdin.
        \item \textbf{search}: Performs a search using static embeddings (similar to grep). Works best with keyword-based queries. Only works with text-based files (requires \texttt{parse} first for PDFs).
        \item \textbf{workspace}: Management for accelerating search over large collections.
      \end{itemize}
    }
    \vspace{4pt}
    {\small\textbf{1. Parse CLI}}
    \small
\begin{verbatim}
parse [OPTIONS] <FILES>...
Options:
  -c, --parse-config <PATH>  Path to config file
  -b, --backend <BACKEND>    Backend type (default: llama-parse)
\end{verbatim}
    \vspace{4pt}
    {\small\textbf{2. Search CLI}}
    \small
\begin{verbatim}
search [OPTIONS] <QUERY> [FILES]...
Options:
  -n, --n-lines <N>       Lines context [default: 3] (Suggest: 30-50)
  --top-k <K>             Top-k files/texts [default: 3]
  -m, --max-distance <D>  Return results below distance threshold
  -i, --ignore-case       Case-insensitive search
\end{verbatim}
    \vspace{4pt}
    {\small\textbf{3. Workspace CLI}}
    \small
\begin{verbatim}
workspace <COMMAND>
Commands:
  use     Use or create a workspace
  status  Show active workspace and stats
  prune   Remove stale files
\end{verbatim}
  \end{tcolorbox}

  \begin{tcolorbox}[psec-role]
    {\small\textbf{\textcolor{snowgray}{Common Usage Patterns}}}
    \small
\begin{verbatim}
# Parse a PDF and search for specific content
parse document.pdf | xargs cat | search "error handling"

# Search with custom context and thresholds
search "machine learning" *.txt --n-lines 5 --max-distance 0.3

# Create/Use a workspace (caches embeddings)
workspace use my-workspace
export SEMTOOLS_WORKSPACE=my-workspace
search "some keywords" ./large_dir/*.txt --n-lines 5 --top-k 10
\end{verbatim}
  \end{tcolorbox}
\end{tcolorbox}
\caption{\textbf{Claude Agent user prompt and CLAUDE.md configuration.} The agent uses the \texttt{claude\_code} system prompt preset with access to Unix-style semantic search tools.}
\label{prompt:claude-agent}
\end{prompt}

\begin{prompt}[h]
\ContinuedFloat 
\begin{tcolorbox}[promptbox-notitle]
  \begin{tcolorbox}[psec-warning]
    {\small\textbf{\textcolor{valencia!60!black}{Evaluation Rules}}}\vspace{2pt}\\
    {\small Verbose answers score POORLY. Answer format rules:\vspace{-6pt}\\
      \begin{itemize}[label={-}, leftmargin=*, topsep=0pt, itemsep=0pt]
        \item answer: A list of SHORT strings with ONLY essential info.
        \item NO full sentences. NO explanations.
        \item Always return a list, even for single answers (e.g., \texttt{["\$1.2M"]}).
        \item Use EXACT words from the document when possible.
        \item NEVER include phrases like "According to...", "The document states...".
      \end{itemize}
    }
  \end{tcolorbox}

  \begin{tcolorbox}[psec-steps]
    {\small\textbf{\textcolor{snowdark}{Examples: Correct vs Incorrect}}\vspace{2pt}\\}
    {\small
      Q: Federal housing laws protect people from discrimination on the basis of what?\vspace{2pt}\\
      \textcolor{codepurple}{\textbf{[WRONG]}} Federal housing laws protect people from discrimination on the basis of race, color...\\
      \textcolor{green!60!black}{\textbf{[CORRECT]}} \texttt{["race", "color", "religion", ...]}\vspace{2pt}\\
      Q: What was the total live weight of sheep?\vspace{1pt}\\
      \textcolor{codepurple}{\textbf{[WRONG]}} The total live weight ... was 2,400 cwt.\\
      \textcolor{green!60!black}{\textbf{[CORRECT]}} \texttt{["2,400 cwt"]}
    }
  \end{tcolorbox}

  \begin{tcolorbox}[psec-output]
    {\small\textbf{\textcolor{snowstar!60!black}{Expected Output Format}}}
    {\small You only need to return the "answer" and "citations".}
    \small
\begin{verbatim}
{
  "id": "test/0",
  "question": "What is the total revenue?",
  "answer": ["$1.2M"],
  "citations": [
    {"document": "report.pdf", "page": 5}
  ],
  "search_history": ["query1", "query2"]
}
\end{verbatim}
  \end{tcolorbox}
\end{tcolorbox}
\caption[]{\textbf{Claude Agent user prompt and CLAUDE.md configuration.} \textit{(Continued)}}
\end{prompt}

\subsection{Recursive Language Models}\label{app:rlm}

Recursive Language Models (RLMs)~\cite{zhang2025recursivelanguagemodels} represent a task-agnostic inference paradigm that enables language models to handle arbitrarily long contexts by \emph{programmatically} examining, decomposing, and recursively calling themselves over the input. Unlike traditional retrieval-augmented generation approaches that rely on fixed chunking and embedding strategies, RLMs offload the entire document corpus as a variable within a REPL (Read-Eval-Print Loop) environment that the model can interact with through code execution.

The RLM framework replaces the canonical \texttt{llm.completion(prompt)} call with an \texttt{rlm.completion(prompt)} call. When invoked, the system:
\begin{enumerate*}[label=(\arabic*)]
    \item Loads the document corpus into a code-accessible variable with explicit structural markers (e.g., \texttt{[FILE: doc.pdf]}, \texttt{[PAGE N]}).
    \item Provides the LLM with a sandboxed execution environment where it can write and execute code.
    \item Enables the model to launch recursive sub-LM calls via \texttt{llm\_query()} to process subsets of the context.
\end{enumerate*}

Our implementation uses the RLM\footnote{\url{https://github.com/alexzhang13/rlm}} library. The document corpus is sourced from OCR-processed markdown representations of the PDF collection (created with Mistral OCR 3), where page boundaries are preserved using horizontal rule separators (\texttt{---}). We augment the corpus with explicit page markers (\texttt{[PAGE N]} ... \texttt{[END PAGE N]}) to enable accurate citation extraction.

The model receives a prompt containing the full corpus text and the question, with instructions to search the corpus and return both the answer and source locations (Prompt~\ref{prompt:rlm}). The raw RLM response is then processed through a structured output extraction step (using the same underlying model) to normalize answers and citations into a consistent format.


\begin{prompt}[h]
\begin{tcolorbox}[promptbox, title={\small\textbf{RLM Prompt}}]
  \begin{tcolorbox}[psec-role]
    \vspace{4pt}
    {\small You are tasked with answering a query with associated context. You can access, transform, and analyze this context interactively in a REPL environment that can recursively query sub-LLMs. You will be queried iteratively until you provide a final answer.}
  \end{tcolorbox}

  \begin{tcolorbox}[psec-steps]
    {\small\textbf{\textcolor{snowdark}{REPL Environment Initialization}}}
    {\small
      \begin{enumerate}[leftmargin=*, itemsep=2pt, topsep=4pt]
        \item \texttt{context}: Variable containing extremely important information. Check its content to understand what you are working with.
        \item \texttt{llm\_query}: Function to query an LLM (handles ~500K chars) inside the REPL.
        \item \texttt{llm\_query\_batched}: Function to query multiple prompts concurrently. \\
        \textit{Signature:} \texttt{llm\_query\_batched(prompts: List[str]) -> List[str]}
        \item \texttt{print()}: Use statements to view REPL output and continue reasoning.
      \end{enumerate}
    }
  \end{tcolorbox}

  \begin{tcolorbox}[psec-role]
    {\small\textbf{\textcolor{snowgray}{Strategy}}}
    \vspace{4pt}
    {\small You will only see truncated outputs from the REPL. Use \texttt{llm\_query} on variables to analyze semantics. Use variables as buffers to build your answer.}\vspace{1em}\\
    {\small\textbf{\textcolor{snowgray}{Suggested Workflow:}}}\vspace{-3pt}\\
    {\small
      \begin{itemize}[label={-}, leftmargin=*, topsep=0pt]
        \item Look at \texttt{context} and determine a chunking strategy.
        \item Break context into smart chunks.
        \item Query sub-LLMs per chunk and save answers to a buffer.
        \item Query an LLM with all buffers to produce the final answer.
      \end{itemize}
      \vspace{5pt}
      \textit{Note:} Sub-LLMs fit $\sim$500K characters. You can feed $\sim$10 documents per sub-LLM query.
    }
  \end{tcolorbox}

  \begin{tcolorbox}[psec-input]
    {\small\textbf{\textcolor{snowpurple}{Execution Format}}}
    \vspace{4pt}
    {\small Wrap Python code in triple backticks with \texttt{repl} identifier:}
    \small
\begin{verbatim}
```repl
chunk = context[:10000]
answer = llm_query(f"What is the magic number? Chunk: {chunk}")
print(answer)
\end{verbatim}
  \end{tcolorbox}

  \begin{tcolorbox}[psec-error]
    {\small\textbf{\textcolor{codepurple}{Final Answer Protocol}}}
    \vspace{4pt}
    {\small
      When the task is complete, you MUST use one of these functions (do not use code blocks for this):
      \begin{itemize}[label={-}, leftmargin=*, topsep=0pt]
        \item \texttt{FINAL(your final answer here)}
        \item \texttt{FINAL\_VAR(variable\_name)}
      \end{itemize}
      Think step by step. Plan and execute immediately. Explicitly answer the original query in your final output.
    }
  \end{tcolorbox}
\end{tcolorbox}
\caption{\textbf{RLM prompt.} The model receives a REPL environment with the document corpus loaded as a variable and functions to spawn recursive sub-LLM queries.}
\label{prompt:rlm}
\end{prompt}

%% file: appendix/failure-analysis.tex
\clearpage
\section{Extended Results}
\label{app:extended-results}

\subsection{Multi-Hop Question Complexity Analysis}
\label{app:multihop}

\begin{figure}[b]
    \centering \includegraphics[width=0.3\linewidth]{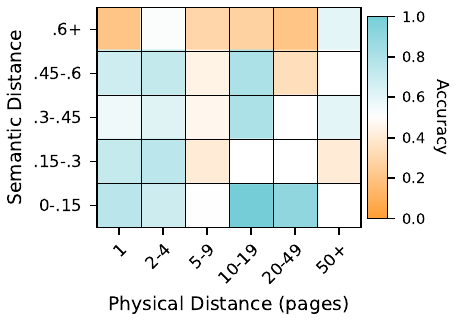}
    \caption{\textbf{Same-document multi-hop accuracy} by physical distance (page gap) and semantic distance (embedding cosine). Semantic distance is a stronger predictor of difficulty.}
    \label{fig:multihop-heatmap}
\end{figure}

Our benchmark contains 372 multi-hop questions requiring evidence integration across multiple pages: 186 same-document (50\%) and 186 cross-document (50\%). For each multi-hop question, we compute:
\begin{enumerate*}[label=(\arabic*)]
    \item \textit{Physical distance.} The page index difference between evidence locations (same-document only).
    \item \textit{Semantic distance.} Cosine distance between page embeddings using Snowflake Arctic Embed (\texttt{snowflake-arctic-embed-m-v2.0}).
\end{enumerate*}
We measure accuracy using five models (Claude 4.5 Sonnet, Claude 4.5 Haiku, Gemini 3.0 Pro, GPT-5, GLM-4.6V).

\paragraph{Same-Document Multi-Hop.}
Table~\ref{tab:multihop-same-doc} shows accuracy by semantic distance. Semantic similarity is a stronger predictor of difficulty than physical page distance ($r$=-0.26 vs $r$=-0.06). Questions with semantically similar evidence achieve 72.4\% accuracy, while dissimilar evidence drops to 34.8\%---a 38 percentage point decline.

\begin{table}[h]
\caption{Same-document multi-hop accuracy by semantic distance between evidence pages.}
\label{tab:multihop-same-doc}
\centering\footnotesize
\begin{tabular}{lcc}
\toprule
\textbf{Semantic Distance} & \textbf{Questions} & \textbf{Accuracy} \\
\midrule
0.0--0.15 (similar) & 29 & 72.4\% \\
0.15--0.3 & 38 & 71.1\% \\
0.3--0.45 & 51 & 56.9\% \\
0.45--0.6 & 45 & 64.0\% \\
0.6+ (dissimilar) & 23 & 34.8\% \\
\midrule
\textbf{Overall} & \textbf{186} & \textbf{61.2\%} \\
\bottomrule
\end{tabular}
\end{table}

\paragraph{Cross-Document Multi-Hop.}
Cross-document questions (n=186) show higher overall accuracy (75.7\%) compared to same-document (61.2\%). The relationship between semantic distance and accuracy is weaker ($r$=-0.09), with accuracy ranging from 72.4\% to 80.0\% across semantic distance bins. This suggests that explicit document boundaries may help models structure the retrieval task.

\begin{table}[h]
\caption{Cross-document multi-hop accuracy by semantic distance between evidence pages.}
\centering\footnotesize
\begin{tabular}{lcc}
\toprule
\textbf{Semantic Distance} & \textbf{Questions} & \textbf{Accuracy} \\
\midrule
0.0--0.15 & 18 & 80.0\% \\
0.15--0.3 & 17 & 80.0\% \\
0.3--0.45 & 26 & 77.7\% \\
0.45--0.6 & 43 & 77.2\% \\
0.6+ & 82 & 72.4\% \\
\midrule
\textbf{Overall} & \textbf{186} & \textbf{75.7\%} \\
\bottomrule
\end{tabular}
\label{tab:multihop-cross-doc}
\end{table}

Counter-intuitively, cross-document multi-hop is easier than same-document multi-hop (75.7\% vs 61.2\%). This may be because cross-document questions often involve explicit comparisons (e.g., ``which event occurs earlier?''), providing clearer task structure, while same-document reasoning requires subtle integration of information across distant sections of a single coherent document.

\subsection{Qualitative Analysis of Human Baseline Errors}
\label{app:human_errors}

During the establishment of the human baseline (Section~\ref{sec:human}), we observed that human participants, despite having access to the same search tools and documents as the agentic systems, occasionally failed to provide correct answers. These instances were not results of data annotation errors, but rather genuine performance failures reflecting the cognitive load and complexity of the benchmark. 

\begin{table*}[h]
\caption{\textbf{Examples of human errors}. Observed during baseline creation. These represent genuine misunderstandings rather than annotation noise, highlighting the reasoning challenges inherent in the benchmark.}
\label{tab:human_errors}
\centering
\footnotesize
\renewcommand{\arraystretch}{1.2}
\begin{tabular}{p{0.3\linewidth} p{0.2\linewidth} p{0.45\linewidth}}
\toprule
\textbf{Question} & \textbf{Error Type} & \textbf{Characterization of Mistake} \\
\midrule
What types of pants are \textbf{not} allowed at MTM? & Negation Blindness & The participant located the correct dress code section but overlooked the negation in the query, listing permitted items rather than prohibited ones. \\
\midrule
Was the Sodium level detected [...] within normal range? & Incomplete Evidence Retrieval & The participant found the specific sodium test result but failed to retrieve the separate page containing the reference ``normal range'' table required for verification. \\
\midrule
Which Forensic Scientist \textbf{approved} the [...] report? & Role Conflation & The participant provided the name of the \textit{Certifying Scientist} who electronically signed the work order, confusing this role with the \textit{Forensic Scientist} mentioned elsewhere in the document chain. \\
\midrule
What water-related \textbf{act} did Pennsylvania introduce...? & Lexical Polysemy & The participant was misled by the polysemy of the word ``act,'' interpreting it as a physical deed (answering that they ``opened a bridge'') rather than a legislative statute. \\
\midrule
\textbf{How long after} public transportation started [...] did deaths decrease? & Temporal Reasoning Failure & The question required calculating a duration (time delta). The participant identified the correct dates but answered with a specific calendar year instead of the elapsed time. \\
\midrule
What is the \textbf{largest} penalty assessed in case 3AN-24-04508CI? & Ordinal Ranking Error & The participant located the correct legal document but reported the second-largest penalty listed, likely due to skimming or a failure to sort the values exhaustively. \\
\bottomrule
\end{tabular}
\end{table*}

These errors underscore the difficulty of the task, specifically regarding attention to detail, cross-page navigation, and semantic precision. Table~\ref{tab:human_errors} categorizes common failure modes observed during the human baseline study.


\subsection{Error Decomposition}\label{subsec:failure-analysis}

\paragraph{Methodology.}
We analyze the BM25 MLLM Agent configuration, which provides the broadest model coverage (17 systems spanning five model families, plus the human baseline with the same search tool).
We decompose every prediction into a four-stage cascade using citation metadata and the paper's LLM judge (Section~\ref{sec:evaluation}) for correctness.
Correctness is determined by the Semantic Accuracy score (LLM judge): a score $\geq 0.5$ counts as correct (with $0.5$ indicating a verbose but semantically valid answer, and $1.0$ a precise match).
For each \emph{incorrect} prediction, the error type is assigned deterministically:
\begin{enumerate}[nosep]
    \item \textbf{Retrieval Failure} --- the system never retrieved the gold document (\docf{} $= 0$).
    \item \textbf{Navigation Failure} --- the gold document was found but the gold page was not (\docf{} $> 0$, \pagef{} $= 0$).
    \item \textbf{Comprehension Failure} --- the correct page was retrieved but the answer is wrong (\pagef{} $> 0$).
    \item \textbf{No Answer / Refusal} --- the system returned an empty response.
\end{enumerate}
This taxonomy requires no subjective LLM classification for the error \emph{type} assignment---only for the correct/incorrect boundary.

\paragraph{Aggregate Error Distribution.}
Across 17 agent systems (8{,}499 predictions, 3{,}273 incorrect), the error breakdown is:

\begin{table}[h]
\centering\footnotesize
\caption{Aggregate error distribution across all agent systems (N=3{,}273 incorrect predictions).}
\label{tab:error-distribution}
\begin{tabular}{lcc}
\toprule
\textbf{Error Type} & \textbf{Count} & \textbf{\%} \\
\midrule
Retrieval (wrong document) & 1{,}167 & 35.7 \\
Comprehension (right page, wrong answer) & 941 & 28.8 \\
Navigation (right doc, wrong page) & 753 & 23.0 \\
No Answer / Refusal & 412 & 12.6 \\
\bottomrule
\end{tabular}
\end{table}

\paragraph{Per-System Failure Profiles.}
Figure~\ref{fig:error-decomposition-main} (main paper) shows the full decomposition across all systems. Key observations:
\begin{itemize}[nosep]
    \item \textit{Claude Sonnet 4.5} has the lowest retrieval failure rate among agents (4.0\%) but the highest comprehension failure rate among top models (8.6\%)---it finds content effectively but misinterprets it.
    \item \textit{Smaller and open-weight models} exhibit the highest retrieval failure rates: \textit{GLM-4.6V Flash} (36.0\%), \textit{Qwen3-VL 8B} (25.2\%), and \textit{Gemini 2.5 Pro} (21.4\%), indicating that search query formulation is a primary bottleneck for less capable agents.
    \item \textit{GPT-4.1 Nano} is dominated by refusals (48.2\% of predictions)---the model abandons search prematurely rather than failing at any specific stage.
    \item The \textit{Human} baseline has the lowest retrieval failure (2.2\%) and navigation failure (4.6\%), but a relatively high comprehension rate (9.4\%), confirming that when humans have the right evidence, occasional attention errors still occur (cf.\ Table~\ref{tab:human_errors}).
\end{itemize}

\begin{figure}[h]
    \centering
    \includegraphics[width=\linewidth]{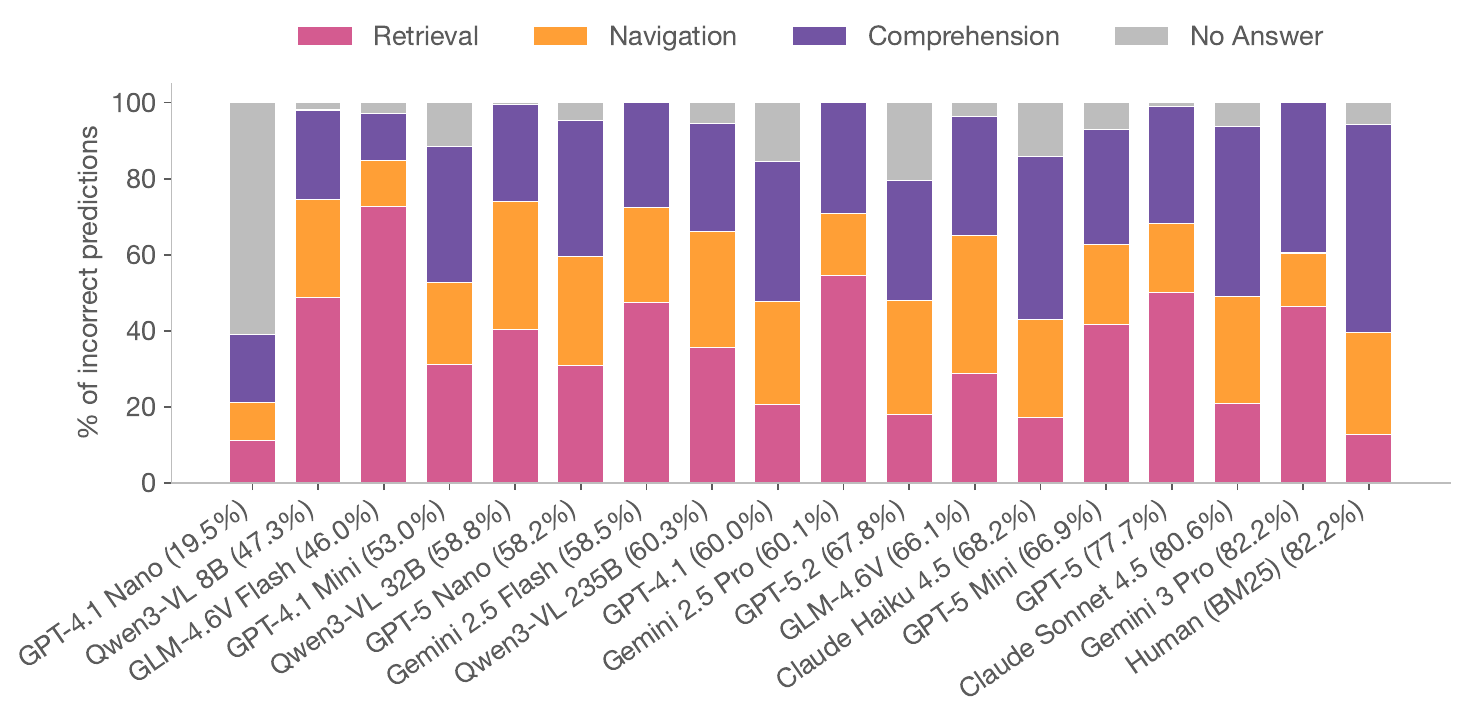}
    \caption{\textbf{Error composition shifts with model capability.} Systems ordered by increasing accuracy (left to right). Weaker models are dominated by refusals; stronger models shift toward retrieval and comprehension failures.}
    \label{fig:error-shift}
\end{figure}

\paragraph{How Error Composition Shifts with Model Capability.}
Figure~\ref{fig:error-shift} shows that as model accuracy increases, the composition of errors changes qualitatively. Weaker models are dominated by refusals (``no answer'')---they give up. As capability increases, retrieval failure becomes the dominant bottleneck. The strongest models have error profiles dominated by comprehension failures, indicating that retrieval is largely solved and the remaining challenge is answer extraction.

\begin{figure}[h]
    \centering
    \includegraphics[width=0.85\linewidth]{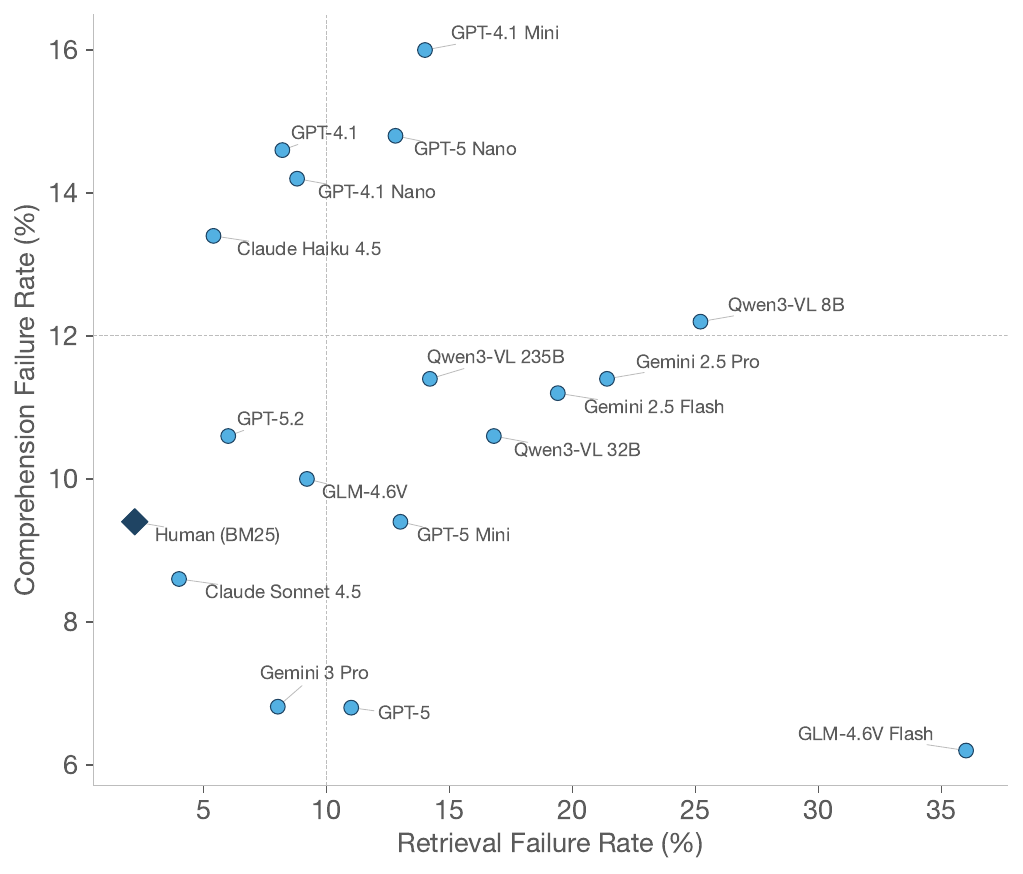}
    \caption{\textbf{Retrieval vs.\ comprehension failure profiles.} Each point is one system. Models in the bottom-left corner fail least on both dimensions. The dashed lines mark approximate cluster boundaries.}
    \label{fig:ret-vs-comp}
\end{figure}

\paragraph{Retrieval vs.\ Comprehension Failure Profiles.}
Figure~\ref{fig:ret-vs-comp} maps each system's retrieval and comprehension failure rates, revealing distinct clusters. The strongest models (Gemini~3~Pro, GPT-5, Claude~Sonnet~4.5) occupy the bottom-left quadrant---low failure rates on both axes. Mid-tier models split into two profiles: \emph{retrieval-limited} systems (Qwen3-VL, GLM-4.6V Flash) that rarely reach the comprehension stage because they fail to find the right document, and \emph{comprehension-limited} systems (GPT-4.1 family, Claude~Haiku~4.5) that retrieve well but struggle with answer extraction. Notably, the human baseline has the lowest retrieval failure rate but non-trivial comprehension failure, consistent with the attention-based errors described in Table~\ref{tab:human_errors}.

\begin{figure}[h]
    \centering
    \includegraphics[width=\linewidth]{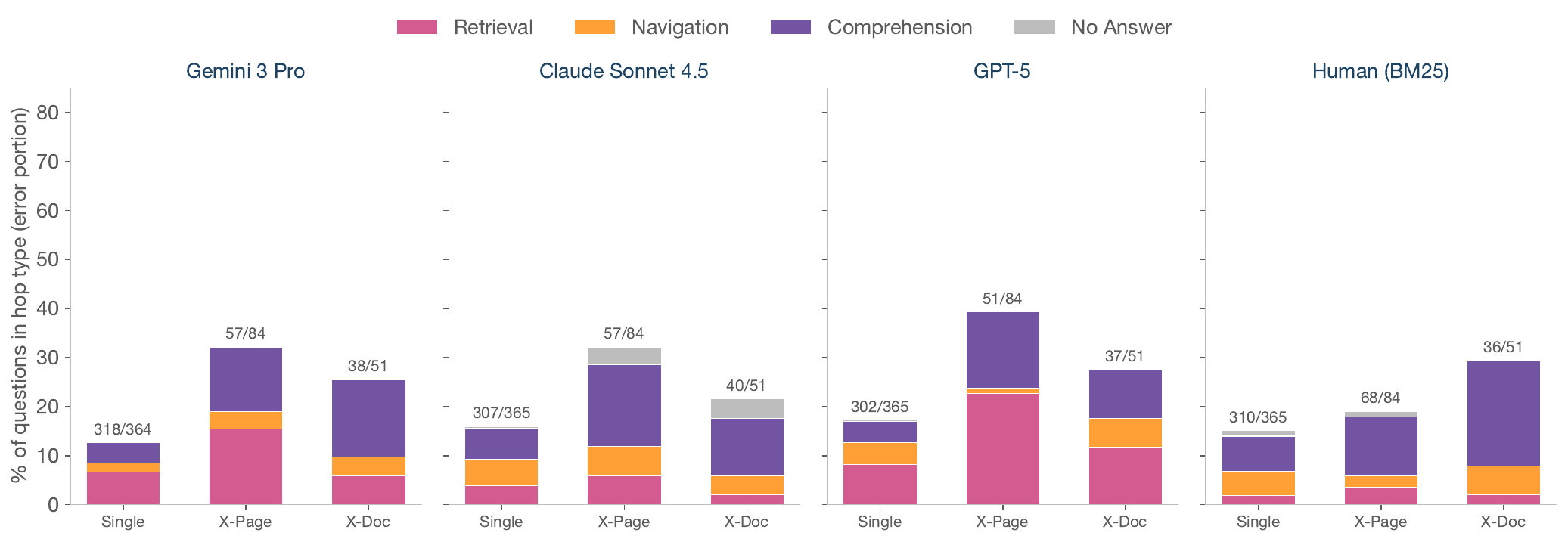}
    \caption{\textbf{Error types by evidence complexity} for top-4 systems. Cross-page (same document) questions are consistently hardest.}
    \label{fig:error-by-hop}
\end{figure}

\paragraph{Error Types by Evidence Complexity.}
Figure~\ref{fig:error-by-hop} breaks down errors by hop type for the top four systems. Cross-page (same-document) questions are hardest, with error rates rising from $\sim$13\% for single-evidence to $\sim$35\% for cross-page. Retrieval failure accounts for a larger share of cross-page errors, suggesting that navigating within long documents is harder than finding separate documents.

\begin{figure}[h]
    \centering
    \includegraphics[width=\linewidth]{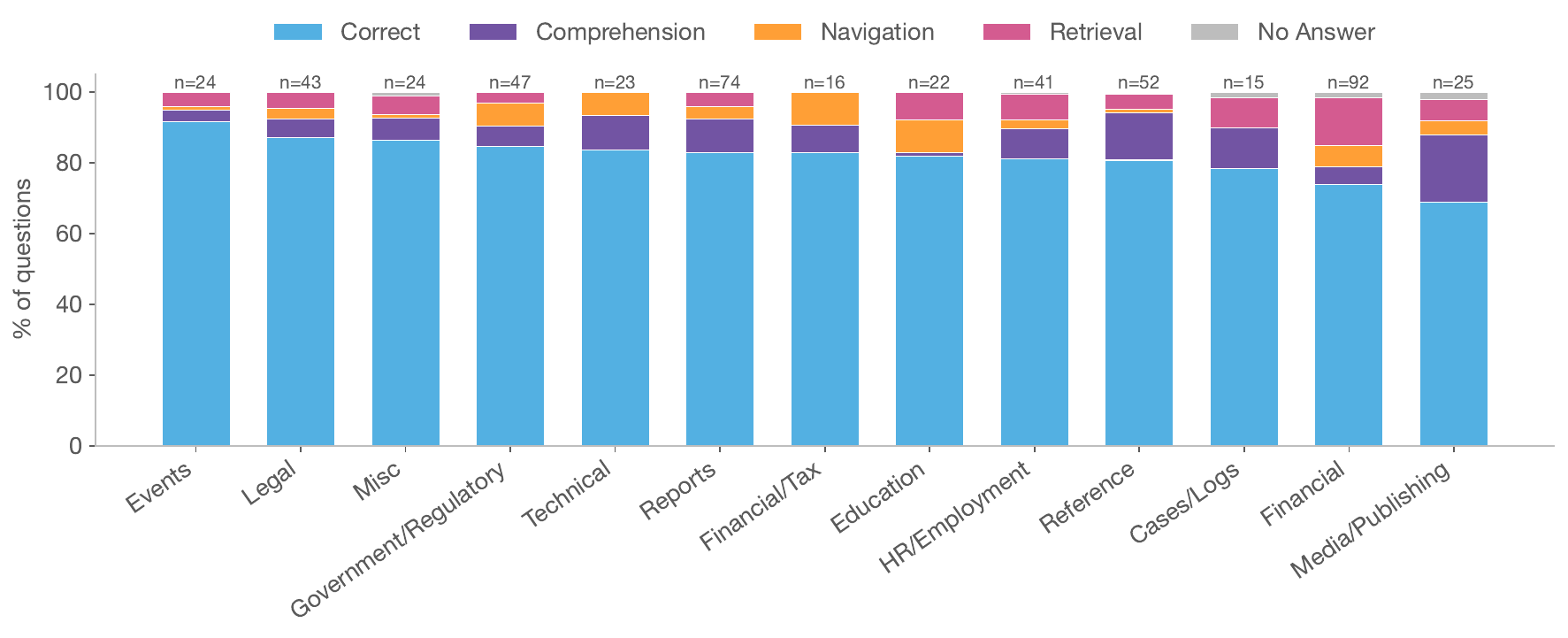}
    \caption{\textbf{Error decomposition by document domain} (top-4 systems aggregated). Media/Publishing and Financial domains are hardest; Events documents are easiest.}
    \label{fig:error-by-domain}
\end{figure}

\paragraph{Error Types by Document Domain.}
Figure~\ref{fig:error-by-domain} aggregates errors by document domain for the top four systems. Event-related documents are easiest (92\% accuracy), likely due to structured formats with clear named entities. Media/Publishing (69\%) and Financial documents (74\%) are hardest, with Financial documents showing the highest retrieval failure rate---consistent with the challenge of navigating dense numerical tables across many pages.

\begin{figure}[h]
    \centering
    \includegraphics[width=\linewidth]{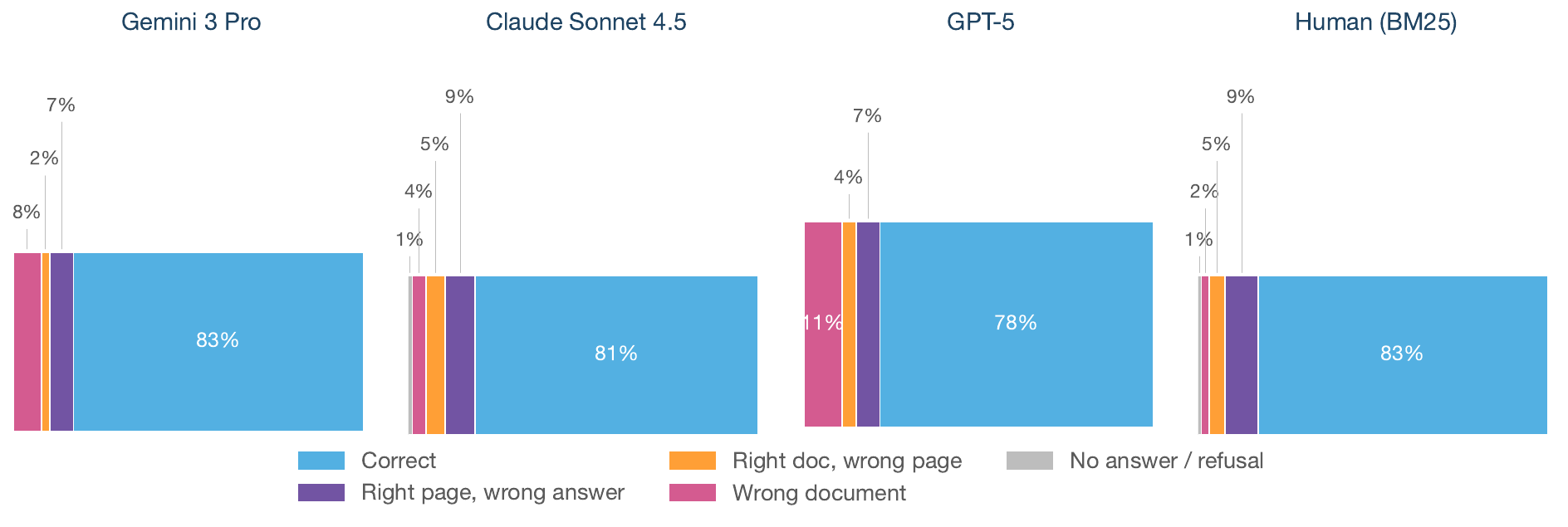}
    \caption{\textbf{Error cascade: where do systems fail?} Each horizontal bar decomposes a system's predictions by error stage. Percentages indicate the share of predictions in each category.}
    \label{fig:cascade-comparison}
\end{figure}

\paragraph{Failure Cascade.}
Figure~\ref{fig:cascade-comparison} presents the failure cascade for four representative systems. Each error stage is shown as a cumulative funnel: total questions $\rightarrow$ document found $\rightarrow$ correct page found $\rightarrow$ correct answer. This visualization makes the progressive loss at each stage immediately visible.

\subsection{Cross-System Item Agreement}\label{subsec:item-agreement}

\paragraph{Methodology.}
For each of the 500 test questions, we build a binary correctness vector across all BM25 MLLM Agent systems and the human baseline. We measure pairwise agreement using Cohen's $\kappa$, which accounts for chance agreement due to marginal accuracy rates.

\begin{figure}[h]
    \centering
    \includegraphics[width=\linewidth]{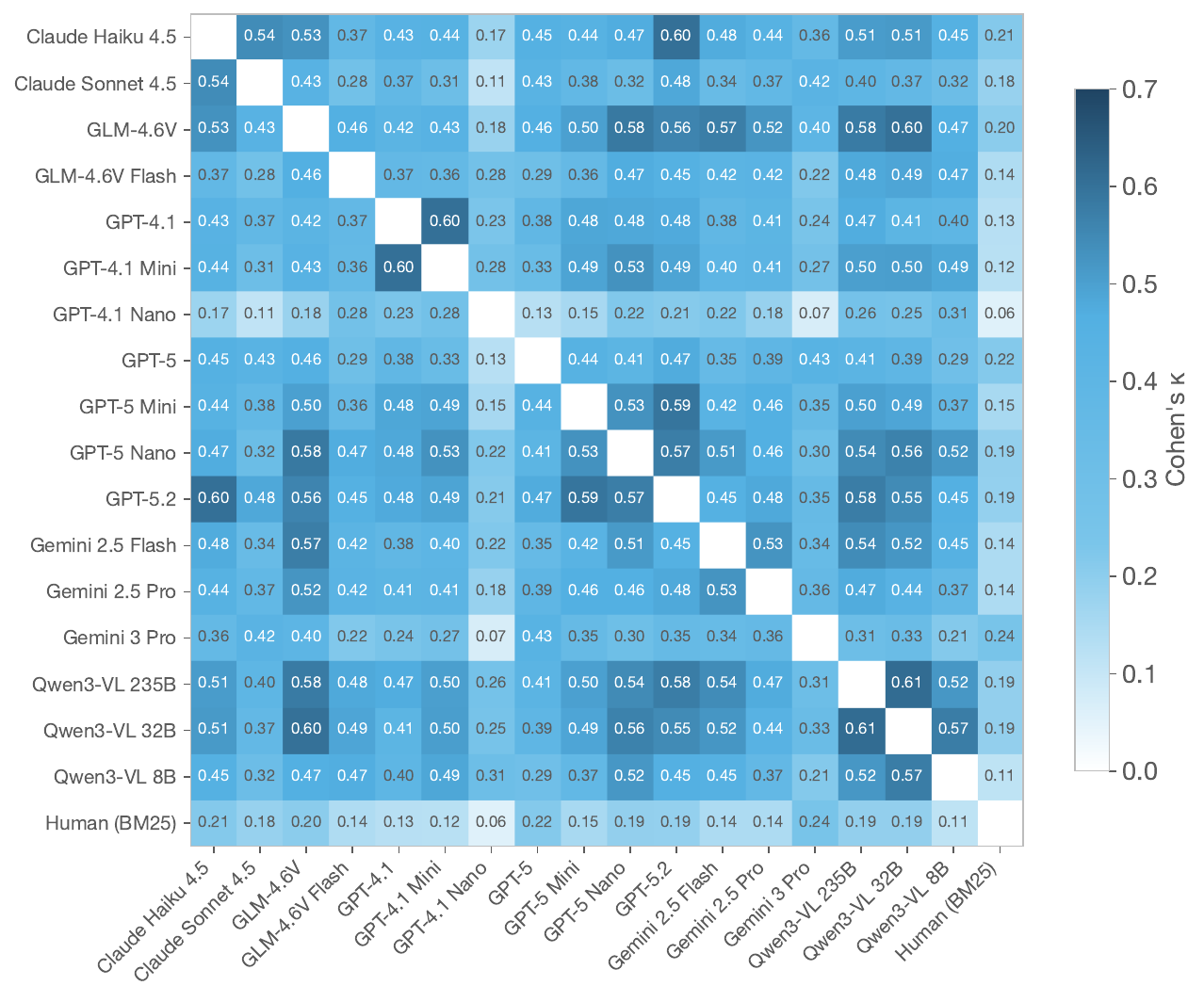}
    \caption{\textbf{Pairwise system agreement (Cohen's~$\kappa$).} Darker cells indicate higher agreement. The Human baseline shows low agreement with all agent systems despite comparable accuracy, revealing complementary error patterns.}
    \label{fig:kappa-heatmap}
\end{figure}

\paragraph{Pairwise Agreement.}
Figure~\ref{fig:kappa-heatmap} shows the full pairwise $\kappa$ matrix. Model--model pairs within similar capability tiers achieve moderate agreement ($\kappa \approx 0.4$--$0.6$), confirming shared systematic strengths and weaknesses. Cross-family agreement is lower (e.g., GPT vs.\ Gemini $\kappa \approx 0.3$--$0.4$), indicating that model families have distinct failure signatures.

The most striking pattern is the \emph{Human row}: agreement between the human baseline and every agent is low ($\kappa = 0.06$--$0.24$), despite similar overall accuracy. For example, Human and Gemini~3~Pro both achieve ${\sim}82\%$ accuracy yet share $\kappa = 0.24$. Of their 107 disagreement items, human-specific failures are dominated by comprehension errors (64\%), while model-specific failures split between retrieval (43\%) and comprehension (43\%). This confirms that humans and agents solve fundamentally different subsets of the benchmark.

\begin{figure}[h]
    \centering
    \includegraphics[width=\linewidth]{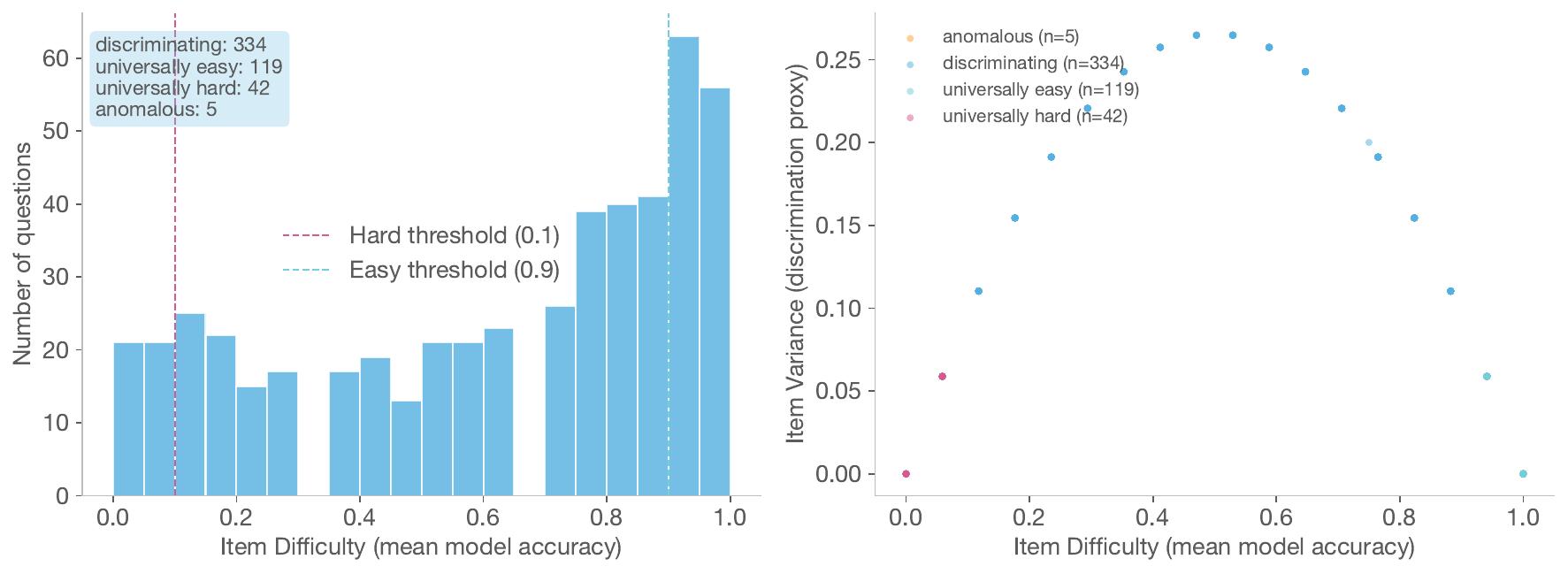}
    \caption{\textbf{Left:} Distribution of item difficulty (mean model accuracy). \textbf{Right:} Difficulty vs.\ variance; high-variance items are the most discriminating. Only 5 anomalous items exist, confirming monotonic difficulty scaling.}
    \label{fig:item-difficulty}
\end{figure}

\paragraph{Item Difficulty Spectrum.}
Across the 500 test items, 23.8\% are universally easy ($>$90\% of models correct), 8.4\% are universally hard ($<$10\% correct), and the remaining 66.8\% are discriminating (Figure~\ref{fig:item-difficulty}). Critically, only 5 items (1.0\%) are \emph{anomalous}---easy for weak models but hard for strong ones---confirming that the benchmark contains no ``trick questions'' and exhibits sound psychometric properties: difficulty increases monotonically with model capability.
Among the 86 items where humans fail, 24 have model difficulty $\geq$70\%, revealing human-specific cognitive limitations (attention fatigue, negation blindness) on questions that most models solve reliably.

\subsection{Search Trajectory Analysis}\label{subsec:trajectory-analysis}

\begin{figure}[h]
    \centering
    \includegraphics[width=\linewidth]{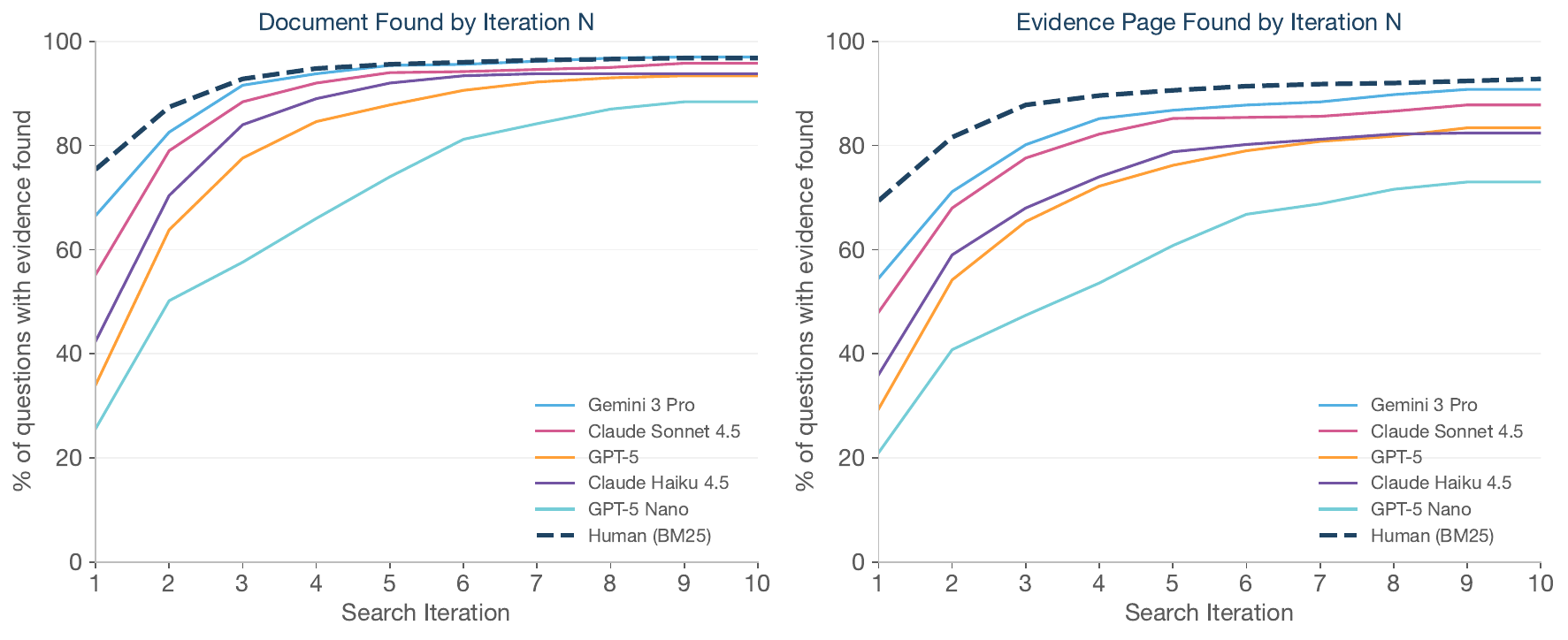}
    \caption{\textbf{Evidence found by iteration~$N$.} Left: document retrieval. Right: page retrieval. Humans find evidence earlier, but top agents converge by iteration 5.}
    \label{fig:survival}
\end{figure}

\paragraph{First-Query Advantage.}
Figure~\ref{fig:survival} plots cumulative document and page retrieval rates by search iteration. Humans find the gold document on their first query ${\sim}80\%$ of the time, whereas the best agent (Gemini 3 Pro) starts at ${\sim}70\%$. The gap narrows by iteration 4--5, but the human first-query advantage persists in page-level retrieval.

\begin{figure}[h]
    \centering
    \includegraphics[width=\linewidth]{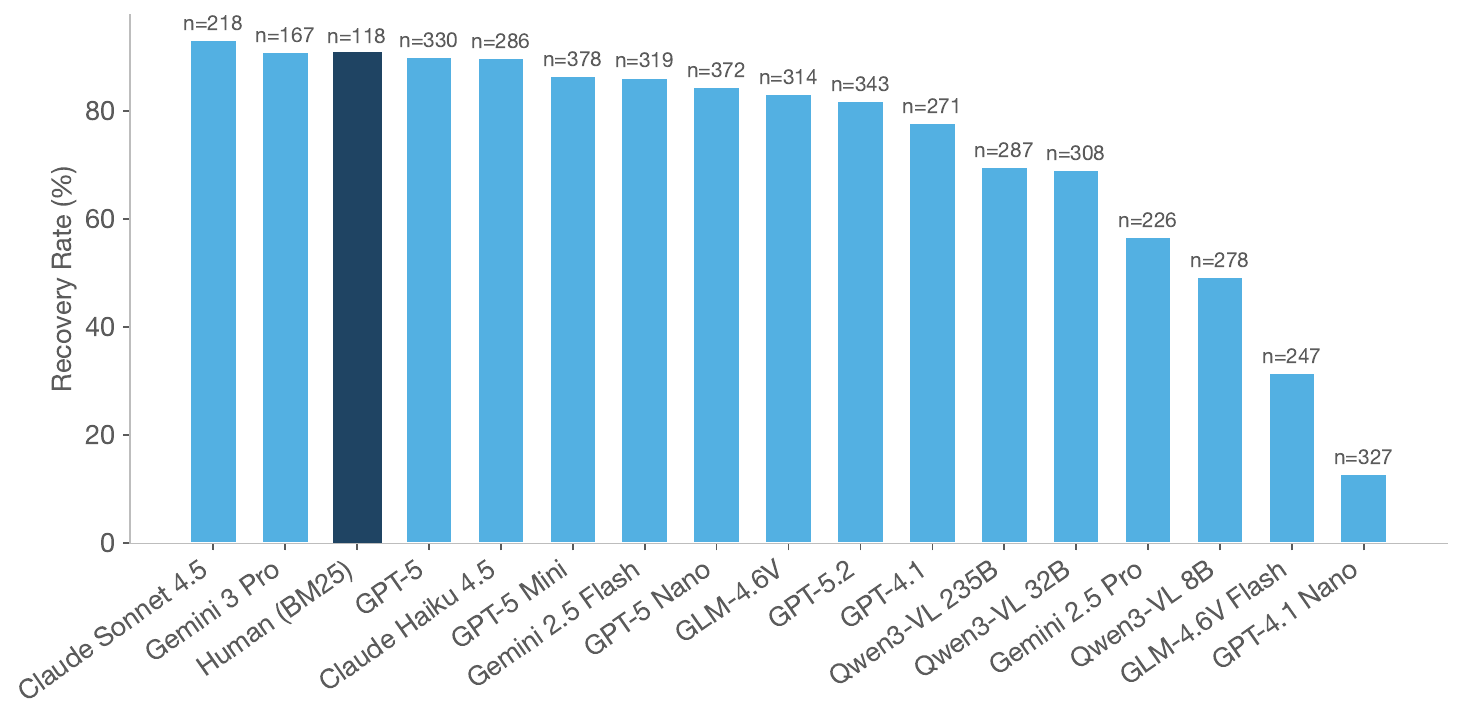}
    \caption{\textbf{Recovery rate after initial query miss.} Percentage of questions where the system eventually finds the gold document after failing on the first query. Labels show the number of questions requiring recovery ($n$). Notably, the human baseline needs recovery for only 118 questions---far fewer than most agents (typically 250--380)---indicating that humans craft more precise initial queries. Among the cases that do require recovery, top agents match human recovery rates (${\sim}93\%$); weaker models rarely recover.}
    \label{fig:recovery-rate}
\end{figure}

\paragraph{Recovery After Initial Failure.}
When the first search query fails to retrieve the gold document, recovery success varies dramatically (Figure~\ref{fig:recovery-rate}). Claude~Sonnet~4.5 and Gemini~3~Pro recover in $>$90\% of cases, matching the human rate ($\sim$97\%). Weaker models (Gemini 2.5 Pro: 56\%, GPT-4.1 Nano: 12\%) rarely recover, explaining much of the accuracy gap.

\subsection{Query Reformulation Analysis}\label{subsec:reformulation-analysis}

\paragraph{Methodology.}
To quantify how aggressively systems reformulate their search queries, we embed all queries using Snowflake Arctic Embed (\texttt{snowflake-arctic-embed-m-v2.0}) and measure cosine distance between consecutive queries for each question trajectory. We define \emph{mean drift per step} as $\bar{d} = \frac{1}{k-1}\sum_{i=1}^{k-1}(1 - \cos(q_i, q_{i+1}))$ for a trajectory with $k$ queries.

\begin{figure}[h]
    \centering
    \includegraphics[width=\linewidth]{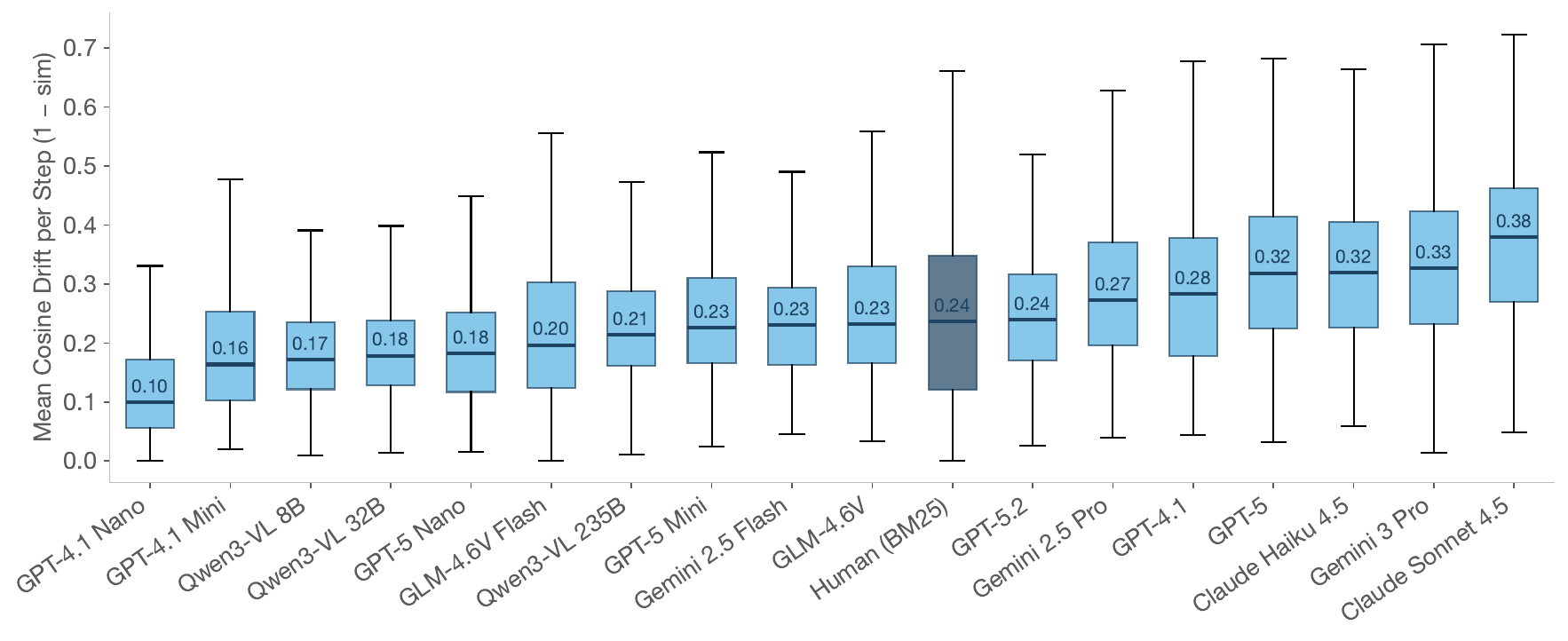}
    \caption{\textbf{Query reformulation magnitude by system.} Higher drift indicates more aggressive query changes. Box shows IQR with median labeled. Systems sorted by median drift.}
    \label{fig:drift-by-system}
\end{figure}

\begin{figure}[h]
    \centering
    \includegraphics[width=\linewidth]{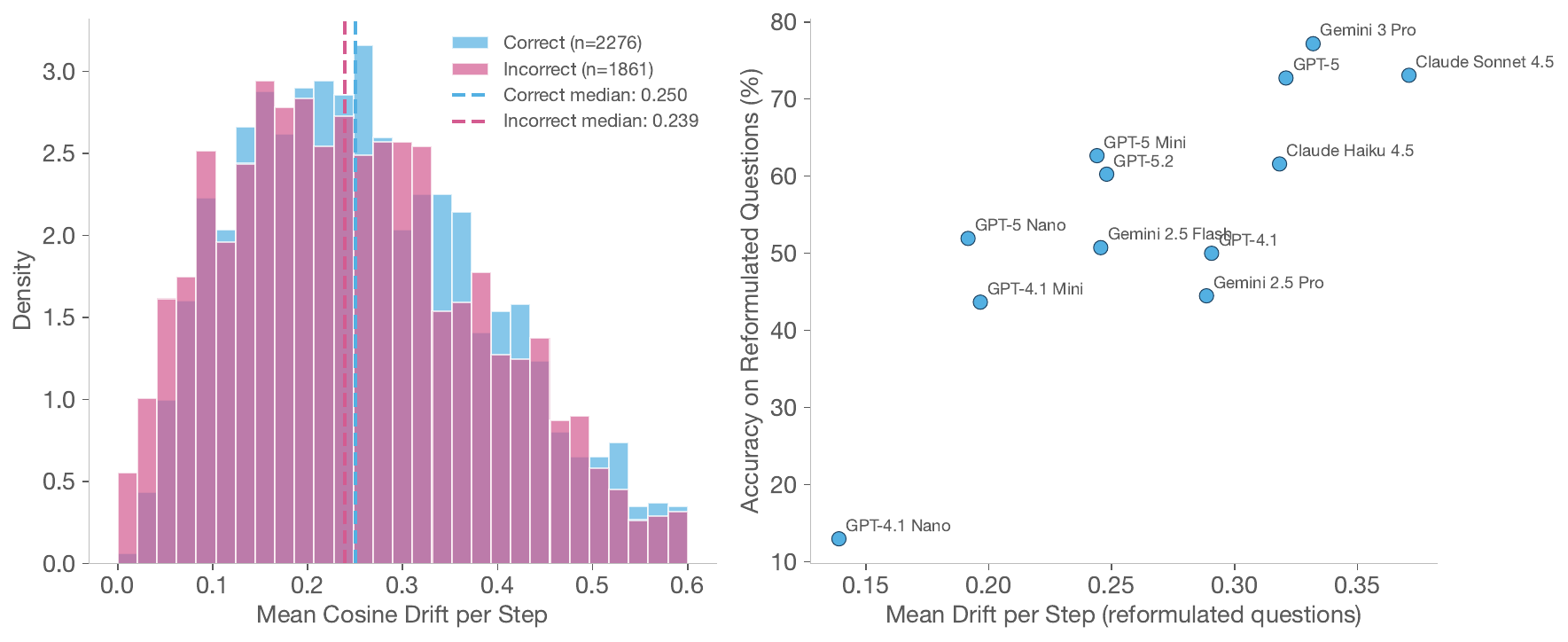}
    \caption{\textbf{Left:} Drift distribution for correct vs.\ incorrect predictions. \textbf{Right:} Per-system mean drift vs.\ accuracy on reformulated questions, showing a positive correlation between reformulation effort and success.}
    \label{fig:drift-vs-success}
\end{figure}

\paragraph{Reformulation Magnitude by System.}
Figure~\ref{fig:drift-by-system} shows the distribution of mean drift per step across systems, ordered by median. Top-performing systems reformulate more aggressively: Claude~Sonnet~4.5 has a median drift of 0.38, while GPT-4.1~Nano barely changes its queries (median drift 0.10). The correlation between reformulation magnitude and accuracy is strong (Figure~\ref{fig:drift-vs-success}, right panel).

\paragraph{Single-Query vs.\ Multi-Query Accuracy.}
Every system performs better on questions answered in a single query. The gap quantifies multi-step difficulty: for example, Human achieves 86.4\% on single-query questions vs.\ 79.5\% on multi-query, while GPT-4.1~Nano drops from 46.1\% to 13.2\%---a 33 percentage point gap that reveals its inability to effectively reformulate.

%% file: appendix/solution-space.tex
\clearpage
\section{Solution Space}\label{app:solution-design-space}

This design space provides a structural lens for interpreting the empirical findings presented in the main text. The comparative analysis highlights critical trade-offs across these specific dimensions:

While flexible \textit{Programmatic} flows (e.g., Recursive Language Models) offer theoretical advantages, our efficiency analysis demonstrates that they currently suffer from catastrophic computational overhead compared to constrained \textit{Iterative} architectures (e.g., BM25 MLLM Agent). This suggests that without robust guardrails, unconstrained reasoning leads to the diminishing returns observed in the efficiency calibration curves.
    
The necessity of the \textit{Visual} primitive is validated by our Construct Validity analysis, which finds that 57.2\% of questions require structural or visual comprehension (e.g., tables, forms). This confirms that systems relying solely on \textit{Textual} representations (Markdown/JSON) are architecturally limited compared to those utilizing full page images or visual crops.
    
With retrieval failures accounting for 39.1\% of all errors, the selection of \textit{Indexing} and \textit{Scope} is paramount. The performance hierarchy suggests that simple Sparse indexing often fails to capture semantic nuance, necessitating more advanced \textit{Late Interaction} or hybrid strategies (like HEAVEN) to handle the high recall requirements of the benchmark.

Finally, the pervasive issue of ``stochastic search'' identified in our behavioral analysis highlights the \textit{Memory} sub-dimension as a critical area for future optimization. Implementing \textit{Episodic Stores} could allow agents to learn corpus-specific terminology across queries, reducing the wasteful exploratory loops observed in current frontier models.

\begin{table*}[h]
\centering\footnotesize
\caption{Morphological Analysis of the Agentic Document Collection Answering Solution Space}
\label{tab:solution_space}
\renewcommand{\arraystretch}{1.2}
\begin{adjustbox}{width=\textwidth}
\begin{tabularx}{\textwidth}{@{}l l X@{}}
\toprule
\textbf{Dimension} & \textbf{Sub-Dimension} & \textbf{Primitives / Options} \\ 
\midrule
\textbf{1. Control Flow} & Architecture & Single-step RAG, Iterative (ReAct), Hierarchical (Manager-Worker), Map-Reduce \\
 & Reasoning & Zero-shot, Chain-of-Thought, Tree-of-Thoughts, Reflexion \\
 & Memory & Stateless, Sliding Window, Summarized History, Episodic Store \\ 
\midrule
\textbf{2. Representation} & Textual & Plain Text, Markdown, JSON, HTML, Semantic Chunks \\
 & Visual & Full Page Image, Bounding Box Crops, Latent Visual Embeddings \\
 & Graph & Knowledge Graph (Entities), Document Structure Graph (DOM) \\ 
\midrule
\textbf{3. Retrieval} & Indexing & Sparse (BM25), Dense (Bi-encoder), Late Interaction (ColBERT/Multi-vector) \\
 & Reranking & None, Cross-Encoder, LLM-based Reranking \\
 & Scope & Chunk-level, Page-level, Document-level, Parent-Child \\ 
\midrule
\textbf{4. Generation} & Synthesis & Extractive (Span selection), Abstractive (New synthesis) \\
 & Grounding & Citation (Sentence-level), Citation (Chunk-level), Visual Bounding Boxes \\
\bottomrule
\end{tabularx}
\end{adjustbox}
\end{table*}

%% file: appendix/related-works.tex

\clearpage

\section{Landscape of Agentic Approaches to Documents}

\input{sections/related}

\subsection{Related Works Assessment}\label{app:related}

Table~\ref{tab:benchmark_comparison} summarizes prior benchmarks along three axes:
(1) document diversity in terms of subject domains and layouts;
(2) whether questions and answers are fully human-annotated or partially automated;
and (3) the task framing (single-document VQA vs.\ document RAG vs.\ agentic research).
We briefly justify our assignments here.

\begin{figure}[h]
    \centering
    \includegraphics[width=\linewidth]{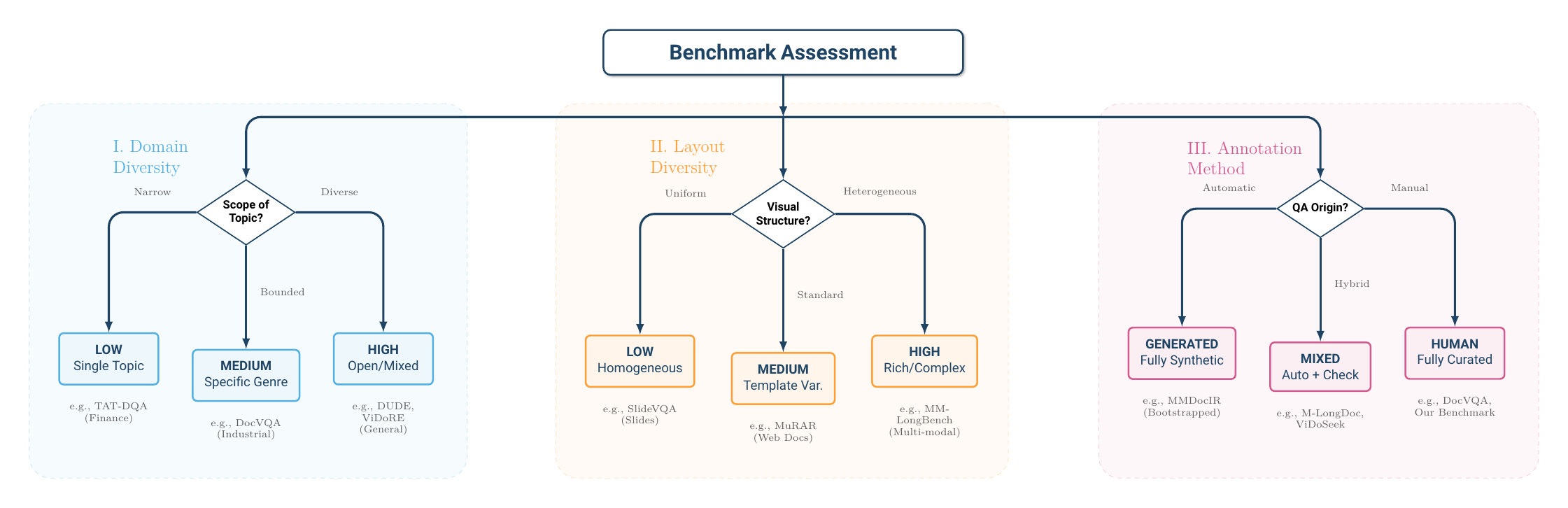}
    \caption{\textbf{Taxonomy of document-based benchmarks.} We structure our assessment of related works along three axes: (I) Domain Diversity, ranging from single-topic to open-domain; (II) Layout Diversity, evaluating visual structural complexity; and (III) Annotation Method, distinguishing between synthetic, hybrid, and fully human-curated pipelines.}
    \label{fig:asessment}
\end{figure}

We use \low{}, \medium{}, and \high{} as coarse, relative categories:
\low{} indicates a single or very narrow domain or layout type;
\medium{} covers a small number of fairly similar domains/layouts;
\high{} corresponds to broad open-domain coverage or many heterogeneous layouts (See Figure~\ref{fig:asessment}).

\input{appendix/related-works-rationale}

%% file: sections/related.tex


Over the past few years, document understanding approaches have progressed from systems combining OCR with layout-aware transformers \citep{Xu_2020,powalski2021goingfulltiltboogiedocument,Garncarek_2021} to specialized OCR-free models \citep{kim2022ocrfreedocumentunderstandingtransformer,lee2023pix2structscreenshotparsingpretraining}, and most recently, MLLMs \citep{chen2024internvlscalingvisionfoundation,bai2023qwenvlversatilevisionlanguagemodel,wu2024deepseekvl2mixtureofexpertsvisionlanguagemodels,openai2024gpt4technicalreport,liu2023visualinstructiontuning}. Despite this architectural evolution, the evaluation remains largely focused on answering questions based on a single input document.

The benchmarks in this area evolved from simpler single-page documents \citep{mathew2021docvqadatasetvqadocument, mathew2021infographicvqa} to slightly longer \citep{tito2023hierarchicalmultimodaltransformersmultipage, vanlandeghem2023documentunderstandingdatasetevaluation}, and massive PDFs \citep{chia2024mlongdocbenchmarkmultimodalsuperlong, ma2024mmlongbenchdocbenchmarkinglongcontextdocument}. 
Although emerging datasets introduce document collections as input, they often rely on semi-automated annotation pipelines \citep{cho2024m3docragmultimodalretrievalneed, dong2025mmdocirbenchmarkingmultimodalretrieval} or frame the problem as single-step retrieval \citep{faysse2025colpaliefficientdocumentretrieval}, failing to capture the iterative planning required for complex reasoning.

To address this complexity, recent research has adopted agentic frameworks  \citep{yao2023reactsynergizingreasoningacting,shinn2023reflexionlanguageagentsverbal}, which enable models to interleave reasoning, retrieval, and self-correction. However, the benchmarks used to evaluate these capabilities \citep{10.1145/3726302.3730275, su2025brightrealisticchallengingbenchmark}, predominantly operate over HTML web pages or plain text.


%% file: appendix/related-works-rationale.tex

\textbf{DocVQA~\cite{mathew2021docvqadatasetvqadocument}.}
DocVQA collects scanned business documents such as invoices, forms, and letters.
The subject matter is mostly industrial/administrative, but spans multiple document types,
so we mark domain diversity as \medium{}.
Layouts vary across templates but remain in the family of structured forms and reports,
hence \medium{} layout diversity.
Questions and answers are human-written and grounded in each single page, so we mark the dataset as fully human-annotated (\cmark).

\textbf{InfographicVQA~\cite{mathew2021infographicvqa}.}
This benchmark uses infographics as the sole document type.
Infographics cover a fairly wide range of topics but are all designed as visual posters,
leading to \medium{} domain diversity and \high{} layout diversity.
Questions and answers are manually annotated on each infographic, so we use \cmark.

\textbf{TAT-DQA~\cite{Zhu_2022}.}
TAT-DQA focuses on question answering over financial tables from corporate reports.
Both the domain (finance) and the layouts (tabular pages with similar structures)
are narrow, which we denote as \low{} domain and \low{} layout diversity.
QA pairs are human-annotated, so we mark \cmark.

\textbf{DUDE~\cite{vanlandeghem2023documentunderstandingdatasetevaluation}.}
DUDE includes multi-page PDFs from several domains (e.g., legal, scientific,
technical), with rich layouts involving paragraphs, tables, and figures.
We therefore assign \high{} diversity for both domains and layouts.
Questions and answers are manually curated, so the dataset is fully human-annotated (\cmark).

\textbf{MP-DocVQA~\cite{tito2023hierarchicalmultimodaltransformersmultipage}.}
MP-DocVQA extends DocVQA to multi-page documents but still focuses primarily
on industrial PDFs (forms, reports).
We rate both domain and layout diversity as \medium{}.
Many QA pairs originate from DocVQA and are extended heuristically to multiple pages,
so we mark the annotation as mixed (\cmark{} / \xmark{}).

\textbf{SlideVQA~\cite{tanaka2023slidevqadatasetdocumentvisual}.}
SlideVQA consists of slide decks (PowerPoint-style documents) from a variety of topics,
yielding \medium{} domain diversity but relatively homogeneous slide layouts (\low{}).
Questions are collected and verified by humans, thus the dataset is fully human-annotated (\cmark).

\textbf{M-LongDoc~\cite{chia2024mlongdocbenchmarkmultimodalsuperlong}.}
M-LongDoc comprises recent, very long multimodal documents with hundreds of pages
from three domains: academic research papers, financial reports, and product manuals.
These three domains justify a \medium{} rating for domain diversity.
The documents exhibit multiple structures (multi-column text, tables, figures, diagrams),
but still within a limited set of genres, so we mark layout diversity as \medium{}.
Questions are generated by several powerful LMMs (Claude 3.5, GPT-4o, Gemini 1.5 Pro)
and then filtered by both automated checks and human annotators, i.e., a semi-automatic pipeline,
so we mark the annotation as mixed (\cmark{} / \xmark{}).

\textbf{MMLongBench-Doc~\cite{ma2024mmlongbenchdocbenchmarkinglongcontextdocument}.}
MMLongBench-Doc is constructed over 130+ long PDF documents from seven diverse domains
with rich multimodal layouts (text, tables, charts, images).
Questions are expert-annotated and carefully designed to require evidence from multiple pages
and multiple modalities.
We therefore assign \high{} domain and \high{} layout diversity, and \cmark{} for fully human-annotated QA.

\textbf{MuRAR~\cite{zhu2025murarsimpleeffectivemultimodal}.}
MuRAR is a multimodal RAG framework evaluated on a knowledge base built from Adobe
Experience League documentation pages.
Because all content comes from a single enterprise product ecosystem, we assign \low{}
domain diversity.
Pages mix text, screenshots, tables, and videos inside a single website template,
which we label as \medium{} layout diversity.
Questions and answers are largely generated and refined by LLMs and then evaluated by humans,
so we mark mixed annotation (\cmark{} / \xmark{}).

\textbf{M$^2$RAG~\cite{liu2025benchmarkingretrievalaugmentedgenerationmultimodal}.}
M$^2$RAG builds on ELI5-style open-domain questions and retrieves multimodal evidence
(web pages plus images) across many topical categories (science, society, health,
politics, etc.).
This leads us to mark both domain and layout diversity as \high{}.
Questions are human-written, but document retrieval, multimodal evidence construction,
and some labels rely on automatic pipelines and LLMs, so we use \cmark{} / \xmark{}.

\textbf{ViDoRE~\cite{faysse2025colpaliefficientdocumentretrieval}.}
ViDoRE is the visual document retrieval benchmark introduced in the ColPali work
and aggregates several page-level tasks (DocVQA, InfographicVQA, TAT-DQA, TabFQuAD,
and synthetic DocQA datasets) across multiple domains and languages.
Documents include forms, infographics, tables, reports, and synthetic PDFs,
so we mark \high{} for both domains and layouts.
Since ViDoRE inherits human QA from existing datasets and also introduces synthetic queries
and QA, we regard the benchmark as having mixed human and automated annotation (\cmark{} / \xmark{}).

\textbf{ViDoRE v3~\cite{vidorev3}.}
Queries are generated via contextually-blind LLM prompting (conditioned on document
descriptions, not the documents themselves), then filtered and verified by multiple human
annotators who also provide reference answers and bounding-box annotations.
While this mitigates extractive bias, the generating model still influences the distribution
of question types and linguistic patterns. We therefore retain the mixed annotation
mark (\cmark{} / \xmark{}), acknowledging that the human component is considerably
stronger than in most benchmarks with this designation.

\textbf{DocBench~\cite{zou2024docbenchbenchmarkevaluatingllmbased}.}
DocBench collects 229 real documents from five domains (scientific articles, annual reports,
legal documents, government reports, and newspaper front pages) with heterogeneous layouts
(multi-column text, figures, tables, charts).
We thus assign \high{} domain and \high{} layout diversity.
Questions and reference answers are partly generated by GPT-4 and subsequently refined
or checked by human annotators, so we mark \cmark{} / \xmark{}.

\textbf{M3DocRAG / M3DocVQA~\cite{cho2024m3docragmultimodalretrievalneed}.}
M3DocRAG introduces M3DocVQA, an open-domain DocVQA benchmark created by taking
MultimodalQA question–answer pairs and converting their supporting Wikipedia pages into PDFs.
Because all documents originate from Wikipedia, topics are broad but still constrained
to encyclopedic style, which we mark as \medium{} domain diversity and \low{} layout diversity
(Wikipedia and its rendered PDFs share fairly uniform templates).
The QA origin is human (from MultimodalQA), but the new open-domain, multi-document setting
and PDF rendering are built automatically, so we use \cmark{} / \xmark{}.

\textbf{MMDocIR~\cite{dong2025mmdocirbenchmarkingmultimodalretrieval}.}
MMDocIR is a multimodal retrieval and DocRAG benchmark constructed by combining many
existing document-understanding datasets (including DocVQA, InfographicVQA, TAT-DQA,
MMLongBench-Doc, DocBench, etc.) and adding 1.6k+ manually annotated queries plus
a much larger automatically bootstrapped set.
It inherits broad domain and layout heterogeneity, so we assign \high{} / \high{}.
Given the mixture of expert labels and bootstrapped or inherited labels, we mark
\cmark{} / \xmark{}.

\textbf{FinRAGBench-V~\cite{zhao2025finragbenchvbenchmarkmultimodalrag}.}
FinRAGBench-V is a financial visual RAG benchmark: its corpus consists of financial
reports, statements, regulatory filings, and related documents, all within the financial
domain but with many kinds of charts, tables, and text-heavy pages.
We therefore mark domain diversity as \low{} (single domain) but layout diversity as \high{}.
The accompanying QA dataset is explicitly described as high-quality and human-annotated,
so we assign \cmark.

\textbf{BRIGHT~\cite{su2025brightrealisticchallengingbenchmark}.}
BRIGHT is a reasoning-intensive \textit{retrieval} benchmark: it comprises 1,384 real-world
queries from 12 datasets spanning biology, earth science, economics, psychology, robotics,
StackOverflow, AoPS, TheoremQA, and more, with documents that include blogs, news articles,
documentation, and reports.
We thus assign \high{} domain diversity and \medium{} layout diversity (web and text-centric pages).
Queries and relevance judgments are built from human-authored sources (no synthetic questions),
so we mark BRIGHT as fully human-annotated (\cmark).

\textbf{Researchy Questions~\cite{10.1145/3726302.3730275}.}
Researchy Questions starts from real user queries in commercial search logs and filters them
to non-factoid, decompositional ``researchy'' questions using GPT-3 / ChatGPT classifiers and GPT-4
for the final filtering stage.
Each question is linked to clicked URLs in ClueWeb22, giving a large, open-domain web corpus
with many site types and layouts; we assign \high{} domain and \high{} layout diversity.
Because questions and clicks are human, but the filtering and decompositional labels
rely heavily on LLMs, we mark mixed annotation (\cmark{} / \xmark{}).
The dataset does not ship reference answers, only queries and document signals.

\textbf{ViDoSeek~\cite{wang2025vidoragvisualdocumentretrievalaugmented}.}
ViDoSeek is a benchmark for visually rich document retrieval–reason–answer, consisting of
PDF documents (often slides, but also standards, technical reports, and whitepapers)
and annotated queries with single- or multi-hop types, reference answers, and reference pages.
Documents span many technical and governmental topics and combine 2D layout, charts, tables,
and free text, so we assign \high{} domain and \high{} layout diversity.
Queries and answers are produced via an LLM-assisted pipeline with human filtering
and quality control, hence we mark \cmark{} / \xmark{}.

\textbf{Our benchmark.}
Our benchmark is built from a curated collection of complex, multi-page PDFs drawn from
heterogeneous real-world sources (e.g., technical reports, standards, slide decks, and
research documents), which we categorize as \high{} domain and \high{} layout diversity.
Importantly, all questions are written by humans specifically for this benchmark, and the
documents themselves are selected by humans to support challenging multi-hop, agentic reasoning.
We therefore mark the dataset as fully human-annotated (\cmark).